\documentclass{article}

\PassOptionsToPackage{numbers, sort&compress}{natbib}
\usepackage[main, final]{neurips_2026}

\usepackage[utf8]{inputenc}
\usepackage[T1]{fontenc}
\usepackage{hyperref}
\usepackage{xurl}        %
\usepackage{booktabs}
\usepackage{array}
\usepackage{multirow}
\usepackage{amsfonts}
\usepackage{amsmath}
\usepackage[table]{xcolor}
\usepackage{amssymb}
\usepackage{amsthm}
\usepackage{nicefrac}
\usepackage{microtype}
\usepackage{xcolor}
\usepackage{graphicx}
\usepackage{xspace}
\usepackage{booktabs}
\usepackage[ruled,linesnumbered,noend]{algorithm2e}
\usepackage{graphicx}
\usepackage{placeins}
\usepackage[table]{xcolor}
\usepackage{booktabs}
\usepackage{calc}
\usepackage{tcolorbox}
\usepackage{subcaption}
\usepackage{wrapfig}
\usepackage{authblk}

\tcbuselibrary{skins}
\newtcbox{\rblock}[1][]{enhanced,frame hidden,boxrule=0pt,arc=3pt,boxsep=0pt,left=0pt,right=0pt,top=0pt,bottom=0pt,colback=white,clip upper,nobeforeafter,tcbox raise base,#1}

\DeclareMathOperator*{\argmin}{arg\,min}

\newcommand{\tabref}[1]{Table~\ref{#1}}
\newcommand{\secref}[1]{Section~\ref{#1}}
\newcommand{\figref}[1]{Figure~\ref{#1}}
\newcommand{\apref}[1]{Appendix~\ref{#1}}

\newcommand{\thref}[1]{Theorem~\ref{#1}}
\newcommand{\algref}[1]{Algorithm~\ref{#1}}

\newcommand{\M}{RSBI\xspace}

\definecolor{commentgreen}{RGB}{60,130,80}

\newcommand{\D}{\mathcal{D}_\epsilon}
\newcommand{\R}{ \mathcal{R}_p }
\theoremstyle{definition}
\newtheorem{definition}{Definition}[section]
\newtheorem{proposition}{Proposition}[section]

\newtheorem{theorem}{Theorem}[section]

\title{Two-Phase Rapid Simulation-Based Inference with Differentiable Simulators}

\author[1, 2, 3]{\textbf{Uros Zivanovic}}
\author[3, 4]{\textbf{Rahul Srinivasan}}
\author[3, 5, 6, 7]{\textbf{Roberto Trotta}}
\author[3]{\textbf{Andre Scaffidi}}

\affil[1]{Max Planck Institute for Intelligent Systems, Tübingen, Germany}
\affil[2]{Max Planck Institute for Gravitational Physics (Albert Einstein Institute), Potsdam, Germany}
\affil[3]{Scuola Internazionale Superiore di Studi Avanzati (SISSA), Trieste, Italy}
\affil[4]{IFPU - Institute for Fundamental Physics of the Universe, Trieste, Italy}
\affil[5]{INFN -- National Institute for Nuclear Physics, Italy}
\affil[6]{Italian Research Center on High-Performance Computing, Big Data and Quantum Computing, Casalecchio di Reno, Italy}
\affil[7]{Imperial College London, United Kingdom}

\begin{document}
\addtocontents{toc}{\protect\setcounter{tocdepth}{-1}}

\maketitle

\begin{abstract}

We introduce Rapid Simulation Based Inference (\M), a diffusion-based variational approach to likelihood-free Bayesian inference that achieves high sampling efficiency under large prior-to-posterior volumes. 
\M builds on advances in Schrödinger Bridge (SB) diffusion sampling to handle multi-modal posteriors, with an initial likelihood supplied by a  distance measure via a differentiable simulator. The key innovation of this initial surrogate likelihood is that it gives a highly efficient proposal distribution, replacing sequential rounds of other methods with an efficient one-shot proposal. Subsequent refinement then transports the proposal to the true posterior as specified by the simulator and the original prior. The SB sampler is warm-started from the proposal, and its energy is corrected by a neural likelihood ratio that is learned only where the proposal places mass. 
As a variational method, \M does not suffer from prior leakage commonly observed in standard sequential posterior estimation methods. To improve mode coverage we optionally utilize the well-tempered meta-dynamics framework, which also encourages exploration of the prior volume. We observe strong performance on standard SBI benchmarks, particularly when the prior volume is increased relative to the ground truth posterior, achieving a performance degradation of only $\sim5\%$ on Two Moons under a highly constrained simulation budget. Additionally, we evaluate \M's performance for gravitational wave ring-down posterior estimation, in a real-world physics benchmark inspired by black hole spectroscopy.

\end{abstract}

\section{Introduction}
\label{sec:introduction}

Many domains in science and engineering rely on stochastic simulators whose
sampling distribution $p(x\mid\theta)$ is intractable but from which data samples $x$ can be
drawn given parameters~$\theta$.  Bayesian parameter inference in
such settings is the subject of \emph{Simulation-Based Inference}
(SBI)~\citep{doi:10.1073/pnas.1912789117}.  Neural SBI methods have matured rapidly over the past five years. Sequential Neural Posterior Estimation
\citep[SNPE;][]{DBLP:conf/nips/PapamakariosM16,DBLP:conf/icml/GreenbergNM19} targets the posterior,
Sequential Neural Likelihood Estimation \citep[SNLE;][]{DBLP:conf/aistats/PapamakariosS019} the likelihood,
and Sequential Neural Ratio Estimation \citep[SNRE;][]{DBLP:conf/icml/HermansBL20,DBLP:conf/icml/Durkan0P20}
the likelihood-to-evidence ratio. Sequential Neural Variational Inference
\citep[SNVI;][]{DBLP:conf/iclr/GlocklerDM22} combines these with variational inference (VI).

A common difficulty arises when the  prior-to-posterior volume ratio $(\R)$ is large: when
$p(\theta)$ is broad compared to the posterior, most simulator calls fall in regions of negligible
posterior density, and efficiency vanishes. Sequential variants can reduce this cost by retraining the
proposal across rounds, but this comes at a price. SNPE with atomic proposals suffers from posterior
\emph{leakage} outside prior support~\citep{DBLP:conf/icml/GreenbergNM19}, and naive implementations
of variational approaches collapse onto a single mode of the posterior~\citep{DBLP:conf/iclr/GlocklerDM22}.
Truncation schemes~\citep{miller2021truncated, deistler2022truncated, dax2025realtime} instead restrict
the effective prior, so their result is exact only if the truncated region contains the posterior,
or else must be corrected at inference time, for example by importance sampling~\citep{dax2023neural}.

A growing class of scientific simulators are \emph{differentiable}: GPU-native models written within the JAX~\citep{jax2018github}/PyTorch~\citep{Ansel_PyTorch_2_Faster_2024} frameworks, learnable physical simulators, and neural
emulators trained to mimic expensive codes all expose the gradient
$\nabla_\theta\,s(\theta)$ of the parameters $\theta$ under the simulator output, $s(\theta)$.  This represents an opportunity for SBI: algorithms can exploit gradients directly rather than
treating the simulator as a black box.

\begin{wrapfigure}{l}{0.55\textwidth}
\vspace{-10pt}
\centering
\includegraphics[width=\linewidth]{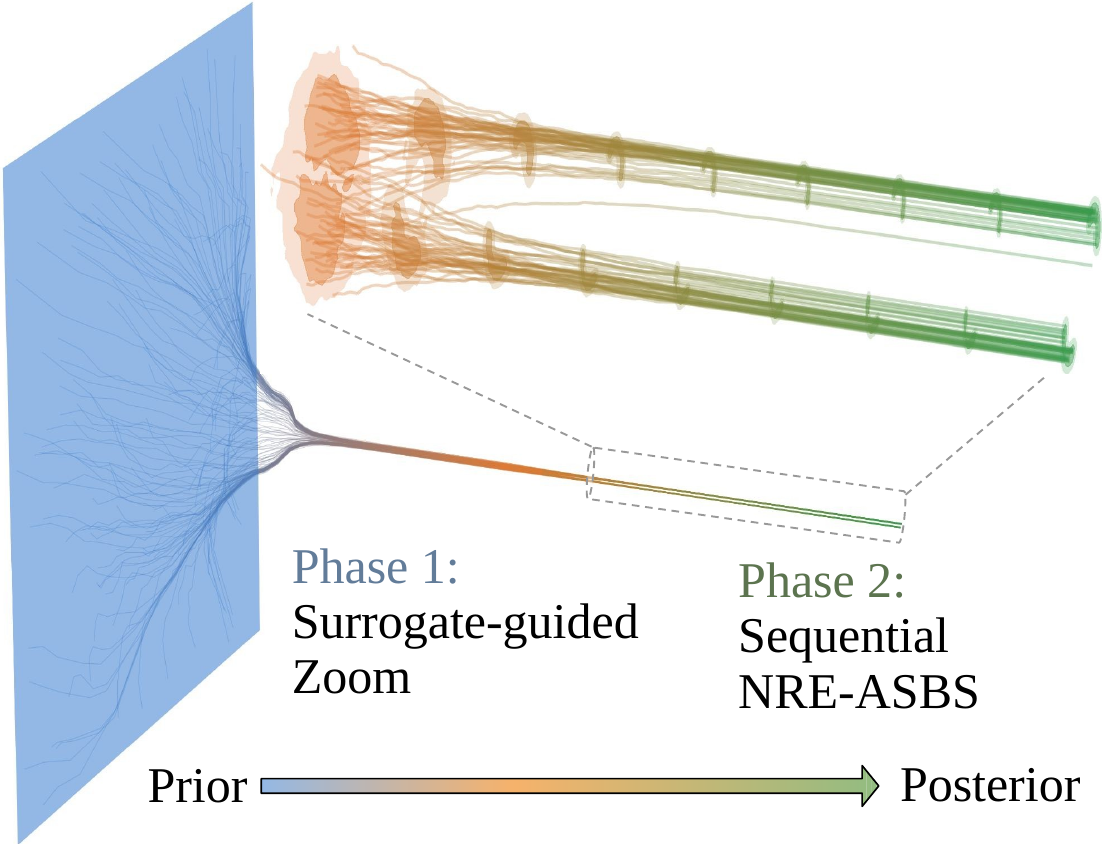}
\caption{Actual diffusion trajectories from the \M pipeline on the Two Moons task detailed in Sec.\ref{sec:experiments}. \M is able to produce a high quality posterior despite the large prior size.}
\label{fig:modes}
\end{wrapfigure}
Recent advances in neural samplers, in particular the Adjoint Schr\"odinger
Bridge Sampler (ASBS;~\citealp{liu2026adjoint}), provide an opportunity to
exploit such gradients. ASBS recasts sampling from an unnormalised
target distribution as a Schr\"odinger bridge problem solved via stochastic optimal
control: a controlled SDE transports a tractable source distribution to the
target, with the control trained by adjoint matching and a learned
corrector~\citep{DBLP:conf/icml/HavensMYDS0WHAK25}. The recent
{Well-Tempered} variant (WT-ASBS;~\citealp{nam2026enhancing}) augments
ASBS with a metadynamics-style bias~\citep{PhysRevLett.100.020603}, overcoming free-energy barriers between modes during training and enabling broad mode coverage at fixed simulation cost. 
Two properties make this family particularly well-suited to SBI. First, ASBS
queries the target only through gradients of its energy (the negative log of the unnormalised target density), $\nabla_\theta E$,
which a differentiable simulator supplies.  Second, the well-tempered bias
addresses the mode-seeking pathology that limits standard variational
SBI~\citep{DBLP:conf/iclr/GlocklerDM22} on multi-modal posteriors. To our
knowledge, neither ASBS nor its well-tempered extension have previously
been deployed for likelihood-free Bayesian inference.

We introduce \textbf{\M}, a two-phase SBI framework that leverages
differentiable simulators while keeping the target posterior fixed as
$p(\theta\mid x_o)$, where $x_o$ denotes the observed data.  Phase~1 places
posterior mass rapidly using a cheap surrogate likelihood, parameterized by bandwidth $\epsilon$ and built from 
distance metric $\D(s,x_o)$ between observed and simulated data, plugged into a Well-Tempered Adjoint Schr\"odinger Bridge Sampler~\citep{liu2026adjoint, nam2026enhancing} (WT-ASBS).  Phase~2 refines the resulting Phase-1 proposal using WT-ASBS, augmented by an NRE trained on simulator samples drawn from the `zoomed in' proposal. Figure.~\ref{fig:modes} outlines the process.

Our contributions are threefold:
\begin{itemize}
    \item \textbf{A two-phase SBI method for differentiable simulators.}
    We introduce \M (Sec.~\ref{sec:method}), a diffusion-based variational approach to likelihood-free
    Bayesian inference that achieves high sampling efficiency under large prior-to-posterior volume
    ratios.
    \item \textbf{First application of WT-ASBS to SBI.}
    We adapt the well-tempered metadynamics extension of ASBS~\citep{nam2026enhancing} to
    likelihood-free Bayesian inference, countering the mode-seeking pathology of reverse-KL
    variational SBI~\citep{DBLP:conf/iclr/GlocklerDM22} on multimodal posteriors.

    \item \textbf{Benchmarking robustness at large prior-to-posterior volume ratio $\R$.}
    We benchmark \M{} and other methods on Two Moons~\citep{pmlr-v130-lueckmann21a},
    SLCP~\citep{pmlr-v130-lueckmann21a, papamakarios2021normalizing}, Lotka--Volterra \cite{pmlr-v130-lueckmann21a},
    Bernoulli GLM~\citep{pmlr-v130-lueckmann21a}, and a new differentiable gravitational-wave
    ringdown task ($\dim\theta = 6$, $\dim x = 820$), sweeping prior-to-posterior volume ratios up
    to $\log_{10}\R \approx 12$.
\end{itemize}

\section{Related Work}
\label{sec:related_work}
\M sits at the intersection of three research threads. The first is
\emph{neural simulation-based inference}, in which a density or ratio
estimator is trained on simulated $(\theta, x)$ pairs to target the
posterior, likelihood, or likelihood-to-evidence
ratio~\citep{DBLP:conf/nips/PapamakariosM16,DBLP:conf/icml/GreenbergNM19,%
DBLP:conf/aistats/PapamakariosS019,DBLP:conf/icml/HermansBL20,%
DBLP:conf/icml/Durkan0P20,DBLP:conf/nips/MillerWF22}. Sequential variants
iteratively refine their proposal to concentrate simulations in informative
regions of parameter space. SNVI+fKL~\citep{DBLP:conf/iclr/GlocklerDM22}
pairs likelihood(-ratio) estimation with VI and is our
closest methodological sibling. Like \M, SNVI combines a learned neural likelihood ratio estimated via Neural Ratio Estimation (NRE) with a
variational sampler, and, like \M, it explicitly addresses the
mode-seeking pathology of standard reverse-KL VI, in its case by training
under the mass-covering forward KL with sampling--importance-resampling for
proposal updates. \M instead handles mode coverage through the
well-tempered metadynamics bias of WT-ASBS~\citep{nam2026enhancing}, which
flattens free-energy barriers between modes. The two
methods also differ in how the initial proposal is obtained: SNVI draws
its first round of simulations from the original prior, whereas \M is
initialised by a one-shot surrogate-guided Schrödinger bridge.

The second thread is the surrogate-likelihood lineage that connects
classical ABC~\citep{10.1093/genetics/162.4.2025} to modern VI. Any differentiable distance metric $\D(s, x_o)$ induces a pseudo-likelihood
whose gradients propagate through a simulator, an idea explored by
ABC-VI~\citep{moreno2016automaticvariationalabc}, 
as well as score-based approaches~\citep{welling2011sgld,10.1214/24-EJS2283}. \M borrows from this for Phase~1, but rather than amortising over
$x$ or running an MCMC against the surrogate, we use it to construct the
energy on which WT-ASBS is trained. This gives a non-mode-seeking
proposal that Phase~2 then transports to the true posterior, with a likelihood specified by an NRE learned only where the proposal places mass. This use of a surrogate kernel as a {warm-start for a downstream refinement},
rather than as the inference target itself, can alleviate the
bandwidth-versus-bias trade-off that limits standalone ABC-VI.

The third thread is the rapidly maturing ecosystem of GPU-native,
auto-differentiable simulators across many scientific fields and neural
emulators~\citep{Li_2024,PhysRevD.110.064028,10.1063/5.0277356,%
10.1093/mnras/staf2071}. Direct access to simulation gradients $\nabla_\theta\,s(\theta)$
enables gradient-based inference for models that previously admitted only
analytical-likelihood treatments:
ASBS~\citep{liu2026adjoint} queries its target
exclusively through energy gradients, which the differentiable simulator
supplies via  a distance metric $\D(s,x_0)$, sidestepping the density-estimation  step of standard SNPE/SNLE/SNRE.
A growing number of works that exploit gradients have been used to improve neural SBI. \citet{brehmer2020mining} augment the training losses of
ratio and likelihood estimators with the joint score and likelihood ratio, and relaxations such as the Gumbel-Softmax
\citep{jang2017categorical} extend such gradients to simulators with discrete randomness. To our knowledge, the only
gradient-informed NPE method adds a score-matching term to the NPE loss \citep{zeghal2022neural}, leaving the proposal,
the simulation allocation and the sampler unchanged. In all of these, gradients regularise an estimator fit to
prior-wide simulations. In RSBI they instead drive the Phase~1 sampler directly, through an auto-encoder-like
comparison of simulations with the observation.

\section{Background}
\label{sec:background}

\subsection{Simulation Based Inference}
\label{sec:background:bi}
We consider a stochastic simulator $s:\Theta\times\Omega\to\mathcal{X}$ with parameters
$\theta\in\Theta\subseteq\mathbb{R}^d$ and outputs $x=s(\theta,\omega)\in\mathcal{X}$, where
$\omega\in\Omega$ collects all internal randomness of the simulator (noise draws, random seeds)
and is drawn from a distribution $p(\omega)$ that does not depend on $\theta$.  The simulator implicitly defines the conditional density $p(x\mid\theta)$ and the distribution of
$s(\theta,\omega)$ under $\omega\sim p(\omega)$. The likelihood $\theta\mapsto p(x_o\mid\theta)$
is this density evaluated at the observed data $x_o$. It can be sampled by running the simulator but
not evaluated.
 We call the simulator
\emph{differentiable} if $\theta\mapsto s(\theta,\omega)$ is differentiable for every 
$\omega$, so that pathwise gradients $\nabla_\theta s(\theta,\omega)$ are available. We write $s(\theta)$ for $s(\theta,\omega)$ with $\omega\sim p(\omega)$,
and $s(\theta,\omega_{1:K}) = \{s(\theta,\omega_k)\}_{k=1}^K$ with $\omega_k\overset{\text{iid}}{\sim}p(\omega)$
for $K$ independent replicates at the same $\theta$. Given a prior $p(\theta)$ and observation
$x_o$, simulation-based inference (SBI) targets the  posterior
$p(\theta\mid x_o)\propto p(x_o\mid\theta)\,p(\theta)$ using only forward
simulations~\citep{doi:10.1073/pnas.1912789117}.

Classical approximate Bayesian computation (ABC) accepts prior draws $\theta\sim p(\theta)$ whose
simulations fall within a tolerance $\epsilon$ of the observation, $\D(s(\theta), x_o) < \epsilon$,
usually after reducing the data to summary statistics~\citep{Marin2011ApproximateBC}. It scales poorly
with $\dim(x)$ and is sensitive to the choice of summaries and of $\epsilon$. Neural SBI instead
trains a density estimator $q_\phi$ on a corpus of simulations $(\theta_i, x_i)$ with $\theta_i\sim p(\theta)$ and
$x_i\sim p(x\mid\theta_i)$, targeting the posterior (NPE)~\citep{DBLP:conf/nips/PapamakariosM16,
DBLP:conf/icml/GreenbergNM19}, the likelihood (NLE)~\citep{DBLP:conf/aistats/PapamakariosS019} or the
likelihood-to-evidence ratio (NRE)~\citep{DBLP:conf/icml/HermansBL20,DBLP:conf/icml/Durkan0P20,
DBLP:conf/nips/MillerWF22}. Amortised variants train once to learn $p(\theta)p(x\mid\theta)$ and apply to
any $x$, whereas sequential variants concentrate simulations around $x_o$, giving up amortisation
for simulation efficiency.

\subsection{Diffusion Sampling}

Neural diffusion samplers learn a Stochastic Differential Equation (SDE) which transports samples $X_{0}\sim\mu(x)$ from source distribution $\mu(x); $ at time $t=0$ to the Boltzmann distribution $v(x)$ at time $t=1$ given by:
\begin{equation}\label{eq:target_dist}
    v(x)=\frac{e^{-E(x)/T}}{Z},\quad\quad Z=\int_{\mathcal{X}}e^{-E(x)/T}dx<\infty
\end{equation}
The distribution is characterized by an unnormalized energy $E:\mathcal{X}\subseteq\mathbb{R}^{d}\to\mathbb{R}$ and temperature $T$, with the normalizing constant $Z$ being intractable.
Diffusion samplers aim to provide faster convergence using fewer evaluations of the energy function than Markov Chain Monte Carlo (MCMC)~\citep{10.1063/1.1699114} methods by learning better proposal distributions.
Especially when $v(x)$ is defined through an energy function, diffusion samplers provide a scalable alternative to sampling methods such as score matching~\citep{DBLP:journals/corr/abs-2402-08667,DBLP:conf/icml/PhillipsDHBDD24} or normalizing flows~\citep{DBLP:conf/aistats/PapamakariosS019, DBLP:conf/nips/PapamakariosM16} whose naive implementations rely on samples from $v(x)$.

\textbf{Adjoint Schr\"odinger Bridge Sampling:}
\label{sec:background:sb}
ASBS\citep{liu2026adjoint} is a state-of-the-art diffusion sampler building on Adjoint Sampling (AS)\citep{DBLP:conf/icml/HavensMYDS0WHAK25} that solves the Schr\"odinger Bridge (SB) problem between source distribution $\mu(x)$ and target $v(x)$ through stochastic optimal control.
Within the context of \M, we consider the following drift-less SDE:
\begin{equation}\label{eq:sde}
    dX_{t}=\sigma_{t}u_{t}(X_{t})dt+\sigma_{t}dW_{t},\quad\quad X_{0}\sim\mu(X_{0}),
\end{equation}
where $\sigma_{t}$ is the diffusion coefficient, $u_{t}$ is a learned control network, and $(W_{t})_{t}$ is a $d$-dimensional Brownian motion.
The optimal control $u_{t}^{*}$ is then given by:
\begin{subequations}\label{eq:asbs_objective}
\begin{equation}
    u^{*}=\argmin_{u}D_{\text{KL}}(p^{u}||p^{\text{base}})=\argmin_{u}\mathbb{E}_{\mathbb{X}\sim p^{u}}\left[\int_{0}^{1}\frac{1}{2}||u_{t}(X_{t})||^{2}dt \right],
\end{equation}
\begin{equation}
        \text{s.t.}\ dX_{t}=\sigma_{t}u_{t}(X_{t})dt+\sigma_{t}dW_{t},\quad X_{0}\sim\mu(X_{0}),\quad X_{1}\sim v(X_{1}).
\end{equation}
\end{subequations}
Therefore, $u^{*}$ minimizes the relative entropy between the path probability $p^{u}$ characterized by \eqref{eq:sde} and the base path probability $p^{\text{base}}$ given when $u=0$.
To solve \eqref{eq:asbs_objective}, ASBS defines an additional corrector network $h$ and iterates over two objectives; Adjoint Matching (AM):
\begin{equation}\label{eq:am_obj}
u^{(k)}=\argmin_{u}\mathbb{E}_{p_{t|0,1}^{\text{base}}p_{0,1}^{\bar{u}}}
\left[\|u_{t}(X_{t})+\sigma_{t}(\nabla_{X_{1}} E(X_{1})+h^{(k-1)}(X_{1}))\|^{2}\right],\quad \bar{u}=\texttt{stopgrad}(u),
\end{equation}
and Corrector Matching (CM):
\begin{equation}\label{eq:cm_obj}
 h^{(k)} = \argmin_{h}\mathbb{E}_{p_{0,1}^{u}}
\left[\|h(X_{1})-\nabla_{X_{1}}\log p^{\text{base}}(X_{1}|X_{0})\|^{2}\right].
\end{equation}

Iterating through AM and CM stages constitutes the well-known iterative proportional fitting\citep{248d76c2-8571-36f6-a45a-066e8859362f} algorithm.
\begin{theorem}[Convergence of ASBS]\label{th:asbs_convergence}
    As $k\rightarrow \infty$ and both AM\eqref{eq:am_obj} and CM\eqref{eq:cm_obj} stages reach their critical points, $u^{(k)}=u^{*}$, the SB solution of \eqref{eq:asbs_objective}.
\end{theorem}
For a proof of \thref{th:asbs_convergence} we refer to Liu et. al.\citep{liu2026adjoint}.
Note that ASBS relies only on the energy gradient $\nabla_{X_{1}} E(X_{1})$ during the AM stage, and does not require any energy evaluations during the CM stage.  This makes the algorithm efficient in terms of energy evaluations, and is what makes it highly suitable for use in \M, where we only have access to the likelihood and prior's gradient.  A consequence is that ASBS distinguishes between two kinds of low-density region. Because the energy
is queried only at the terminal point $X_1$, trajectories can pass through regions of low or zero
target density in transit without obstruction, unlike MCMC, whose chains must traverse every
intermediate state. Regions that separate components of the terminal distribution are different. Adjoint Matching trains
on trajectories drawn from the current controller, so a component that no trajectory reaches provides
no training signal and can remain uncovered. The well-tempered bias of Appendix.~\ref{sec:well_tempering}
addresses this by pushing samples towards regions not yet visited. 

\section{Method}
\label{sec:method}

\begin{figure}[t]
    \centering
    \includegraphics[width=.4\linewidth]{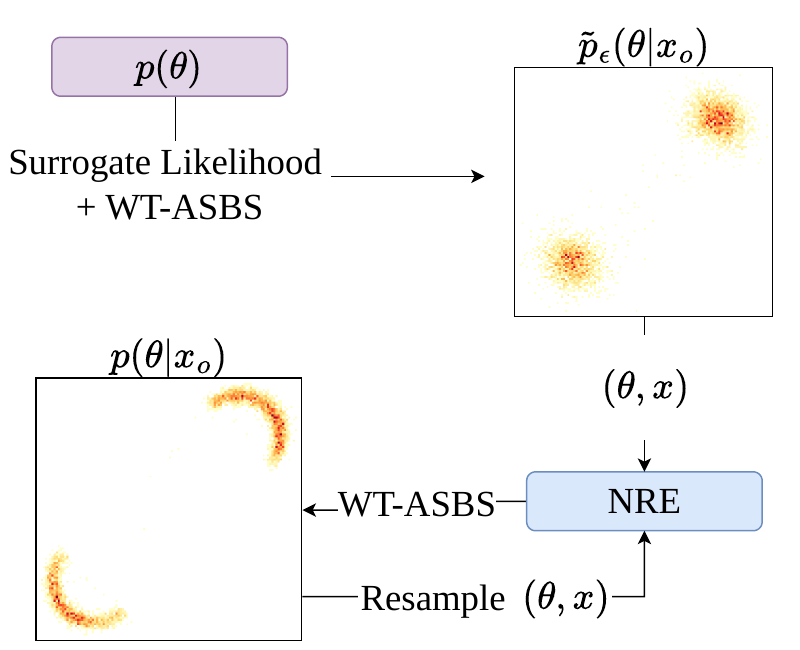}\includegraphics[width=0.6\linewidth]{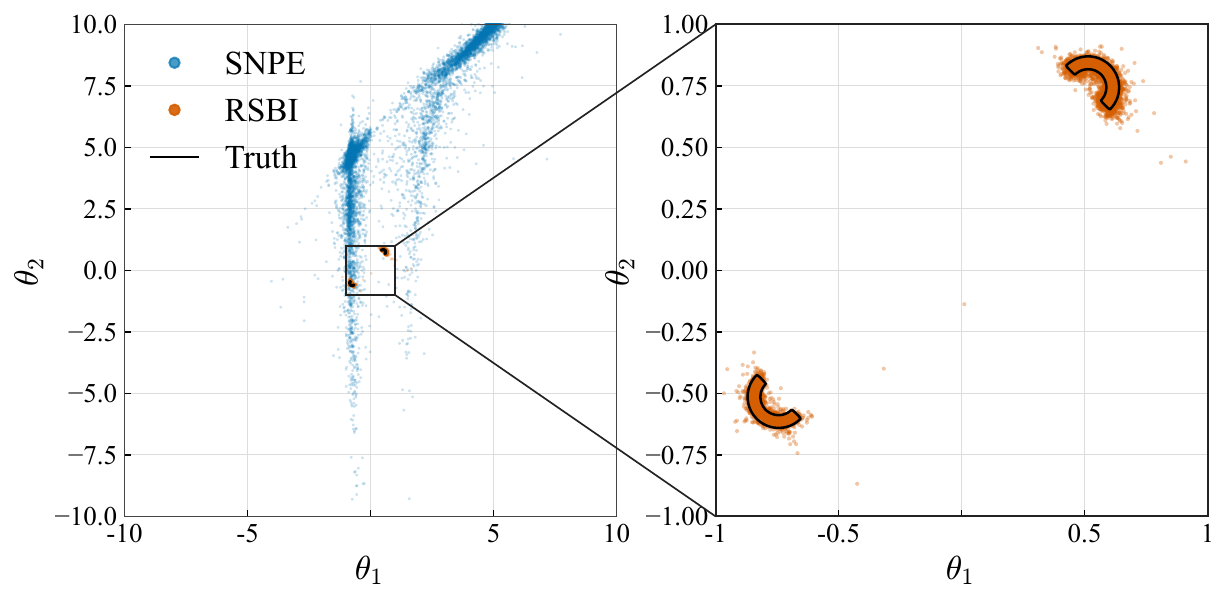}
    \caption{\textbf{Left:} Schematic representation of the two-phase procedure of RSBI. \textbf{Right:}  Comparison of posteriors from RSBI (our method) and SNPE on a Two Moons task with $10^4$ simulations and $ \R \sim 10^{4.3}$, with two sequential steps for both SNPE and Phase 2 of RSBI.    }
    \label{fig:method_zoom}
\end{figure}

\begin{algorithm}[b]
\caption{Well-Tempered ASBS - One Round}\label{alg:asbs}
\SetAlgoLined
\DontPrintSemicolon
\SetKwProg{Def}{def}{:}{}
\SetKwFor{For}{for}{:}{}
\SetKwIF{If}{ElseIf}{Else}{if}{:}{elif}{else:}{}
\SetKwComment{tcp}{\# }{}
\SetKwFunction{WTASBS}{wt\_asbs}
\SetKwFunction{RSBIfn}{rsbi}
\Def{\WTASBS{source $\mu$, energy gradient $\nabla E$, control $u$, corrector $h$, bias $V(\xi(\theta))$, stages $S$}}{
\For{stage $k$ in $1, 2, \dots, S$ and each epoch}{
Obtain $X_1$ by drawing $X_0\sim\mu$ and integrating \eqref{eq:sde}  \;
\uIf{$k$ is odd (AM stage)}{
Add $\bigl(X_0,\, X_1,\, \nabla E(X_1) + h(X_1) + \nabla(V\!\circ\!\xi)(X_1)\bigr)$ to the replay buffer\;
\lIf{$k < S$}{$V\leftarrow V+$ WT update \eqref{eq:wt_update_rule} at $\xi(X_1)$}
Update $u$ on the buffer through WT-AM \eqref{eq:wt_am_objective}\;
}
\lElse{add $(X_0, X_1)$ to the replay buffer and update $h$ by solving CM \eqref{eq:cm_obj}}
}
\KwRet{$u$, $h$, $V$, saved simulations $\mathcal{B}$}
}
\end{algorithm}

We adopt the SBI setting of Section~\ref{sec:background:bi}: a simulator
$s(\theta,\omega)$, prior $p(\theta)$, and observation $x_o$, with the
unnormalized posterior $p(\theta\mid x_o)$ as the inference target. We additionally
assume that $s$ is differentiable in $\theta$, so
that $\nabla_\theta\,s(\theta,\omega)$ is available for any 
$\omega\sim p(\omega)$.
A detailed illustration of \M can be found in \figref{fig:method_zoom}. We also provide simplified pseudocode for ASBS in \algref{alg:asbs} and for \M in \algref{alg:sgsbi}.

\subsection{Phase 1: Surrogate-Guided Zoom}
\label{sec:stage1}

When the prior-to-posterior volume ratio $\R$ is large, learning a neural likelihood or posterior
over the full prior support is impractical: most simulator calls fall in regions of negligible
posterior density. Phase~1 sidesteps this bottleneck with an analytic surrogate likelihood, in the
spirit of ABC.
Given a differentiable distance $\D(s(\theta), x_o)$ between simulated and observed data, we define
the \textit{surrogate likelihood}
\begin{equation}\label{eq:surrogate}
    \tilde{p}_\epsilon(x_o \mid \theta)
    \;\propto\;
    \exp\!\Big(-\,\mathbb{E}\,\D(s(\theta),\, x_o)\Big),
\end{equation}
where $s(\theta)$ inside $\D$ denotes $K$ replicates $s(\theta,\omega_{1:K})$ and the expectation is
over $\omega_{1:K}$. The bandwidth $\epsilon > 0$ is the scale on which $\D$ compares simulated and
observed data, and so controls how sharply $\tilde p_\epsilon$ concentrates around $x_o$. I.e, the noise
variance of a Gaussian negative log-likelihood, or the length scale of an energy distance or MMD
kernel. $\epsilon$ is either fixed or set adaptively from the replicates. We detail how the adaptive bandwidth is set for different surrogate likelihoods $\D$ and conduct ablations on it's choice in Appendix~\ref{app:surrogate}. We also discuss a heuristic method that exploits a cheap survey of the gradient landscape in order to select $\epsilon$ in Appendix~\ref{sec:energy_landscape_opt}. 

The surrogate posterior $\tilde{p}_\epsilon(\theta\mid x_o) \propto
\tilde{p}_\epsilon(x_o\mid\theta)\,p(\theta)$ has the Boltzmann form of \eqref{eq:target_dist}.
ASBS only queries the gradient of its energy, so $\D$ need only be known up to an additive constant:

\begin{theorem}[Equivalence under differentiation]\label{th:equivalence}
Let $\tilde{\D}:\mathcal{X}^K\times\mathcal{X}\to\mathbb{R}$ be differentiable and equal to a
positive-semi-definite (PSD) distance $\D$ up to a constant, $\D(x,x') = \tilde{\D}(x,x') + C$
for some $C\in\mathbb{R}$ and all $x,x'$. Then $\nabla\tilde{\D} = \nabla\D$.
\end{theorem}

The energy gradient can therefore be computed from any such $\tilde\D$,
\begin{equation}\label{eq:surrogate_energy}
    \nabla_\theta E_s(\theta)
    \;=\;
    \mathbb{E}\,\nabla_\theta \tilde{\D}(s(\theta),\, x_o)
    \;-\;
    \nabla_\theta \log p(\theta),
\end{equation}
where the gradient passes inside the expectation because $p(\omega)$ does not depend on $\theta$. We hand \eqref{eq:surrogate_energy} to ASBS with the prior
$p(\theta)$ as source (full derivation in Appendix~\ref{app:surrogate:derivation}). By
Theorem~\ref{th:asbs_convergence}, this converges to the Schr\"odinger-bridge solution that transports
$p(\theta)$ to the surrogate posterior. Although it may seem trivial, Theorem~\ref{th:equivalence} lets standard losses be used without
their constants, such as the Gaussian NLL without its normalising constant and MSE, alongside
PSD discrepancies such as the energy distance~\citep{https://doi.org/10.1002/wics.1375} and
MMD~\citep{DBLP:journals/jmlr/GrettonBRSS12}. A Gaussian NLL with adaptive variance is not of this form but still gives a proper surrogate posterior (Appendix~\ref{app:surrogate:nll})

Because  $\tilde{p}_\epsilon(\theta\mid x_o)$ is defined directly through the simulator rather than learned, Phase~1
needs no density estimator. Each simulator call moves the proposal towards $x_o$ instead of providing a
training example for an estimator fitted from prior draws, which is what limits sequential neural
methods at large $\R$. The distance
gradient points toward $x_o$ in data space, which the differentiable simulator pulls back into
parameter space via $\nabla_\theta s(\theta)$. Phase~1 therefore efficiently allocates posterior mass to
high-density regions under a constrained simulator budget, producing the focused proposal
$\tilde{p}_\epsilon(\theta\mid x_o)$ that Phase~2 refines. Appendix~\ref{app:surrogate:derivation} gives the
admissibility class for $\D$,  whilst Appendix.~\ref{app:surrogate:tradeoffs} discusses the bandwidth--metric trade-offs inherited from ABC, and the surrogates used in Section~\ref{sec:experiments}.

\begin{algorithm}[t]
\caption{Rapid Simulation-Based Inference}
\label{alg:sgsbi}
\SetAlgoLined
\DontPrintSemicolon
\SetKwProg{Def}{def}{:}{}
\SetKwFor{For}{for}{:}{}
\SetKwIF{If}{ElseIf}{Else}{if}{:}{elif}{else:}{}
\SetKwComment{tcp}{\# }{}
\SetKwFunction{WTASBS}{wt\_asbs}
\SetKwFunction{RSBIfn}{rsbi}
\Def{\RSBIfn{prior $p(\theta)$, simulator $s(\theta,\omega)$, observation $x_{o}$, surrogate distance $\D(s,x_o)$, temperature $T$, ratio network $r_\psi(\theta, x)$, rounds $R$}}{
\tcp{\textbf{Phase 1: Surrogate-guided zoom}}
$E_s(\theta) \leftarrow [\mathbb{E}\,\D(s(\theta),\, x_o) - \log p(\theta)] / T$\;
$\bar{u},\, \bar{V},\, \mathcal{B}\leftarrow$ \WTASBS{$\mu = p(\theta)$, $\nabla E = \nabla E_s$, new $u$, new $h$, $V = 0$}\;
Draw $\theta_i \sim \texttt{resample}_{\bar{V}}(p_1^{\bar{u}})$, top up $\mathcal{B}$ with new $(\theta_i, s(\theta_i))$ \;
$\mu \leftarrow p_1^{\bar{u}}$, optionally $\texttt{resample}_{\bar{V}}(p_1^{\bar{u}})$;\quad $u_0,\, h_0 \leftarrow$ fresh networks\;
\BlankLine
\tcp{\textbf{Phase 2: Sequential NRE-ASBS}}
\For{$r$ in $1, 2, \dots, R$}{
    Optimise $r_\psi(\theta, x)$ on $\mathcal{B}$ via the NRE-C loss \eqref{eq:nrec}\;
    $E_r(\theta) \leftarrow -\log r_\psi(\theta, x_o) - \log p(\theta)$ \eqref{eq:phase_two_energy}\;
    $u_r,\, h_r,\, V_r,\, \_ \leftarrow$ \WTASBS{$\mu$, $\nabla E = \nabla E_r$, $u_{r-1}$, $h_{r-1}$, $V = 0$}\;
    \lIf{$r < R$}{draw $\theta_i \sim p_1^{u_r}$, top up $\mathcal{B}$}
}
\KwRet{$\mu$, $u_R$, $V_R$}
}
\end{algorithm}

\subsection{Phase~2: Sequential NRE-ASBS Refinement}
\label{sec:stage2}

In Phase~2 we start from the surrogate posterior
$\tilde{p}_\epsilon(\theta \mid x_o)$ produced by Phase~1 and refine it
to recover the simulator posterior $p(\theta \mid x_o)$ (\secref{sec:background:bi}). Refinement proceeds over $R$
rounds. Each round trains a neural ratio estimator~\citep{DBLP:conf/icml/HermansBL20, DBLP:conf/icml/Durkan0P20}
$r_\psi(\theta, x)$ on $(\theta, x)$ pairs simulated so far, which are drawn from the round's proposal
$q_r(\theta)$ as well as optionally earlier ones used during phase 1 training. It then solves the ASBS objective \eqref{eq:asbs_objective}
against the resulting target energy, with $\tilde{p}_\epsilon(\theta \mid x_o)$
as the source distribution in every round. The SB therefore always transports samples from the Phase-1
posterior, and the sampler of round $r$ provides the proposal $q_{r+1}$ for the next round's simulations.
We use the contrastive (NRE-C)
loss~\citep{DBLP:conf/nips/MillerWF22} throughout. 
To improve mode-discovery over both phases we also optionally use a simplified version of the recent Well-Tempered metadynamics (WT-ASBS)~\citep{nam2026enhancing} extension that augments the energy landscape, for more details on WT-ASBS we refer to Appendix~\ref{sec:well_tempering}. 

At optimum the classifier learns
$r_\psi^\ast(\theta, x) = p(x \mid \theta) / q_{r}(x)$, where $q_{r}(x)$ is the local marginal. 
Because Phase~2 conducts multiple rounds of refinement, the proposal
$q_{r}(\theta)$ changes between rounds as simulations concentrate in
regions of high posterior probability.  Following the same
construction as the Phase~1 energy~\eqref{eq:surrogate_energy}, the
round-$r$ energy is:
\begin{equation}\label{eq:phase_two_energy}
    -\nabla_\theta \log p(\theta \mid x_o)
    \;=\;
    -\nabla_\theta \log\frac{p(x_o \mid \theta)}{q_{r}(x_o)}
    \;-\;\nabla_\theta \log p(\theta)
    \;=\;
    \nabla_\theta E_r(\theta),
\end{equation}
where the proposal-induced marginal $q_{r}(x_o)$ has zero gradient in
$\theta$ and so drops out - this is a core appeal of using a diffusion sampler against an NRE-derived energy. 
Solving \eqref{eq:asbs_objective} via
ASBS with energy $E_r(\theta)$ and source
$\tilde{p}_\epsilon(\theta \mid x_o)$ therefore recovers the
Schr\"odinger-bridge solution connecting the surrogate posterior and
the simulator posterior. To avoid retraining from scratch each round we
warm-start the controller $u$ and corrector $h$ from the previous
round's parameters.  
Because $r_\psi$ is only identified where the classifier has seen training parameters, a user who trusts the Phase~1
posterior to cover the simulator posterior can optionally truncate the ratio outside the region spanned by the Phase~1
samples (Appendix~\ref{app:nre_truncation}).  Because $E_r$ in~\eqref{eq:phase_two_energy}
explicitly retains $\log p(\theta)$ as an additive term, the round
output requires no post-hoc correction for leakage outside prior
support. Natural posterior leakage mitigation is an additional benefit of RSBI which we demonstrate with an additional experiment in Appendix \ref{app:leakage}.

Because Phase~2 \textit{transports} the Phase~1 sampler rather than refitting a global density over the prior, it
remains accurate where SNPE and SNVI, warm-started from the same Phase~1 samples
and given the same simulation budget, still degrade or collapse onto spurious regions of a wide prior. We demonstrate this in an ablation in Appendix~\ref{app:tm-phase2-ablation}.

The full Phase 2 procedure including NRE-C training, the well-tempered bias
update~\eqref{eq:wt_update_rule}, and the augmented Adjoint Matching
objective~\eqref{eq:wt_am_objective} is described in
Appendices~\ref{sec:well_tempering} and~\ref{app:nre_loss}, alongside the convergence
guarantee (Proposition~\ref{prop:rsbi_convergence}) for \M.

\textbf{Preconditioning the Phase-2 sampler for unimodal posteriors. }

The Phase~2 sampling process is isotropic, so on ill-conditioned posteriors (e.g strongly correlated
parameters, narrow ridges) no single noise scale $\sigma_t$ suits every direction. We therefore provide an option to run Phase~2 in
coordinates whitened by the mean $\bar{\theta}$ and covariance $\Sigma = LL^\top$ of the Phase~1 samples,
$z = L^{-1}(\theta - \bar{\theta})$, mapping only the energy gradient, $\nabla_z E_r = L^\top \nabla_\theta E_r$.
The affine map leaves the target unchanged and costs no simulations. It is the analogue of mass-matrix  adaptation in HMC~\citep{neal2011mcmc,hoffman2014nuts}, with the metric supplied by Phase~1. As one
covariance cannot describe several modes,  Phase-2 preconditioning is useful for unimodal targets. For more discussion see Appendix.~\ref{app:phase2-precondition}.

\section{Experiments}
\label{sec:experiments}

We benchmark \M against two established neural-SBI methods:
SNPE-C~\citep{DBLP:conf/icml/GreenbergNM19} and SNVI+fKL~\citep{DBLP:conf/iclr/GlocklerDM22} (see Appendix~\ref{app:baselines} for further details), on
Two~Moons~\citep{DBLP:conf/icml/GreenbergNM19},
SLCP~\citep{DBLP:conf/aistats/PapamakariosS019}, Bernoulli~GLM \cite{pmlr-v130-lueckmann21a} and Lotka--Volterra~\citep{pmlr-v130-lueckmann21a},  as implemented in the \texttt{sbibm}~\citep{pmlr-v130-lueckmann21a} benchmarking suite, 
as well as  a new differentiable gravitational-wave ringdown task (\texttt{gw\_ringdown}).
To make them differentiable, we re-implement the benchmarks from \texttt{sbibm} in PyTorch.
For full details on the re-implementation we refer to Appendix~\ref{app:task_details}. 
We make public our full code and checkpoints\footnote{\url{https://github.com/Chromeilion/rsbi_supplementary}}. \\
Posterior quality is measured by the classifier two-sample test
(C2ST,~\citealp{DBLP:conf/iclr/Lopez-PazO17}) against task-appropriate
reference posteriors, obtained via rejection sampling for Two~Moons, \texttt{sbibm}'s reference samplers for SLCP,
Bernoulli~GLM, resampling the \texttt{sbibm} default for Lotka--Volterra, and MCMC with \texttt{emcee}~\citep{2013PASP..125..306F}
for \texttt{gw\_ringdown} (see Appendix~\ref{app:references} for details). For each task we sweep several prior sizes (uniform boxes, or Gaussian prior scales for Bernoulli~GLM and Lotka--Volterra) and three simulation budgets
$N_\text{sim} \in \{10^3, 10^4, 10^5\}$, using the log prior-to-posterior volume ratio $\log_{10}\mathcal{R}_p$, averaged
over the observations, to quantify the volume reduction (see App.~\ref{sec:volume} for the definition and
App.~\ref{app:prior_volums} for the values adopted here). The smallest priors are the ones implemented in the \texttt{sbibm} package. All baselines use their released \texttt{sbibm} or SNVI defaults at matched total simulator-call budget, with the number of SNVI rounds set to 5 and SNPE rounds set to 2-3 (instead of 10) to match \M closer and reduce computational cost.
Lastly, the surrogate bandwidths were taken to be adaptive where possible. The Bernoulli-GLM experiment instead implemented the heuristic criteria of
Appendix~\ref{sec:energy_landscape_opt} to chose $\epsilon$. \texttt{gw\_ringdown} 
takes $\epsilon=1$.  Full task definitions, prior scenarios, hyperparameters, 
reference-sampler
configurations and the \M training schedule are in
Appendix~\ref{app:task_details}.    

\textbf{Compute-Matched Experiment:} Because doing the full forward + backward pass through a 
simulator is more expensive than only a forward pass, we also investigate 
\M's performance under a compute-matched budget. These results are shown in \apref{sec:compute_cost}. 

\subsection{Main results: gravitational-wave ringdown}
\label{sec:exp_gw}

\begin{wraptable}{R}{0.46\textwidth}
\vspace{-\intextsep}
\caption{Mean C2ST on \texttt{gw\_ringdown} against the \texttt{emcee}
reference at $N_\text{sim} = 10^5$, averaged over 10 simulated
observations. For every method the phases $\varphi_{220}$, $\varphi_{221}$ are
wrapped to $[0, 2\pi)$ before scoring.}
\label{tab:gw}
\centering
\small
\begin{tabular}{@{}lcc@{}}
\toprule
                      & \multicolumn{2}{c}{Prior (see Table~\ref{tab:prior_scenarios})} \\
\cmidrule(l){2-3}
Method                & default & wide \\
$\log_{10}\mathcal{R}_p$ & $\approx 2.6$ & $\approx 2.8$ \\
\midrule
SNVI+fKL              & $0.958\,(0.032)$ & $0.986\,(0.022)$ \\
SNPE                  & $0.935\,(0.071)$ & $0.985\,(0.015)$ \\
\M (ours)             & $\mathbf{0.891\,(0.070)}$ & $\mathbf{0.903\,(0.069)}$ \\
\bottomrule
\end{tabular}
\end{wraptable}

\begin{figure}[!htbp]
\centering
\includegraphics[width=\linewidth]{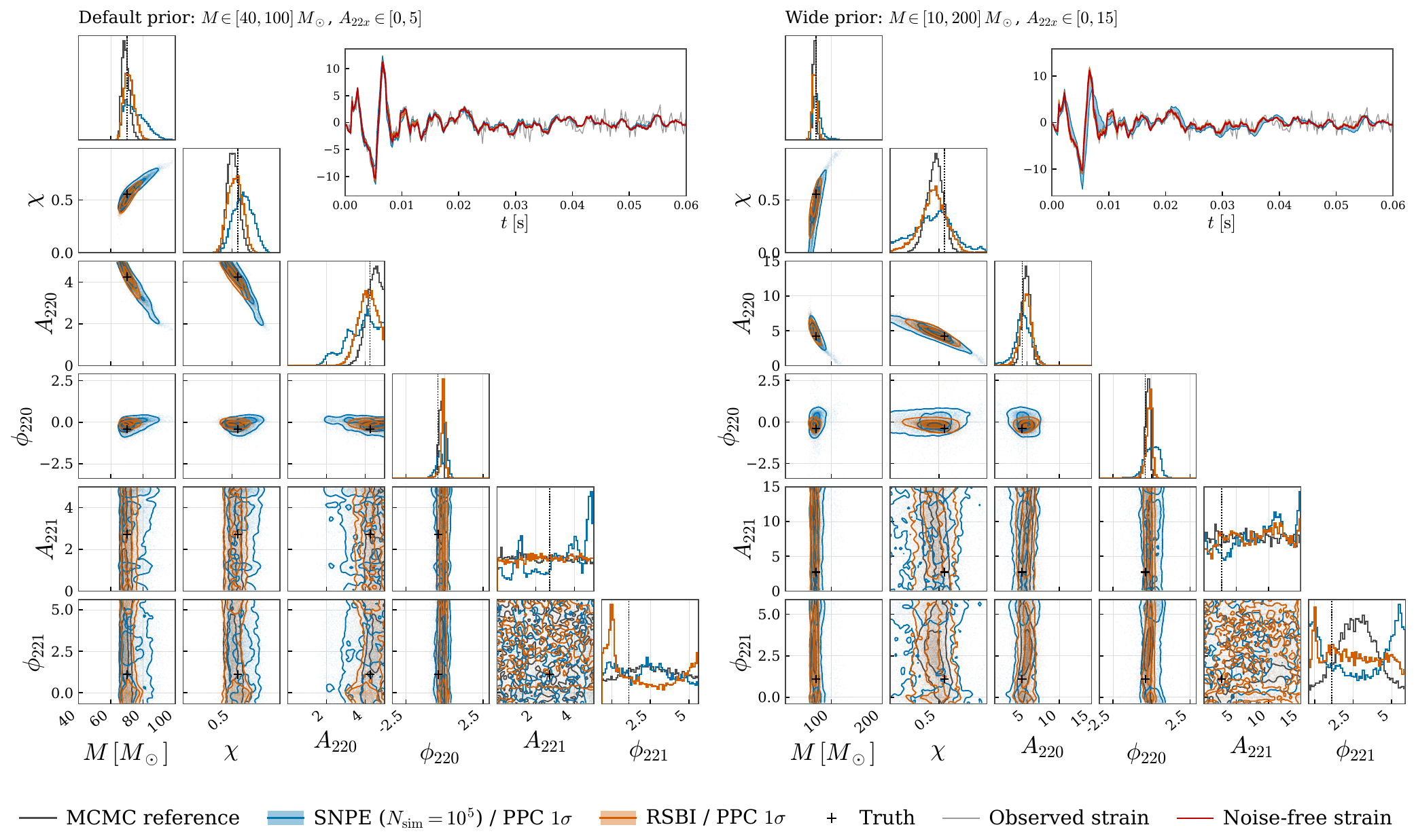}
\caption{6D \texttt{gw\_ringdown} posteriors of \M, SNPE and the MCMC reference
for observation~10 at $N_\text{sim}=10^5$, under the default prior (left,
$\log_{10}\mathcal{R}_p \approx 2.62$) and the wide prior (right,
$\approx 2.84$; Table~\ref{tab:prior_scenarios}). Insets: whitened observation
(grey), noise-free injection (red) and the $1\sigma$ posterior-predictive bands
of SNPE and \M from $200$ re-simulated posterior samples each.}
\label{fig:gw_corner}
\end{figure}

We first evaluate \M on a real-world inference task: estimating six
ringdown parameters $(M, \chi, A_{220}, \varphi_{220},
A_{221}, \varphi_{221})$ from a black hole post-merger gravitational wave signal,
with extrinsic dimensions fixed to truth (the natural single-detector
setup with known sky position and distance). Simulator construction
and reference posterior generation are described in
Appendix~\ref{app:gw_task}. Figure~\ref{fig:gw_corner} shows the $d_\theta = 6$ 2-d and 1-d marginal
posteriors on observation~10 at the default and wide priors
from $N_\text{sim} = 10^5$: \M tracks the \texttt{emcee} reference closely across all six parameters, with a more constrained posterior in comparison to that of SNPE, as also evident in their respective C2ST scores (see Table~\ref{tab:gw}) for the default prior (0.891 and 0.935, respectively) and wide prior (0.903 and 0.985, respectively).
The strain inset shows the $2\sigma$ posterior-predictive band of \M tightly envelopes the injected ringdown signal, in comparison to the larger uncertainties of SNPE.
We highlight the difference in the MCMC and RSBI posteriors, especially visible in the 1-dimensional marginal plot of $A_{220}$ (see top row of the $A_{220}$ corner column).

\subsection{Main results: robustness under \texorpdfstring{$\mathcal{R}_p$}{Rp} increase}
\label{sec:exp_main}

\begin{figure}[!htbp]
    \centering
    \includegraphics[width=\linewidth]{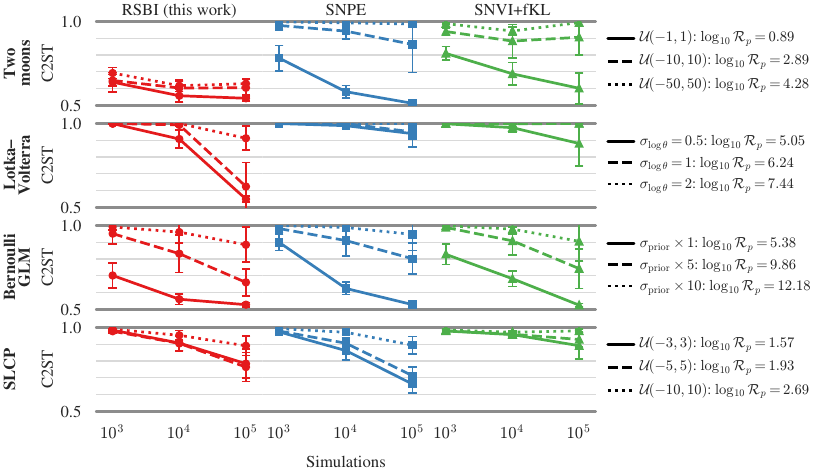}
 
    \caption{Robustness to prior-to-posterior volume ratio on 
    Two Moons, Lotka--Volterra, Bernoulli GLM and SLCP benchmarks. 
    Mean C2ST ($\pm1\sigma$) calculated against the reference 
    posterior vs.\ simulation budget $N_\text{sim}$ over 10 observations.
    Legends report $\log_{10}\mathcal{R}_p$, full line is the \texttt{sbibm} default. 
    C2ST $= 0.5$ matches the reference, $1.0$ is fully distinguishable.}
    \label{fig:c2st_main}
\end{figure}

Figure~\ref{fig:c2st_main} reports C2ST against the simulation budget ($N_\text{sim}$) on 
the Two Moons, Lotka-Volterra, Bernoulli GLM, and SLCP benchmarks~\citep{pmlr-v130-lueckmann21a}. 
On Two Moons and Bernoulli GLM, all three methods start to converge towards $0.5$ C2ST at $N_\text{sim} = 10^5$.
On the Lotka-Volterra and SLCP benchmarks this is not the case, with only \M converging on Lotka-Volterra.
We also find that the C2ST gap widens as $\mathcal{R}_p$ increases. 
On Two~Moons at $\log_{10}\mathcal{R}_p \approx 4.3$, 
the sequential sbi baselines saturate at C2ST $\approx 1$ for every budget, while 
\M maintains $\text{C2ST} \lesssim 0.7$ throughout. On SLCP, all baselines 
degrade by $0.1$--$0.2$ C2ST between the sbibm default and 
$\log_{10}\mathcal{R}_p \approx 2.69$ at $10^5$
simulations. With the smallest $\mathcal{R}_p$ difference between priors, this experiment details a regime for which the gains of RSBI are not extpected to be as pronounced. 
On Bernoulli GLM, all methods agree at the default prior, but at $\log_{10}\mathcal{R}_p \approx 9.9$
\M reaches a C2ST of $0.66$ at $10^5$ simulations, against $0.74$ for SNVI+fKL and $0.80$ for SNPE, and
at $\log_{10}\mathcal{R}_p \approx 12.2$ $0.89$ against $0.91$ and $0.95$. With Lotka-Volterra we find only \M is able to achieve strong results within the simulation budget,
reaching a C2ST of $0.55$ at $10^5$ simulations against $0.88$ for SNVI+fKL and $0.94$ for SNPE. On Lotka-Volterra we also implement the Phase 2 preconditioning of Sec.\ref{sec:stage2} which considerably enhanced performance. 

To investigate the convergence of RSBI under large simulation budgets, we also conduct 
an additional run on Two Moons, extending the number of sims 
to $10^{6}$ on the default prior. For this run we achieve a mean C2ST ($\pm1\sigma$) of 0.536 ($\pm 0.014$), a 
slight improvement over the $10^{5}$ run, which achieves 0.543 ($\pm 0.018$).

\section{Discussion}
\label{sec:discussion}

The \M methodology introduces two novel components that can each be adopted independently of ASBS. The
first is a surrogate proposal from a differentiable simulator. \eqref{eq:surrogate} is an
unnormalised log density that can be differentiated with respect to $\theta$ by backpropagating
through the simulator, so any gradient-based method can target it. For example, VI can be run as in
SNVI~\citep{DBLP:conf/iclr/GlocklerDM22} with the learned likelihood replaced by the surrogate
distance. Alternatively, ASBS can be swapped for annealed Langevin dynamics or another diffusion
sampler designed for multimodal targets, which avoids the sampler-specific hyperparameters that ASBS
requires. 

The second component is diffusion-based NRE refinement via transport. A ratio estimator remains valid under
any proposal, so every simulation is can be reused, and $\log p(\theta) + \log r_\psi(\theta,x_o)$ is
sampled starting from the Phase~1 proposal rather than from the prior or a truncated version of it.
 The proposal is not sufficient on its own. Our ablation (Appendix~\ref{app:tm-phase2-ablation})
shows that seeding SNPE and SNVI with the Phase~1 proposal improves them, but a substantial further
gain comes only when Phase~2 refines the Phase 1 proposal via NRE-ASBS  transport. Additionally, preconditioning Phase 2 by whitening the Phase 1 proposal (Sec.\ref{sec:stage2}) can alleviate issues arising from ill-conditioned  unimodal posteriors, potentially improving the performance of the ASBS.
On gw\_ringdown (Sec.~\ref{sec:exp_gw})
\M is the only method substantially below a C2ST of $1$, but the remaining discrepancy with the reference is concentrated in the periodic phases $\phi_{220},\phi_{221}$, which no method resolves well. The cost of backpropagating through the simulator is incurred only in Phase~1 (Appendix~\ref{sec:compute_cost}).

A further consequence of driving the proposal with a surrogate is that it can be engineered in ways
a learned estimator cannot. Since the surrogate only has to point towards the posterior, it can be
relaxed for discrete simulators, as with the straight-through Bernoulli relaxation
(Appendix~\ref{app:bernoulli-hparams}). The prior also enters both Phase 1 and Phase 2 energies exactly, mitigating posterior mass leakage outside the support (Appendix~\ref{app:leakage}), whereas
truncation-based methods must correct for this after the fact.

\paragraph{Limitations.}\label{sec:limitations}
\M requires a differentiable simulator. For black-box simulators, classical sequential SBI remains
the appropriate choice. A backward pass through the simulator costs $1.7$ to $4.0$ times a forward
pass (Appendix~\ref{sec:compute_cost}), although this applies only to the Phase~1 share of the budget.
Phase~2 costs more per iteration than amortised NPE, so \M pays off when simulation rather than
network training is the bottleneck. It is also not amortised across $x_o$, a new $x_o$ requires 
re-running both phases. Phase~1 adds three inference-level hyperparameters, the surrogate bandwidth 
$\epsilon$, the distance $\D$ and the temperature $T$ (Appendices~\ref{app:surrogate-ablations} and~\ref{app:bandwidth-ablations}). 
Finally, MCMC is exact in the limit of infinite samples, whereas \M, like other neural SBI methods, 
is bounded by network capacity and optimization. On two moons, for example, accuracy saturates 
even at $10^6$ simulations (Sec.~\ref{sec:exp_main}).

A further set of limitations stems from our choice of ASBS as the sampler rather than from \M itself.
ASBS is sensitive to its own hyperparameters, notably the reference noise $\sigma_{\max}$, the
temperature and the training schedule. We set these per benchmark, keeping them as agnostic as
possible to the surrogate bandwidth by using an adaptive variant or, for the ablations in
Appendix~\ref{app:surrogate}, by inspecting the gradient landscape. The guarantees of the \M target against prior leakage and
mode collapse only carry over to the samples if the ASBS sampler converges (Appendix~\ref{app:leakage}).
When the reference noise $\sigma_{\max}$ is comparable to the separation between modes, the
mode-seeking objective and the self-refilled buffer can amplify small imbalances until modes empty.
An unconverged controller can also scatter samples outside the support. These are failures of the
sampler's optimisation rather than of the target, but in practice they require $\sigma_{\max}$ to be
set relative to the posterior geometry. We stress that the novelty of \M, a surrogate-driven proposal
refined against a target that retains the prior and the simulator, holds for any choice of sampler,
and ASBS can be replaced by any method that requires only $\nabla_\theta E$.

\section{Conclusion}
\label{sec:conclusion}

We introduced \M, a sequential SBI method that uses gradients of a differentiable simulator to zoom
from a broad prior to the posterior before any estimator is trained, subsequently
refining the prediction by transporting the proposal via diffusion under an NRE 
target. Each \M component is compatible with
existing sequential pipelines and with any gradient-based sampler or VI. 
We showed \M is competitive across simulation budgets on broad priors where prior-initialized
methods spend most of their simulations far from $x_o$.
\paragraph{Future work.}
A systematic study of the gradient landscape induced by
differentiable simulators would show when this signal is reliable, and could yield curricula that
adapt automatically instead of being set per task. 
Furthermore, the Phase~2 preconditioning we utilize is a
single global affine whitening estimated from Phase~1 samples (Appendix~\ref{app:phase2-precondition}).
We will quantify how much preconditioning ASBS needs as the posterior becomes more anisotropic or
curved, and replace the global map with adaptive or nonlinear alternatives, such as running estimates
of the posterior covariance or learned transport maps.
 
Two high-impact scientific applications stand out. One is extreme-mass-ratio inspirals (EMRIs), where the broad-prior
regime is even more pronounced than for ringdown. The other is inference with neural emulators: any
trained emulator is a differentiable simulator usable within our framework, and it reduces the
per-call simulator cost to a network forward pass. The \M target is
independent of the sampler, so in future work we will pair it with other samplers such as Langevin
dynamics, flow-based proposals, or ASBS with geometry-adapted reference processes.
\newpage
\section{Acknowledgments}
The authors thank the International Max Planck Research School for Intelligent Systems (IMPRS-IS) for supporting Uros Zivanovic. This work was supported by the Max Planck Artificial Intelligence Network (MP-AIX). RT and AS acknowledge funding from Next Generation EU, in the context of the National Recovery and Resilience Plan, Investment PE1 Project FAIR ``Future Artificial Intelligence Research''. This resource was co-financed by the Next Generation EU [DM 1555 del 11.10.22]. RT is partially supported by the Fondazione ICSC, Spoke 3 ``Astrophysics and Cosmos Observations'', Piano Nazionale di Ripresa e Resilienza Project ID CN00000013 ``Italian Research Center on High-Performance Computing, Big Data and Quantum Computing'' funded by MUR Missione 4 Componente 2 Investimento 1.4: Potenziamento strutture di ricerca e creazione di ``campioni nazionali di R\&S (M4C2-19)'' - Next Generation EU (NGEU).

\bibliographystyle{unsrtnat}
\bibliography{neurips_2026.bib}

\newpage
\appendix

\addtocontents{toc}{\protect\setcounter{tocdepth}{2}}
\renewcommand{\contentsname}{Appendix contents}
\tableofcontents

\section{Additional methodological details}
\label{app:additional_details}
\subsection{Countering Mode-Seeking}
\label{sec:well_tempering}
RSBI is variational, so training the controlled SDE against $E_r$
amounts to minimising a reverse KL between the controlled-SDE marginal
and the round-$r$ target. Standard reverse-KL VI is
mode-seeking~\citep{DBLP:conf/iclr/GlocklerDM22}: when the target has
multiple disconnected modes, the optimal proposal concentrates on
whichever mode is cheapest to cover and abandons the rest. ASBS
partially counters this at no extra cost: its replay buffer maintains
trajectories from earlier in training, preventing the controller from
collapsing onto a single mode within an ASBS round, and its small base
diffusion coefficient $\sigma$ preserves separation between modes that
have already been discovered. Neither mechanism, however, drives the
controller toward modes it has not yet visited; on multimodal posteriors
with high inter-mode barriers, the ASBS optimum can still be a strict
subset of the true target's support.

We address this directly with the well-tempered ASBS extension
of~\citet{nam2026enhancing}: a low-dimensional collective variable
$\xi : \Theta \to \mathcal{S} \subseteq \mathbb{R}^m$ ($m \leq d_\theta$,
chosen per-task) is introduced, and a bias $V_k(\xi(\theta))$ is
deposited online during training. The bias accumulates in regions
of CV space the sampler has already visited and pushes future samples
toward novel ones. With kernel height $h_{\text{wt}}$, kernel width $\sigma_\mathrm{cv}$,
and bias factor $\gamma > 1$, the update rule is
\begin{equation}\label{eq:wt_update_rule}
    V_{k+1}(s)
    \;=\;
    V_k(s) \;+\; h_{\text{wt}} \sum_{i=1}^N
    \exp\!\left(-\,\frac{\beta}{\gamma - 1}\, V_k\!\bigl(\xi(\theta^{(i)})\bigr)\right)
    K_{\sigma_\mathrm{cv}}\!\left(s,\, \xi(\theta^{(i)})\right),
\end{equation}
where $K_{\sigma_\mathrm{cv}}$ is a Gaussian kernel and the
$\theta^{(i)}$ are the i.i.d.\ samples produced by the current
controlled SDE. The well-tempering factor
$\exp(-\beta V_k / (\gamma{-}1))$ shrinks the deposition rate as bias
accumulates, so $V_k$ converges to a finite limit rather than diverging.
The Adjoint Matching objective \eqref{eq:am_obj} is augmented with the
bias gradient through the chain rule:
\begin{equation}\label{eq:wt_am_objective}
    u^{(k)} = \argmin_u\,
    \mathbb{E}_{p_{t|0,1}^\text{base}\, p_{0,1}^{\bar u}}
    \left[\bigl\|u_t(X_t) + \sigma_t\bigl(\nabla E(X_1) + \nabla(V_k\!\circ\!\xi)(X_1) + h^{(k-1)}(X_1)\bigr)\bigr\|^2\right].
\end{equation}
After training, the bias is removed via importance reweighting by
$w(\theta) \propto \exp(\beta V^\ast(\xi(\theta)))$, recovering weighted
samples from the original Phase-2 target.

\begin{proposition}[Convergence of \M]
\label{prop:rsbi_convergence}
Within each Phase-2 round, under the regularity conditions of
Proposition~3.1 of~\citet{nam2026enhancing}, compact CV domain,
positive-definite Gaussian kernel, and Robbins-Monro step
schedule, the bias $V_k$ in \eqref{eq:wt_update_rule} converges almost
surely to $V^\ast(s) = -(1 - 1/\gamma)\,F_r(s) + \mathrm{const}$, where
$F_r(s)$ is the round-$r$ PMF along $\xi$. Combined with
Theorem~\ref{th:asbs_convergence} and the consistency of the NRE-C
estimator $r_{\psi_r} \to p(x \mid \theta)/q_r(x)$ (Appendix~\ref{app:nre_loss}), when
Algorithm~\ref{alg:sgsbi} is run to convergence the round-$r$ marginal
of the controlled SDE, reweighted by $w$, converges to the target
posterior $p(\theta \mid x_o)$.
\end{proposition}

Algorithm~\ref{alg:sgsbi} details the full \M procedure. The structure
mirrors that of \citet{DBLP:conf/iclr/GlocklerDM22}: a proposal is
refined across rounds, but the target $p(\theta\mid x_o)$ is held fixed
throughout and reweighting is analytic rather than learned.

\subsection{Training the ratio estimator}
\label{app:nre_loss}
Phase~2's energy $E_r$ requires a ratio estimator $r_{\psi_r}(\theta, x_o)$
that tracks the log-ratio
$\log p(x_o \mid \theta) / q_r(x_o)$, whose dependence on $\theta$ is proposal-invariant. We train $r_\psi$ with the contrastive
NRE-C loss of~\citet{DBLP:conf/nips/MillerWF22}, which generalises NRE-A
\citep{DBLP:conf/icml/HermansBL20} and NRE-B
\citep{DBLP:conf/icml/Durkan0P20}. Given a minibatch of $B$ paired samples
$(\theta_b, x_b)_{b=1}^B$ with $\theta_b \sim q_r(\theta)$ and
$x_b = s(\theta_b, \omega_b)$ from the round-$r$ training set $\mathcal{B}$ of
Algorithm~\ref{alg:sgsbi}, where $q_r$ is the mixture of all proposals so far, plus
in-batch contrastive parameters $\theta_k \sim q_r$ drawn for each anchor from the other pairs,
the NRE-C objective with $K_{\text{class}}$ classes is
\begin{equation}
\label{eq:nrec}
\begin{split}
    &\mathcal{L}_\mathrm{NRE\text{-}C}(\psi)
    \;=\;
    -\frac{1}{1+\eta}\,
       \mathbb{E}_{q_r(x)\, q_r^{\otimes K_{\text{class}}}(\theta_{1:K_{\text{class}}})}
    \left[
        \log\frac{K_{\text{class}}}{K_{\text{class}} + \eta\sum_{k=1}^{K_{\text{class}}} r_\psi(\theta_k, x)}
    \right] \\
    &\qquad\quad -\frac{\eta}{1+\eta}\,
       \mathbb{E}_{q_r(\theta_0)\, p(x \mid \theta_0)\, q_r^{\otimes (K_{\text{class}}-1)}(\theta_{1:K_{\text{class}}-1})}
    \left[
        \log\frac{\eta\, r_\psi(\theta_0, x)}
                 {K_{\text{class}} + \eta\sum_{k=0}^{K_{\text{class}}-1} r_\psi(\theta_k, x)}
    \right],
\end{split}
\end{equation}
where $\eta$ weights the joint against the marginal term. We use $\eta = 1$
throughout (NRE-B is recovered as $\eta \to \infty$, and NRE-A at $K_{\text{class}} = 1$,
$\eta = 1$). The minimiser is the ratio
$r_\psi^\ast(\theta, x) = p(x \mid \theta) / q_r(x)$, whose gradient in $\theta$ does not depend on the
proposal $q_r$ used to generate training data, the property our Phase-2
derivation in Section~\ref{sec:stage2} explicitly relies on, and the
property that distinguishes the NRE-C target from the
proposal-conditioned target of SNPE-C/APT~\citep{DBLP:conf/icml/GreenbergNM19}.

\subsection{Derivation of the Surrogate Energy}
\label{app:surrogate:derivation}

Starting from the surrogate posterior
$\tilde{p}_\epsilon(\theta\mid x_o)$ specified through the surrogate
likelihood~\eqref{eq:surrogate}:
\begin{equation}\label{eq:surrogate_posterior}
    \begin{split}
        \log \tilde{p}_\epsilon(\theta\mid x_o)
        = \;-\,\,\mathbb{E}\,\D(s(\theta), x_o) + \log p(\theta).
    \end{split}
\end{equation}
This has the same exponential structure as the Boltzmann
distribution~\eqref{eq:target_dist}, with the energy occupying the
entire exponent. Sign-reversing and differentiating gives the
surrogate energy gradient under a sufficient distance measure $\D$:
\begin{equation*}
    \nabla_\theta E_s(\theta) \;=\;
    \,\mathbb{E}\,\nabla_\theta \D(s(\theta), x_o)
    \;-\; \nabla_\theta \log p(\theta).
\end{equation*}

For $\D$ to define a proper probability density $\tilde{p}_\epsilon(\theta \mid x_o)$,
the requirement is that of a loss in a general Bayesian update~\citep{bissiri2016general}:
\begin{definition}[Sufficient distance measure]\label{def:sufficient_measure}
    A function $\D:\mathcal{X}^K\times\mathcal{X}\to\mathbb{R}$ is a
    \emph{sufficient distance measure} for a prior $p(\theta)$ and observation
    $x_o$ if it is differentiable in its first argument and the surrogate
    posterior has a finite, non-zero normalising constant:
    \[
        0 \;<\; \int_\Theta
            \exp\!\Big(-\,\mathbb{E}\,\D\bigl(s(\theta), x_o\bigr)\Big)\,
            p(\theta)\, d\theta \;<\; \infty .
    \]
\end{definition}
Since by Theorem~\ref{th:equivalence}, any differentiable $\tilde{\D}$ that
equals $\D$ up to an additive constant satisfies $\nabla\tilde{\D} =
\nabla\D$, as far as ASBS is concerned \eqref{eq:surrogate_energy} we can replace
$\D$ with $\tilde{\D}$, equivalent under differentiation.
In practice this relaxation matters because it admits standard losses whose constants are
conventionally dropped, such as the $\tfrac{d_x}{2}\log 2\pi$ of the Gaussian NLL.

\subsection{Optional truncation of the ratio estimator}
\label{app:nre_truncation}

A ratio estimator is only identified where its training data have support. Trained on $(\theta_i, x_i)$ with
$\theta_i \sim q_r(\theta)$, the classifier $r_\psi(\theta, x_o)$ is constrained near the bulk of $q_r$, but in regions
the proposal never visited its value is an extrapolation of the network. Because the Phase~2 sampler minimises
$E_r = -\log r_\psi - \log p(\theta)$ over the whole prior, a spuriously large extrapolated ratio can pull samples into
regions with no posterior mass. We optionally allow RSBI to guard against this by restricting where the ratio is trusted. This may be the case, for example, when the Phase~1 zoom is assumed to be locally concentrated around the true posterior mass. The choice firmly remains with the user.

Let $\{\theta_j\}$ be the parameters the region is fitted on (the Phase~1 sampler draws, or
all NRE training parameters of the round). We compute their mean $\bar{\theta}$ and covariance $\Sigma$, discard the points whose
Mahalanobis distance exceeds the $q$-quantile of the distances, and recompute $\bar{\theta}$ and $\Sigma$ on the remaining core, so that
outlying draws do not inflate the region. The distance of a parameter to the core is
$d(\theta) = \sqrt{(\theta - \bar{\theta})^\top \Sigma^{-1} (\theta - \bar{\theta})}$. For multimodal posteriors a single ellipsoid around
separated modes would cover most of the prior, so we first split $\{\theta_j\}$ into $J$ clusters with k-means, fit
$(\bar{\theta}_c, \Sigma_c)$ per cluster as above and set $d(\theta) = \min_c d_c(\theta)$. The truncated ratio is
\begin{equation}\label{eq:nre_truncation}
  \log \bar r_\psi(\theta, x_o) = \log r_\psi(\theta, x_o) - \lambda\,\mathrm{relu}\bigl(d(\theta) - k\bigr)^2 ,
\end{equation}
which leaves the ratio unchanged where $d(\theta) \le k$ and adds a soft quadratic wall of strength $\lambda$ beyond.
Phase~2 then uses $E_r = -\log \bar r_\psi - \log p(\theta)$. The wall is refitted after every NRE update, acts only on
the ratio, leaves the prior term untouched and needs no simulations beyond those already in the budget.

Restricting inference to a region that provably contains the
posterior is a common device in sequential SBI. Truncated sequential NPE (TSNPE) \citep{deistler2022truncated} draws
each round's simulations from the prior restricted to the highest-probability region of the current posterior
estimate. Truncated marginal NRE (TMNRE) \citep{miller2021truncated} shrinks the prior to a box around the region
where the ratio estimate is non-negligible and retrains within it.  All of these
methods, and ours, rest on the same observation: a learned density or ratio is only reliable where it has data, so
inference is confined to that region rather than trusted everywhere. Our variant differs in three respects. (i) The
region is fitted on samples that are already available (the Phase~1 output and the round's training parameters), so it
costs no additional simulations. (ii) The truncation is soft and acts on the energy rather than on the prior or the proposal,
so the sampler is steered back into the region rather than forbidden from leaving it, and the prior in $E_r$ stays the
original one. (iii) The region is an ellipsoid (or a union of ellipsoids), not an axis-aligned box, and so follows
correlated posteriors.

 Inside $\{d(\theta) \le k\}$ the truncated energy equals the untruncated one, so the
Phase~2 target is unchanged wherever the region covers the posterior. As for TSNPE and TMNRE, the truncation biases the
result only if non-negligible posterior mass lies outside the region, which $k$ and the trim quantile $q$ control. A
larger $k$ is safer but restricts less, and a smaller $q$ fits a tighter core. The wall is a safeguard against
extrapolation and is not needed when the ratio is well behaved, so it is optional and off unless stated otherwise.

 Among the benchmark tasks the truncation is used only for Lotka--Volterra
(App.~\ref{sec:lokta-volterra}), where it is enabled for all prior widths and simulation budgets. Its Phase~1
posterior reliably covers the reference posterior there, while the oscillatory likelihood makes the ratio
extrapolate poorly away from the training parameters. We use a single cluster ($J = 1$) fitted on the NRE training
parameters of the current round and refitted every round, with $\lambda = 50$, $k = 3$ and $q = 0.99$. The only
exception is the large prior at $N_\text{sim} = 10^5$, where $k = 2$ and $q = 0.9$ give a tighter core. For
Lotka--Volterra, $\log r_\psi$ is in addition softly capped at the $0.999$-quantile of its values over the training
parameters plus a margin of $2$, which guards against the same far-field blow-up of the classifier logits. Two~Moons,
SLCP, Bernoulli~GLM and \texttt{gw\_ringdown} use the untruncated ratio.

\subsection{Preconditioning the Phase~2 sampler}
\label{app:phase2-precondition}

The Phase~2 sampler is an adjoint Schr\"odinger bridge sampler (ASBS;~\citealp{liu2026adjoint}) whose
reference process is isotropic with the same noise scale $\sigma_t$ is applied in every direction of
$\theta$-space. This choice is natural when the target has a comparable width in all directions, but it
is poorly suited to \emph{ill-conditioned} targets, whose posterior is very narrow along some directions
and broad along others. Such geometry arises whenever parameters are strongly correlated, when the
posterior concentrates on a curved ridge, or when the posterior widths of different parameters differ by
orders of magnitude. A noise scale small enough to resolve the stiff directions moves far too slowly along
the shallow ones, while a noise scale matched to the shallow directions overshoots the stiff ones; in
either case the sampler concentrates poorly within a fixed training budget. The same difficulty is
well known for gradient-based MCMC: the efficiency of the Metropolis-adjusted and unadjusted Langevin
algorithms degrades with the condition number of the target~\citep{roberts2001optimal,dalalyan2017theoretical},
and Hamiltonian Monte Carlo with an identity mass matrix must choose a step size limited by the
narrowest direction~\citep{neal2011mcmc}. The standard remedy is to precondition the dynamics with a
metric that approximates the posterior covariance, either fixed~\citep{neal2011mcmc}, adapted from
earlier samples~\citep{haario2001adaptive,hoffman2014nuts,carpenter2017stan}, position-dependent~\citep{girolami2011riemann},
or given by a learned transport map~\citep{hoffman2019neutra}. For diffusion samplers the difficulty lies in learning rather than in sampling: on Gaussian targets the
sampling error of a diffusion sampler with optimised hyperparameters does not depend on the condition
number~\citep{perbost2026diffusion}, but regression-trained drifts fit high-variance directions quickly
and low-variance ones slowly when the target is ill-conditioned, which preconditioning into an isotropic
space removes~\citep{ahamed2026preconditioned}.

In our method, Phase~1 already provides the information needed for such a metric, at no extra
simulation cost. Its samples are concentrated around the posterior: we use the round-0 NRE training
parameters $\{\theta_j\}$ for Lotka--Volterra, and $10^4$ fresh draws from the trained Phase-1 sampler
for \texttt{gw\_ringdown}. Their mean $\bar{\theta}$ and covariance $\Sigma = LL^\top$, with $L$ the Cholesky factor,
are estimated from the samples whose Mahalanobis distance to the sample mean lies below its $q$-quantile,
where $q$ is the \emph{whitening quantile}. We use $q = 0.99$, so that a few stray Phase-1 samples do
not inflate the covariance (at $10^{5}$ simulation budget, $q = 0.9$ for the large Lotka--Volterra prior, 
whose Phase-1 proposal has heavier tails).
 Together, $\bar{\theta}$ and $\Sigma$ describe the location, scale and orientation of the posterior. We
then run Phase~2 in the whitened coordinates
\begin{equation}
  z = L^{-1}(\theta - \bar{\theta}), \qquad \theta = \bar{\theta} + L z ,
\end{equation}
in which the Phase-1 sample cloud has approximately zero mean and unit covariance. The energy is still
evaluated in the original coordinates, and only its gradient is transformed:
\begin{equation}
  E_r(z) = E_r(\bar{\theta} + L z), \qquad \nabla_z E_r = L^\top \nabla_\theta E_r ,
\end{equation}
where $E_r(\theta) = -\log p(\theta) - \log r_{\psi_r}(\theta, x_o)$ is the round-$r$ energy
of~\eqref{eq:phase_two_energy}.

Because the map is an affine change of variables with constant Jacobian, it leaves the target posterior
unchanged: samples drawn in $z$ are mapped back by $\theta = \bar{\theta} + L z$, and all simulations, NRE training
and C2ST evaluation take place in $\theta$. The transform requires no additional simulator calls, since it
is computed only from Phase-1 samples. It is fixed after round~0, so the Phase~2 networks, which are
warm-started from one round to the next, remain valid across rounds. The construction is the analogue
of mass-matrix adaptation in HMC~\citep{neal2011mcmc,hoffman2014nuts,carpenter2017stan}, and of the
affine invariance of ensemble samplers~\citep{goodman2010ensemble}, with the metric supplied by Phase~1
rather than by a separate warm-up or tuning run. One consequence is that the reference noise scale
$\sigma_t$ is specified in $z$-units, that is, relative to the spread of the Phase~1 proposal rather than
in the units of $\theta$.

Preconditioning is most useful for unimodal posteriors with an ill-conditioned energy landscape, such as
strongly correlated parameters, long narrow ridges, or widths that differ by orders of magnitude between
directions. For Lotka--Volterra, for instance, the posterior is about $40\times$ narrower across its ridge
than along it. There, a single affine map removes most of the anisotropy, after which the isotropic
reference process is appropriate. For multimodal posteriors a global affine map is of little use, the mean
and covariance of a multimodal Phase~1 proposal span all modes and match the local geometry of none of
them.
Handling such targets would require a position-dependent metric~\citep{girolami2011riemann} or a
nonlinear transport map~\citep{hoffman2019neutra}; here we leave the sampler unpreconditioned. Developing such geometry-adapted
reference processes for diffusion samplers, including learned transport maps and position-dependent metrics
built from the Jacobian of the differentiable simulator, is the subject of our future work.
 For posteriors that are already
well conditioned, preconditioning reduces to a near-identity rescaling and has little effect. In our
experiments we therefore precondition Phase~2 for Lotka--Volterra and \texttt{gw\_ringdown}, and not for
the multimodal Two Moons and SLCP posteriors.

Important to note is that preconditioning does not give RSBI an advantage over the baselines, since SNPE and SNVI represent the posterior with
normalising flows, whose learned
bijections adapt to the posterior's scale and orientation by
construction. Preconditioning simply gives the Phase~2 sampler the same affine information, estimated from RSBI's
Phase~1 samples without additional simulator calls.

\section{Additional Experiments}

\subsection{Compute-Matched Performance}
\label{sec:compute_cost}

\begin{figure}[htbp]
\centering
\includegraphics[width=\linewidth]{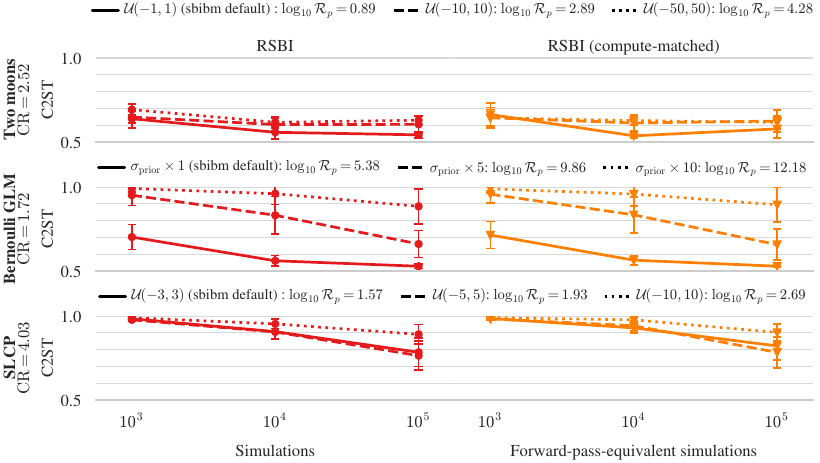}
\caption{\textbf{Left:} C2ST of \M against number of simulations. \textbf{Right:} C2ST of \M against the equivalent compute-matched simulation budget (reported in \tabref{tab:compute_matched_sims}). Calculated for Two Moons, Bernoulli GLM and SLCP benchmarks. Y-axis labels include the compute ratio $\mathrm{CR}$ charged per Phase-1 simulation (Table~\ref{tab:simulator_runtime}). Lines are the prior scenarios of Figure~\ref{fig:c2st_main} with error bars being the standard deviation over 10 observations.}
\label{fig:compute_matched}
\end{figure}

\begin{table}[b]
\centering
\caption{Effect of backwards pass on per-batch mean compute time across 3 \texttt{sbibm} benchmarks.}
\label{tab:simulator_runtime}
\begin{tabular}{lccc}
\toprule
\textbf{Simulator} & \textbf{Forward (ms)} & \textbf{Forward+Backward (ms)} & \textbf{Ratio} \\
\midrule
Two Moons & 0.1401 & 0.3528 & 2.52\,× \\
SLCP & 0.2113 & 0.8524 & 4.03\,× \\
Bernoulli GLM & 0.8057 & 1.3854 & 1.72\,× \\
\bottomrule
\end{tabular}
\end{table}

For all three simulators we time 20 forward passes and 20 forward + backward passes using a batch size of 32 and 5 warmup batches. 
Gradients are calculated with regards to the sum of the outputs of the simulator (e.g., \texttt{x.sum()}). 
The timings are done on an NVIDIA GTX 1660 Ti laptop GPU. 
The per-batch timings are reported in \tabref{tab:simulator_runtime}, C2ST under a compute-matched budget is reported 
in \figref{fig:compute_matched}, and the number of simulator calls used is reported in \tabref{tab:compute_matched_sims}.
Across all simulators we see an increased cost when doing the additional backward pass, ranging from $\sim1.7\times$ to $\sim4.0\times$.
We note that although the computational costs of doing the backward pass are significant, \M only requires this extra compute during Phase~1, which can take up only a small portion of the total simulation budget. 
We find across all benchmarks that \M maintains a strong performance across all simulation budgets even when compute-constrained.
Especially in Two Moons and Bernoulli GLM, \M maintains it's strong C2ST scores.
We stress that the measured timings in \tabref{tab:simulator_runtime} are hardware-dependent.

\begin{table}[htbp]
\caption{Total simulator calls made by the compute-matched runs of Figure~\ref{fig:compute_matched}.}
\label{tab:compute_matched_sims}
\centering\small
\begin{tabular}{@{}llccc@{}}\toprule
Task & Prior & $N_\text{sim} = 10^3$ & $N_\text{sim} = 10^4$ & $N_\text{sim} = 10^5$ \\\midrule
\multirow{3}{*}{Two Moons} & default & 724 & 5\,064 & 70\,037 \\
 & wide10x & 800 & 9\,203 & 99\,203 \\
 & wide50x & 800 & 9\,203 & 99\,203 \\
\midrule
\multirow{3}{*}{Bernoulli GLM} & default & 820 & 8\,200 & 83\,455 \\
 & medium & 820 & 8\,200 & 82\,015 \\
 & large & 820 & 8\,200 & 82\,015 \\
\midrule
\multirow{3}{*}{SLCP} & default & 624 & 6\,364 & 59\,859 \\
 & wide5x & 624 & 6\,364 & 59\,859 \\
 & wide10x & 624 & 6\,364 & 59\,859 \\
\bottomrule\end{tabular}
\end{table}

\subsection{Posterior mass leakage mitigation}
\label{app:leakage}

\begin{figure}[!htbp]
\centering
\includegraphics[width=\linewidth]{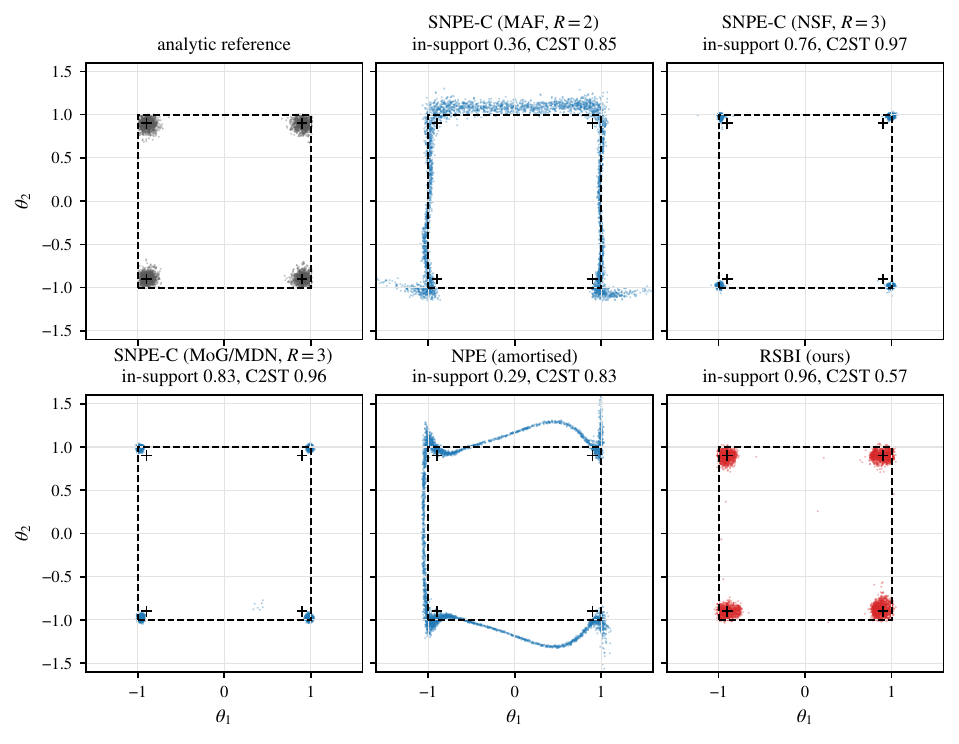}
\caption{Demonstration of posterior leakage and it's mitigation with RSBI: Posterior samples on the tractable squared-location task ($d = 2$, $N_\mathrm{sim} = 10^4$, one
observation). Dashed square: the support of the uniform prior, $[-b, b]^2$; crosses: the four true modes at
$(\pm c, \pm c)$, which lie close to the corners of the support. Each panel shows the raw sampler output of one
method (before rejection for the flows, before removing the metadynamics bias for RSBI), with the fraction of
its mass inside the support. The C2ST against the analytic reference ($0.5$ = indistinguishable) is computed on
each method's final posterior restricted to the support: the in-support samples of the flows and NPE, and the
in-support reweighted samples of RSBI. The atomic SNPE-C flows and amortised NPE place much of their mass outside
the support; RSBI keeps it inside while covering all four modes, like the analytic reference (top left).}
\label{fig:leakage-samples}
\end{figure}

Leakage arises when an algorithm learns a density over $\theta$ that is not tied to the prior support. Atomic
SNPE-C fits a conditional density estimator $q_\psi(\theta \mid x)$, an unbounded flow or Gaussian mixture on
$\mathbb{R}^d$, to simulated pairs, and its proposal correction normalises the estimator only over the finite
set of contrasting atoms. Nothing in the loss evaluates the estimator outside the support, so when the posterior
modes sit at the boundary, the smooth estimator spills mass across it and the leaked mass must be removed
afterwards by rejection. Truncated SNPE (TSNPE;~\citealp{deistler2022truncated}) avoids the atomic correction by drawing simulations
from the prior truncated to a high-probability region and training with the standard NPE loss, which removes
the optimisation pathology behind leakage; its estimator is nevertheless still a free density on
$\mathbb{R}^d$, so any residual out-of-support mass is again removed by rejection. RSBI never learns such a density. Its Phase~2 sampler is trained against the energy
$E_r(\theta) = -\log p(\theta) - \log r_{\psi_r}(\theta, x_o)$, which contains the prior exactly. Outside the
support $\log p(\theta) = -\infty$, so the target assigns zero mass there whatever the ratio estimator outputs:
an unconstrained classifier can be wrong about $r_{\psi_r}$ off the support, but the product $p(\theta)\,
r_{\psi_r}(\theta, x_o)$ cannot be. Because the sampler is fitted to this target, mass placed outside the support
is penalised by the energy, and its gradient pushes trajectories back inside. In practice we implement the box
prior as a stiff quadratic wall rather than an infinite one, to keep gradients finite, so a small residual
fraction of the raw samples falls just outside the boundary (Figure~\ref{fig:leakage-samples}). This fraction is
set by the wall stiffness and the sampler's training error rather than by the estimator, and it vanishes from the
final posterior, which is restricted to the support. The same argument applies to SNVI~\citep{DBLP:conf/iclr/GlocklerDM22},
whose variational posterior is likewise fitted to the prior-weighted ratio $p(\theta)\, r_\psi(\theta, x_o)$
rather than learned as a free density, and whose final importance-resampling step assigns zero weight to
samples outside the support. This guarantee holds only to the extent that the ASBS sampler fits its target. When the controller fails to
converge, for instance by collapsing onto a subset of modes or scattering trajectories across the box, its samples
need not respect the energy, and out-of-support mass can reappear. This may not be in general true for other samplers/models used in 
Phase 1 of RSBI, which we discuss as a limitation of choosing WT-ASBS in Section~\ref{sec:limitations}.

To isolate how bounded prior support affects inference (leakage), we use a tractable benchmark with a
closed-form posterior. The parameters are drawn from a uniform box, $\theta \sim \mathcal{U}([-b, b]^d)$, and the
data are $x = \theta^2 + \sigma\epsilon$, $\epsilon \sim \mathcal{N}(0, I)$, so the posterior for
$x_o = (c^2, \ldots, c^2)$ has $2^d$ symmetric modes at $(\pm c, \ldots, \pm c)$. Taking $c$ close to $b$ places
every mode near a corner of the prior support (in $d = 2$, a square), the regime in which density-based
posteriors are known to ``leak'' probability mass outside the support. Because the reference posterior is
available analytically, we can measure leakage exactly. We use $d = 2$, $b = 1$, $c = 0.9$ and $\sigma = 0.1$.

At a fixed simulation budget ($N_{\text{sim}} = 10^4$) we compare atomic SNPE-C~\citep{DBLP:conf/icml/GreenbergNM19},
using the standard \texttt{sbi} implementation~\citep{tejero2020sbi}, with masked autoregressive
flow~\citep{papamakarios2017masked} and neural spline flow~\citep{durkan2019nsf} density estimators, the
non-atomic mixture-of-Gaussians variant (whose posterior estimator is a Gaussian mixture), and amortised NPE
(a single round with a masked autoregressive flow), against RSBI. We report, on the raw sampler output, the
fraction of mass inside the prior support and the fraction of the $2^d$ modes recovered (mode recall), and the
classifier two-sample test (C2ST) against the analytic reference on each method's final posterior restricted
to the support.

For RSBI, Phase~1 uses the Gaussian negative log-likelihood surrogate with the exact noise variance
$\sigma^2$ and reference noise scale $\sigma_{\max} = 0.3$, and is stopped after $5 \times 10^3$ simulations.
Phase~2 spends the remaining $5 \times 10^3$ simulations in a single round: a contrastive NRE is trained on
fresh draws from the Phase~1 sampler, its ratio is evaluated only within the truncation region of
Appendix~\ref{app:nre_truncation}, and the Phase~2 sampler, again with $\sigma_{\max} = 0.3$, targets the posterior
under the uniform box prior.

The flow-based amortised and sequential methods place a substantial fraction of their mass outside the prior
support, and the mixture-of-Gaussians variant mitigates but does not eliminate this
(Figure~\ref{fig:leakage-samples}). In contrast, RSBI samples under a constrained energy and keeps almost all
of its mass inside the support while recovering all four modes. This controlled experiment shows that the
leakage pathology is structural to unconstrained density estimation on boundary-mode targets, and that the
constrained Phase~1 and 2 samplers remove it by construction.

\subsection{Ablation: replacing Phase~2 by standard sequential methods}
\label{app:tm-phase2-ablation}

\M has two components that could each explain its advantage on wide priors. Phase~1 localises the posterior by
training ASBS on a surrogate, so it zooms from the prior onto the region that carries posterior mass. Phase~2 then
refines this estimate with an NRE and an ASBS sampler that transports the Phase~1 sampler locally. To separate the
two, we keep Phase~1 fixed and replace Phase~2 with established sequential methods that use Phase~1 as their first
proposal, at the same simulation budget. The obvious candidate, atomic SNPE-C~\citep{DBLP:conf/icml/GreenbergNM19}, is
poorly suited to this setting. Its contrastive loss normalises the posterior only over a few atoms drawn from the
proposal, so posterior mass leaks into the large region of the prior that the localised proposal never visits. With
the Phase~1 sampler as its first proposal it reaches a C2ST of only $0.96$ and $0.93$ at $10^3$ and $10^4$
simulations, far from \M. We therefore use truncated SNPE (TSNPE)~\citep{deistler2022truncated}, which restricts
the prior to a highest-density region (HPR) of the current posterior estimate and trains with the maximum-likelihood
loss. It avoids leakage by construction and is the natural way to hand a localised proposal to a neural posterior
estimator. As a second, variational baseline we use SNVI~\citep{DBLP:conf/iclr/GlocklerDM22}, which, like Phase~2 of
\M, fits a sampler to the prior times a learned ratio and draws the next round of simulations from it. Unlike \M,
it fits a normalising flow over the whole prior by forward KL, so the flow is free to move mass anywhere in the prior box, including regions where the ratio estimator has never seen data. On a wide prior the classifier extrapolates there without control. In SNVI's self-normalised importance weights a single proposal in such a region can dominate all others, so the flow is pulled onto spurious high-ratio regions even when initialised at the Phase~1 sampler, and the next round of simulations is spent there. RSBI's Phase-2 instead, transports particles locally from the Phase-1 proposal due to ASBS using Phase 1's $\tilde{p}_\epsilon\left(\theta \mid x_o\right)$ as the source distribution.

We use Two Moons with the widest prior, $[-50, 50]^2$ ($\log_{10}\mathcal{R}_p \approx 4.3$), at
$N_\text{sim} \in \{10^3, 10^4, 10^5\}$ and observations 1--10. For each cell we reload the trained Phase~1 sampler
of the corresponding \M run, without retraining. TSNPE then runs two rounds with the same per-round allocation as
\M's ratio estimator, $600 / 1\,000 / 5\,000$ simulations in round 1 and $400 / 9\,000 / 95\,000$ in round 2.
Round 1 draws from the prior truncated to the $1 - \epsilon$ HPR of a 10-component Gaussian mixture fitted to the
Phase~1 samples, and round 2 from the prior truncated to the HPR of the round-1 posterior at $x_o$, with
$\epsilon = 10^{-4}$. TSNPE cannot be trained on the simulations \M's ratio estimator uses. Maximum-likelihood NPE
trained on draws from a proposal $\tilde p(\theta)$ targets $\tilde p(\theta \mid x) \propto \tilde p(\theta)\,
p(x \mid \theta)$, which equals the posterior inside the truncation region only when $\tilde p$ is the truncated
prior. The ratio estimator instead learns $p(x \mid \theta)/q(x)$ (Eqn.~\ref{eq:phase_two_energy}) for any proposal whose support covers the
posterior, so \M can train on the Phase~1 pairs and on draws from its Phase~2 sampler directly. Using the same
pairs for TSNPE would require atomic SNPE-C, which suffers from leakage on wide priors, or importance weights
$p(\theta)/\tilde p(\theta)$, which need a sampler density that ASBS does not provide. We therefore use the
Phase~1 samples only to define the first truncation region and simulate afresh in both rounds, so TSNPE uses
$n_1$ more simulations than \M, which favours TSNPE. This flexibility in the choice of proposal is a structural
advantage of the ratio-based Phase~2. The density estimator
(NSF, 50 hidden units, independent $z$-scoring) and the C2ST protocol are those of the main SNPE baseline. In every
cell both truncation regions contain all reference samples, so the truncation never removes posterior mass.

Phase~1 + SNVI uses the configuration of our SNVI+fKL baseline (SNRE-B with a ResNet classifier, a neural spline
flow fitted by forward KL, and sampling-importance resampling) and is matched to \M in both simulations and rounds.
Its ratio estimator is trained in round 1 on the same $600 / 1\,000 / 5\,000$ Phase~1 pairs as \M's, and round 2
simulates the remaining budget from the round-1 variational posterior. Like the ASBS sampler of \M, the flow starts
at the Phase~1 sampler, fitted to its samples by maximum likelihood. We compute the importance weights of the forward
KL in log space, which avoids a numerical overflow of the released implementation that is detailed in
Appendix~\ref{app:baselines}.

\begin{table}[t]
\centering
\caption{Phase-2 ablation on Two Moons with the widest prior ($[-50, 50]^2$): mean (sd) C2ST over ten observations at matched simulation budget. Phase~1 + TSNPE uses the trained Phase~1 sampler of each \M run as the first truncation of TSNPE~\citep{deistler2022truncated}, with the same per-round simulation allocation as \M's ratio estimator ($n_1$ more simulations, since the Phase~1 pairs are not truncated-prior draws). Phase~1 + SNVI replaces \M's NRE and ASBS sampler by SNVI~\citep{DBLP:conf/iclr/GlocklerDM22} (SNRE-B and a normalising flow trained by forward KL), trained on exactly the same simulations in round 1 and the same number in round 2. SNPE and SNVI+fKL from the prior are the main-benchmark configurations.
 Phase~1 alone (524 simulations) reaches $0.995\,(0.001)$.}
\begin{tabular}{lccc}
\toprule
Method & $N_\text{sim}=10^3$ & $10^4$ & $10^5$ \\
\midrule
SNPE (prior, $R{=}2$) & $0.997\,(0.002)$ & $0.991\,(0.015)$ & $0.985\,(0.009)$ \\
SNVI+fKL (prior, $R{=}5$) & $0.987\,(0.004)$ & $0.943\,(0.023)$ & $0.993\,(0.020)$ \\
Phase~1 + TSNPE ($R{=}2$) & $0.925\,(0.032)$ & $0.873\,(0.083)$ & $0.738\,(0.107)$ \\
Phase~1 + SNVI ($R{=}2$) & $0.938\,(0.061)$ & $0.770\,(0.101)$ & $0.848\,(0.023)$ \\
\M (Phase~1 + NRE/ASBS) & $\mathbf{0.693\,(0.033)}$ & $\mathbf{0.617\,(0.034)}$ & $\mathbf{0.629\,(0.026)}$ \\
\bottomrule
\end{tabular}

\label{tab:tm-phase2-ablation}
\end{table}

\begin{figure}[t]
\centering
\includegraphics[width=0.6\linewidth]{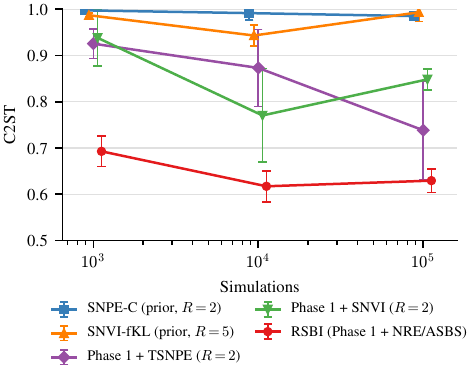}
\caption{C2ST against the reference posterior on Two Moons with the widest prior (mean $\pm$ sd over ten
observations, lower is better). Values as in Table~\ref{tab:tm-phase2-ablation}.}
\label{fig:tm-phase2-ablation}
\end{figure}

Table~\ref{tab:tm-phase2-ablation} and Figure~\ref{fig:tm-phase2-ablation} show that both phases are needed. On its
own, Phase~1 is a poor posterior estimate (C2ST $0.995$) because the surrogate is coarse, but it localises the
posterior reliably. Using it as the first proposal improves TSNPE by $0.07$, $0.12$ and $0.25$ in C2ST at $10^3$,
$10^4$ and $10^5$ simulations. SNPE from the prior barely improves at any budget, so without this zoom a
standard sequential method spends almost its whole budget in regions of negligible posterior mass.

The zoom alone is not sufficient, however. Given the same Phase~1 and the same per-round budget, replacing TSNPE by
\M's variational Phase~2 lowers the C2ST by a further $0.23$, $0.26$ and $0.11$, and \M is better than
Phase~1 + TSNPE on 10, 10 and 7 of the ten observations. Phase~2 accounts for most of the gain at $10^3$ and $10^4$
simulations, and \M reaches at $10^3$ simulations ($0.693$) an accuracy that Phase~1 + TSNPE does not reach at
$10^5$ ($0.738$). Two properties of the variational Phase~2 explain this. First, the ratio estimator is valid under
any proposal, so every simulation is used, including the Phase~1 pairs, whereas TSNPE must discard them and draw
prior-shaped samples from the truncation region. Second, ASBS samples directly from the current posterior estimate,
the prior times the learned ratio, so the round-2 simulations concentrate where the posterior has mass. TSNPE can
only spread them over the prior restricted to an HPR of its round-1 estimate, and after a coarse zoom on the wide
prior this region is still large compared with the posterior. At $10^5$ simulations TSNPE closes part of the gap, as
expected from a consistent estimator trained on a well-localised proposal, but it remains less accurate and much
less stable across observations (sd $0.107$ against $0.026$). Phase~1 + SNVI confirms that substantial gain comes from how Phase~2 refines the posterior and not from the ratio estimator
alone. With the same Phase~1 pairs and the same rounds as \M, it improves on SNVI+fKL from the prior at every
budget, but \M is better on all ten observations at every budget. At $10^3$ simulations the variational posterior
moved onto a spurious high-ratio region at the boundary of the prior in 4 of the 10 observations (C2ST $1.000$),
the failure described above. Both SNVI variants also share a non-monotonic dependence on the budget, with C2ST
rising from $0.943$ to $0.993$ (prior, $R{=}5$) and from $0.770$ to $0.848$ (Phase~1, $R{=}2$) between $10^4$ and
$10^5$ simulations. The two use different implementations, numbers of rounds and initialisations, so this
degradation reflects SNVI's variational refinement on a wide prior. \M improves
monotonically with the budget up to $10^4$ simulations and remains stable at $10^5$.
The advantage of \M on wide priors in this ablation study is therefore derivative of both the Phase~1 zoom and the variational ASBS refinement that follows it.

\section{Surrogate Likelihoods: Details and ablations}
\label{app:surrogate}

This appendix expands on the surrogate likelihood used in Phase~1 (Section~\ref{sec:stage1}). We stress at the
outset that the choice of distance, and hence of surrogate likelihood, is task dependent and deserves care from the
analyst. The ABC literature offers a wide range of choices, each motivated by a particular class of problems. These
include hand-crafted or learned summary statistics~\citep{fearnhead2012constructing}, Gaussian synthetic
likelihoods~\citep{wood2010statistical,price2018bayesian}, and full-data discrepancies such as the
Wasserstein distance~\citep{bernton2019approximate}, the maximum mean discrepancy~\citep{park2016kernelabc}, the
Kullback-Leibler divergence~\citep{jiang2018approximate} and the energy statistic~\citep{nguyen2020approximate}.
Comparative studies find that no single choice is best across problems~\citep{drovandi2022comparison}, and the ABC
kernel itself can be read as a model of the error between simulator and data~\citep{wilkinson2013approximate}, so the
bandwidth $\epsilon$ is a modelling choice as much as a tuning parameter.

\begin{figure}[t]
\centering
\includegraphics[width=\linewidth]{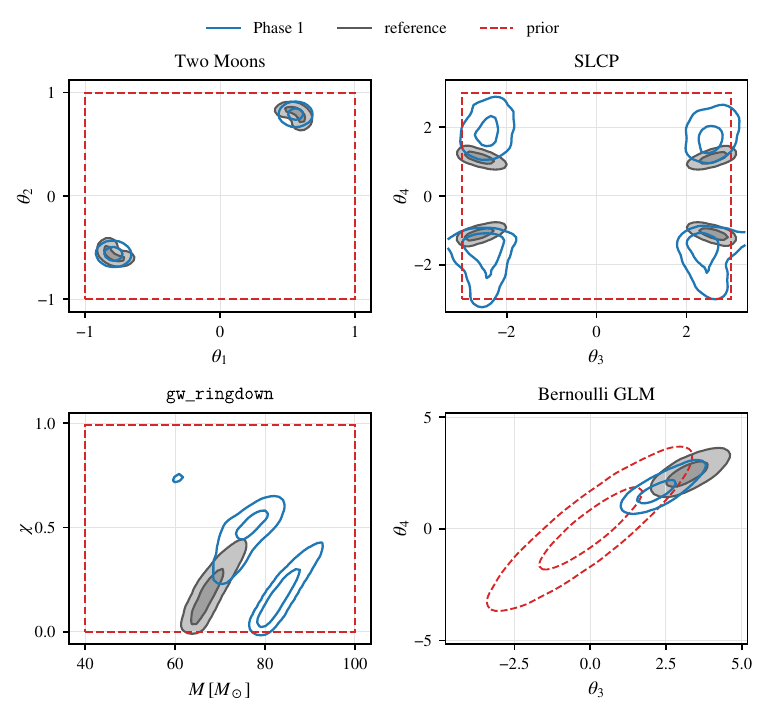}
\caption{Phase-1 posteriors used for the main results in Fig.~\ref{fig:c2st_main} for Two Moons, SLCP,
\texttt{gw\_ringdown} and the Bernoulli GLM (top left to bottom right), each at its default prior with
$N_\mathrm{sim} = 10^5$ at observation~1 (see Sec.~\ref{sec:hyperparam_tables} for the Phase-1 configuration). We
show the 1- and 2-$\sigma$ contours of the Phase-1 (blue) and reference (grey) posteriors over the prior (dashed red:
the prior box for the uniform priors, and the 1- and 2-$\sigma$ contours of the Gaussian prior for the Bernoulli
GLM). For SLCP and \texttt{gw\_ringdown} we show the 2-D marginals of the full 5-D and 6-D posteriors for the two
parameters with the most pronounced structure. These are $(\theta_3, \theta_4)$, which separate the four SLCP modes,
and the mass and spin pair $(M, \chi)$, whose posterior is a single curved ridge (Phase~1 also places small islands of
mass off the ridge). For the unimodal Bernoulli GLM we show the most strongly correlated pair of its 10 parameters.}
\label{fig:phase1-modes}
\end{figure}

We first describe the surrogate likelihoods used in this work and then discuss the bandwidth and metric trade-offs
known from the ABC literature. Next, we show how inspecting the energy and gradient landscape of a surrogate can
inform the choice of the bandwidth $\epsilon$. We close with ablations over the three surrogate likelihoods and over
$\epsilon$. Figure~\ref{fig:phase1-modes} shows the Phase-1 posteriors obtained with the surrogates used for the
main results. Phase~1 already concentrates on the posterior support and recovers its modes before any NRE is
trained.

\subsection{Gaussian negative log-likelihood with adaptive variance}
\label{app:surrogate:nll}

For a simulator with output $x = s(\theta, \omega) \in \mathbb{R}^{d_x}$, we model per-dimension
residuals as zero-mean Gaussian with diagonal covariance
$\mathrm{diag}\bigl(\epsilon_1(\theta), \dots, \epsilon_{d_x}(\theta)\bigr)$, which plays the role of
the bandwidth $\epsilon$. The surrogate distance is
\begin{equation}
    \D^\mathrm{NLL}\!\bigl(s(\theta), x_o\bigr)
    \;=\;
    \tfrac{1}{2}\sum_{d=1}^{d_x}
    \frac{\bigl(x^{(d)}_o - \bar{x}^{(d)}(\theta)\bigr)^2}{\epsilon_d(\theta)}
    \;+\;
    \tfrac{1}{2}\sum_{d=1}^{d_x}\log \epsilon_d(\theta),
    \label{eq:nll}
\end{equation}
where, as in Sec.~\ref{sec:stage1}, $s(\theta)$ denotes the $K$ replicates
$x_k = s(\theta,\omega_k)$, $\omega_k \overset{\text{iid}}{\sim} p(\omega)$, sharing the parameter
$\theta$ but differing in the simulator's internal randomness, and $\bar{x}^{(d)}(\theta)$ and
$\epsilon_d(\theta)$ are their per-dimension mean and unbiased variance.
Heteroscedastic Gaussian losses of this form have a long history in neural
regression~\citep{nix1994estimating} and are now standard for deep predictive uncertainty
quantification~\citep{lakshminarayanan2017deepensembles}. We use PyTorch's
\texttt{gaussian\_nll\_loss} with \texttt{full=False}, which omits the constant
$\tfrac{d_x}{2}\log 2\pi$ (irrelevant for gradients by Theorem~\ref{th:equivalence}). The variance
term in~\eqref{eq:nll} acts as an entropy-like penalty that prevents the bridge from concentrating
in regions where the simulator is highly stochastic.

We support two options for the variance $\epsilon_d(\theta)$.
\textbf{(i) Fixed:} $\epsilon_d(\theta) = \epsilon^\star_d$ for a user-supplied constant. A single
simulator call per $\theta$ suffices ($K=1$), appropriate when the noise model is exact.
\textbf{(ii) Adaptive:} $\epsilon_d(\theta)$ is recomputed at every $\theta$ from the
$K$ replicates. It varies smoothly with $\theta$ but is noisy at small $K$.

With fixed variance, \eqref{eq:nll} is a weighted squared distance plus the constant
$\tfrac12\sum_d\log\epsilon^\star_d$, hence a sufficient distance measure
(Definition~\ref{def:sufficient_measure}) up to a constant, covered by
Theorem~\ref{th:equivalence}. In the adaptive case, $\epsilon_d = \epsilon_d(\theta,\omega_{1:K})$ is
the sample variance of the replicates drawn at $\theta$, recomputed at every evaluation and
differentiated through. The term $\tfrac12\sum_d\log\epsilon_d$ is then not a constant, since its
expectation varies with $\theta$ wherever the simulator noise is heteroscedastic, so
\eqref{eq:nll} is not a PSD distance up to a constant. It nonetheless defines a proper surrogate:
the variance is floored, so $\D^\mathrm{NLL}$ is bounded below, $\exp(-\mathbb{E}\,\D^\mathrm{NLL})\,p(\theta)$
is integrable for any proper prior as per Def.\ref{def:sufficient_measure}, and \eqref{eq:surrogate_energy} remains the gradient of a well-defined energy that ASBS targets as before. Any mismatch with the true likelihood is removed
in Phase~2.

\subsection{Energy distance}
\label{app:surrogate:ed}
We also consider the energy statistic of~\citet{szekely2013energy}, a
characteristic-kernel discrepancy known to be equivalent to MMD with a
distance-induced kernel~\citep{sejdinovic2013equivalence,%
DBLP:journals/jmlr/GrettonBRSS12}, whose sample estimator differentiates
only through the (smooth) simulator output and never through a
non-differentiable comparison. We instantiate it as $\D$ by identifying
$P$ with the simulator distribution $p(x \mid \theta)$ and $Q$ with
the (point-mass) data distribution at $x_o$. For two distributions
$P, Q$ on $\mathbb{R}^{d_x}$ and the Euclidean kernel
$k(u, v) = -\|u - v\|_2$, the squared energy distance is
\begin{equation}
    \mathcal{E}^2(P, Q)
    \;=\;
    2\,\mathbb{E}_{X \sim P,\,Y \sim Q}\|X - Y\|
    \;-\;
    \mathbb{E}_{X, X' \sim P}\|X - X'\|
    \;-\;
    \mathbb{E}_{Y, Y' \sim Q}\|Y - Y'\|,
    \label{eq:ed_pop}
\end{equation}
with $\mathcal{E}^2(P, Q) = 0$ iff $P = Q$~\citep{szekely2013energy}.
The bracket in Eq.~\eqref{eq:ed_unbiased} is its minimum-variance
unbiased $U$-statistic. For each $\theta$ in the training minibatch we
draw $K$ i.i.d.\ simulator replicates
$\{x_k\}_{k=1}^{K} = \{s(\theta;\, \omega_k)\}_{k=1}^{K}$ as samples from
$P$, and treat the $N$ available observations $\{y_n\}_{n=1}^{N}$ as
samples from $Q$ (in our setting $N = 1$ and $y_1 = x_o$). The unbiased
$U$-statistic estimator, divided by a length scale $\epsilon > 0$, is then
\begin{equation}
\begin{split}
     \D^\mathrm{ED}\!\bigl(s(\theta), x_o\bigr)
  ={}&
    \frac{1}{\epsilon}\Biggl[
    \frac{2}{K N}\sum_{k=1}^{K}\sum_{n=1}^{N}\!\|x_k - y_n\|
    \;-\;
    \frac{1}{K(K-1)}\sum_{k\neq k'}\|x_k - x_{k'}\| \\
   &\;-\;
    \frac{1}{N(N-1)}\sum_{n\neq n'}\|y_n - y_{n'}\|
    \Biggr],
\end{split}
    \label{eq:ed_unbiased}
\end{equation}
with the diagonals $k = k'$ and $n = n'$ excluded; the third term
vanishes when $N = 1$. For a fixed $\epsilon$ the estimator is unbiased, and its expectation over the replicates is exactly the population
quantity, $\mathbb{E}\,\D^\mathrm{ED}(s(\theta),x_o) = \mathcal{E}^2\bigl(p(x\mid\theta),\delta_{x_o}\bigr)/\epsilon$,
so the surrogate of Sec.~\ref{sec:stage1} is the energy distance itself. For $N=1$ the estimator is
also non-negative, since $\|x_k - x_{k'}\| \le \|x_k - x_o\| + \|x_{k'} - x_o\|$, so it is a sufficient
distance measure (Definition~\ref{def:sufficient_measure}).
 Since $\mathcal{E}$ carries the units of $x$,
$ \frac{1}{\epsilon}$ makes $\D^\mathrm{ED}$ unit-less. It does not change the
minimiser and only weights the discrepancy against the prior. By default $\epsilon$ is adaptive. It is the median pairwise distance
among simulations at $256$ prior draws (counted toward the simulation
budget) and updated at every energy evaluation by an exponential moving
average (decay $0.95$) of the minibatch-averaged median pairwise
distance among $\{x_k\}_{k=1}^{K} \cup \{x_o\}$, with no gradient
flowing through $\epsilon$. Because $\epsilon$ is a single detached scale, updated with weight $0.05$ from a
minibatch average, it depends only weakly on the replicates of any one $\theta$. Within
each evaluation we treat it as fixed, so unbiasedness holds conditionally on $\epsilon$,
while across training $\epsilon$ drifts and the Phase-1 target with it. 
With a single replicate ($K = 1$, used on Two~Moons) the second term in \eqref{eq:ed_unbiased} is absent and
$\D^\mathrm{ED}$ reduces to $2\|x_1 - x_o\|/\epsilon$, whose expectation exceeds
$\mathcal{E}^2/\epsilon$ by $\mathbb{E}\|X - X'\|/\epsilon$. For Two~Moons the simulator noise is
additive, so this offset does not depend on $\theta$ and, by Theorem~\ref{th:equivalence}, the
gradient is that of the energy distance itself. The median over $\{x_1, x_o\}$ is then simply
$\|x_1 - x_o\|$, so $\epsilon$ tracks a moving average of the minibatch-mean distance between the
simulations and $x_o$. It starts at the scale of the prior predictive and shrinks as Phase~1
concentrates on the posterior, which anneals the surrogate from smooth to sharp during training
(Appendix~\ref{app:bandwidth-ablations} compares this with fixed length scales).

\subsection{Maximum mean discrepancy (MMD)}
\label{app:surrogate:mmd}

We also consider the squared maximum mean discrepancy
(MMD)~\citep{DBLP:journals/jmlr/GrettonBRSS12} with the Gaussian (RBF)
kernel $k_\epsilon(u, v) = \exp\!\bigl(-\|u - v\|_2^2 / (2\epsilon^2)\bigr)$
of bandwidth $\epsilon > 0$, which is characteristic, so that
$\mathrm{MMD}_\epsilon^2(P, Q) = 0$ iff $P = Q$. As for the energy
distance, we identify $P$ with the simulator distribution
$p(x \mid \theta)$ and $Q$ with the (point-mass) data distribution at
$x_o$, and
\begin{equation}
    \mathrm{MMD}_\epsilon^2(P, Q)
    \;=\;
    \mathbb{E}_{X, X' \sim P}\, k_\epsilon(X, X')
    \;+\;
    \mathbb{E}_{Y, Y' \sim Q}\, k_\epsilon(Y, Y')
    \;-\;
    2\,\mathbb{E}_{X \sim P,\,Y \sim Q}\, k_\epsilon(X, Y).
    \label{eq:mmd_pop}
\end{equation}
For each $\theta$ in the training minibatch we draw $K$
i.i.d.\ simulator replicates
$\{x_k\}_{k=1}^{K} = \{s(\theta;\, \omega_k)\}_{k=1}^{K}$ and treat the $N$
observations $\{y_n\}_{n=1}^{N}$ as samples from $Q$ (here $N = 1$,
$y_1 = x_o$). Since the unbiased $U$-statistic of the $Q$--$Q$ term is
undefined for $N = 1$, we use the plug-in ($V$-statistic) estimator
\begin{equation}
\begin{split}
    \D^\mathrm{MMD}\!\bigl(s(\theta), x_o\bigr)
  ={}&
    \frac{1}{K^2}\sum_{k=1}^{K}\sum_{k'=1}^{K} k_\epsilon(x_k, x_{k'})
    \;+\;
    \frac{1}{N^2}\sum_{n=1}^{N}\sum_{n'=1}^{N} k_\epsilon(y_n, y_{n'}) \\
   &\;-\;
    \frac{2}{K N}\sum_{k=1}^{K}\sum_{n=1}^{N} k_\epsilon(x_k, y_n),
\end{split}
    \label{eq:mmd_v}
\end{equation}
diagonals included, so that for $N = 1$ the second term equals $1$. As
$\D^\mathrm{MMD} \in [0, 2]$ is unit-less, no length-scale normalisation
is applied. The bandwidth $\epsilon$ is either fixed or, by default, set
by the median heuristic separately for each $\theta$: $\epsilon(\theta)$
is the median pairwise Euclidean distance among
$\{x_k\}_{k=1}^{K} \cup \{x_o\}$, recomputed at every energy evaluation
from the current replicates (no pilot simulations, no smoothing), and
gradients are propagated through $\epsilon(\theta)$. The $V$-statistic lies in $[0,2]$, so it is a sufficient distance measure
(Definition~\ref{def:sufficient_measure}), but unlike the energy distance its expectation is biased,
$\mathbb{E}\,\D^\mathrm{MMD} = \mathrm{MMD}_\epsilon^2 + \tfrac{1}{K}\bigl(1 - \mathbb{E}\,k_\epsilon(X,X')\bigr)$
for fixed $\epsilon$, an additional penalty on simulator spread that vanishes as $K\to\infty$. With
the median heuristic, $\epsilon(\theta)$ depends on the replicates and is differentiated through, as
for the adaptive variance of Appendix~\ref{app:surrogate:nll}. MMD is used only in the surrogate
ablations (Appendices~\ref{app:surrogate-ablations} and~\ref{app:bandwidth-ablations}).

\subsection{Bandwidth and Metric Trade-offs}
\label{app:surrogate:tradeoffs}

The impact of $\tilde{\D}$ and $\epsilon$ on $\tilde{p}_\epsilon(x_o\mid\theta)$ has been studied
extensively in the ABC literature~\citep{Marin2011ApproximateBC, 10.1093/genetics/162.4.2025,
fearnhead2012constructing, park2016kernelabc}. Under mild regularity, $\tilde{p}_\epsilon \to p$ as
$\epsilon \to 0$, but small $\epsilon$ degrades acceptance and inflates Monte Carlo variance, while
large $\epsilon$ over-smooths the likelihood and biases the posterior~\citep{blum2010nonparametric}.
The choice of $\tilde{\D}$ involves similar considerations, and kernel-based discrepancies such as
MMD and the energy distance can improve the bias-variance trade-off relative to a fixed summary
statistic.

In \M the trade-off takes a different form, because Phase~1 samples the surrogate with gradients
rather than by acceptance. A small $\epsilon$ gives a sharp energy whose gradient is dominated by
Monte Carlo noise near $x_o$ and vanishes far from it. A large $\epsilon$ flattens the data term
until the drift is set by the prior alone. Since Phase~2 NRE-ASBS refinement targets the true posterior,
the surrogate need not approximate $p(\theta\mid x_o)$ precisely. It only needs to place mass in the
region that Phase~2 then refines, so larger $\epsilon$ values can be tolerated than in standalone ABC.
The surrogates we use are not guaranteed to reproduce the mode, shape or variance of the true
likelihood, and Phase~2 corrects the resulting bias.

Our ablations show that both choices depend on the simulator and the data, and that no single
surrogate or bandwidth rule is best across tasks. Among surrogates (Appendix~\ref{app:surrogate-ablations}),
the energy distance gives the best Phase-1 proposal on Two Moons and the Bernoulli GLM, where MMD
fails to concentrate at all. On SLCP all three surrogates are within one standard deviation of each
other in C2ST, and MMD places much of its mass on the edges of the prior box. Whether the bandwidth
should be adaptive or fixed is equally task dependent (Appendix~\ref{app:bandwidth-ablations}). On SLCP
an adaptive per-$\theta$ variance is needed for the Gaussian NLL, because $\theta_3$, $\theta_4$ and
$\theta_5$ enter the simulator only through the noise covariance, and no fixed variance discovers the
modes better than prior samples do. On the Bernoulli GLM the opposite holds. Per-$\theta$ variances
estimated from a few binary replicates are too noisy, adaptive bandwidths fail (C2ST $\geq 0.995$) and
a fixed bandwidth is required. Sensitivity to $\epsilon$ also varies. Misspecifying the best bandwidth
by an order of magnitude costs $0.13$ to $0.18$ in Phase-1 C2ST on Two Moons and raises it to $0.98$ or
more on the Bernoulli GLM, whereas on SLCP the C2ST changes by less than $0.03$ over four orders of
magnitude. The pilot landscape heuristic of Sec.~\ref{sec:energy_landscape_opt} chooses bandwidths
without the reference posterior that lie within half a decade of the best value on Two Moons and
coincide with it on the Bernoulli GLM. It remains a heuristic, however, and it does not decide between
surrogates or between adaptive and fixed bandwidths. Making these choices in a principled way requires
a better understanding of the gradient landscape that a differentiable simulator and a distance induce
over the prior, which we leave to future work.

\subsection{Bandwidth and temperature hyperparameter guesses via energy and gradient landscape}
\label{sec:energy_landscape_opt}
\begin{figure}[t]
\centering
\includegraphics[width=\linewidth]{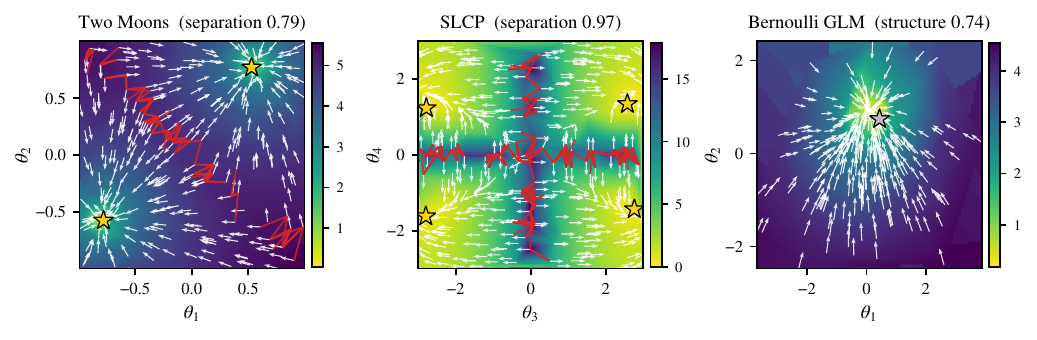}
\caption{Pilot energy landscapes for Two Moons (left), SLCP (centre) and the Bernoulli GLM (right), computed with the Phase-1 surrogate, bandwidth and temperature used for the benchmark main results at the default prior on  observation~1 (see Sec.~\ref{sec:hyperparam_tables} for Phase-1 hyperparamters and surrogates used).  Colour is $\log(1 + E/T - \min E/T)$; white arrows show the direction of $-\nabla_\theta E$; stars are basin sinks (gold: significant); red edges are ridges. SLCP and the Bernoulli GLM are shown on 2-D slices through $(\theta_3, \theta_4)$ and $(\theta_1, \theta_2)$. The number in each panel title is the score used to rank bandwidths for that posterior: separation for the multimodal Two Moons and SLCP, structure for the unimodal Bernoulli GLM. Scores for these settings: Two Moons: separation 0.79, structure 0.79; SLCP: separation 0.97, structure 0.21; Bernoulli GLM: separation 0.00, structure 0.74.}
\label{fig:landscape}
\end{figure}

We suggest a cheap heuristic to aid in choosing the surrogate bandwidth $\epsilon$ and the Phase-1 temperature $T$. This method leverages a cheap pilot run of $\mathcal{O}(10^2)$ draws from the prior in order to observe the landscape of gradients and energy throughout the prior volume. The assumption is that bandwidth $\epsilon$ and temperature $T$ configurations that provide informative gradients for the Phase-1 zoom   should produce a structured  landscape with gradients that cohesively point to one or more energy wells, no matter where they are within the prior. We detail this process below, noting that exploring a fully fledged hyperparameter optimization regime for WT-ASBS is generally interesting and could constitute future work.

Before any training, we evaluate the surrogate energy
$E(\theta) = \D\bigl(s(\theta), x_o\bigr) - \log p(\theta)$ on a small pilot batch of prior draws with a few
simulator replicates $\omega_{1:K}$ each, look at its landscape, and score candidate bandwidths on the same simulations. The
heuristic needs no training and no reference posterior, and its only cost is the pilot simulations. 

At each pilot draw we record the energy and the drift $-\nabla_\theta E$, obtained by back-propagating through the
differentiable simulator. Splitting the replicates into two halves gives two independent estimates $E_1, E_2$, and
the Monte Carlo noise of the energy is $\sigma_E = \tfrac12\,\mathrm{median}\,|E_1 - E_2|$. All quantities below are
computed on the $k$-nearest-neighbour graph of the pilot draws, so every candidate $\epsilon, T$ is evaluated on the
same simulations without new simulator calls. In more than two dimensions a projection of the draws mixes points
that differ in the hidden parameters,  so the maps are drawn on 2-D slices of two nominal parameters only.

Each draw steps to the neighbour that is lower in energy and best aligned with its drift. Following these steps
downhill ends in a sink, and the draws that end in the same sink form one basin, or well; its share of the draws is
the share of prior mass it attracts. Two neighbouring draws in different basins whose drifts point away from each
other, each drift having a negative projection onto the direction to the other draw, lie on a \textit{ridge}. The barrier
$\Delta E$ of a basin is the lowest energy on its boundary (the higher endpoint of a cross-basin edge) minus the
basin's minimum. Adjacent basins are merged unless the barrier between them is at least $3\sigma_E$ deep and at least 30\% of their boundary pairs diverge, so that only boundaries where the drift genuinely splits between two wells remain.

We introduce three metrics that can be used to score a given surrogate bandwidth $\epsilon$. First, the \emph{cohesion}: the median cosine between a
draw's drift and the direction to its basin's sink, which measures whether the drift points into its own well.  Second, the
\emph{ridge} fraction: the share of cross-basin neighbour pairs whose drifts diverge, which measures whether the boundaries between wells are genuine ridges. Third, the \emph{depth}: the mean over basins of $\min(1, \Delta E / 10\sigma_E)$,
which measures whether the barriers stand above the Monte Carlo noise. For a single well the ridge term is $1$ and
the depth uses the energy contrast $(E_{95} - E_{\min})/\sigma_E$ in place of a barrier. They combine into
\begin{equation}
  \mathrm{structure} = \mathrm{cohesion} \times \mathrm{ridge} \times \mathrm{depth}, \qquad
  \mathrm{separation} = \begin{cases} \mathrm{ridge} \times \mathrm{depth}, & \ge 2 \text{ wells,} \\
                                      0, & \text{one well.} \end{cases}
\end{equation}
Structure rewards a drift that points coherently into wells that are deep and cleanly separated, whereas separation
rewards only the separation of several wells, regardless of the drift inside each. For a multimodal energy landscape we
take the $\epsilon$, or the adaptive variant, with the highest separation.  For a unimodal energy landscape, we take the highest
structure.For certain priors, a large $\epsilon$ leaves only the prior's own
well, which every score rates as perfect, so only settings whose median data gradient is at least three times the
prior gradient are considered. 

The landscape can inform the temperature $T$, but only when WT is on. The target $\exp(-E/T)$ crosses a barrier of height $\Delta E$ with a
probability that scales as $\exp(-\Delta E/T)$ (Sec.~\ref{sec:well_tempering}), so what matters is the barrier in units of the temperature,
$\Delta E/T$. Well-tempered metadynamics (WT) with bias factor $\gamma$ fills the wells and lowers the effective 
barrier to roughly $\Delta E/\gamma$, leaving $\Delta E/(\gamma T)$. Let $\kappa$ denote the barrier height, in units
of $T$, that we want to remain after the bias. Setting $\Delta E/(\gamma T) = \kappa$, with $\Delta E$ the median
barrier of the significant wells, gives
\begin{equation}
  T =
  \begin{cases}
    1, & \text{one well (no WT),}\\[2pt]
    \dfrac{\Delta E}{\gamma\,\kappa}, & \text{several wells.}
  \end{cases}
\end{equation}
The value of $\kappa$ trades mode separation against mixing. A remaining barrier of $\kappa \approx 3$ gives a
crossing factor $e^{-3} \approx 5\%$. Much smaller values flatten the wells and blur the modes, and much larger ones let the sampler stay trapped in the first well it finds. We use $\kappa = 3$ as a heuristic default rather than a tuned value. A single well needs neither tempering nor a bias, so $T = 1$.

Figure~\ref{fig:landscape} shows such landscapes for the Two moons, SLCP and Bernoulli GLM tasks. The colour shows the energy, with the white arrows depicting the 
drift/gradients from the average pilot. The stars depict the location of the basin sinks and the red edges show ridges such that the wells and barriers between energy basins can be
inspected directly. Two Moons shows two wells separated by a ridge along $\theta_1 = -\theta_2$ whilst SLCP shows four wells, one for each sign combination of $(\theta_3, \theta_4)$, separated by ridges along $\theta_3 = 0$. We note that this energy landscape was only seen for the Gaussian surrogate on SLCP. The Bernoulli GLM shows a single well with the drift pointing into it from everywhere. Table~\ref{tab:landscape-scores} lists the scores of these landscapes.

\begin{table}[t]
\centering
\begin{tabular}{lcccccc}
\toprule
Task & wells & cohesion & ridge & depth & structure & separation \\
\midrule
Two Moons & 2 & 1.00 & 0.79 & 1.00 & 0.79 & 0.79 \\
SLCP & 4 & 0.22 & 0.97 & 1.00 & 0.21 & 0.97 \\
Bernoulli GLM & 1 & 0.74 & -- & 1.00 & 0.74 & 0.00 \\
\bottomrule
\end{tabular}
\caption{Landscape scores of the pilot batches in Fig.~\ref{fig:landscape}. \emph{Wells}: significant basins. \emph{Cohesion}: median cosine between a draw's drift and the direction to its basin's sink. \emph{Ridge}: share of cross-basin neighbour pairs whose drifts point apart (-- for a single well, where the ridge term is $1$). \emph{Depth}: mean over basins of $\min(1, \Delta E / 10\sigma_E)$, or $\min(1, (E_{95} - E_{\min}) / 10\sigma_E)$ for a single well. Structure $=$ cohesion $\times$ ridge $\times$ depth. Separation $=$ ridge $\times$ depth for two or more wells and $0$ otherwise.}
\label{tab:landscape-scores}
\end{table}

\subsection{Surrogate likelihood ablations}
\label{app:surrogate-ablations}

\begin{figure}[h]
\centering
\includegraphics[width=\linewidth]{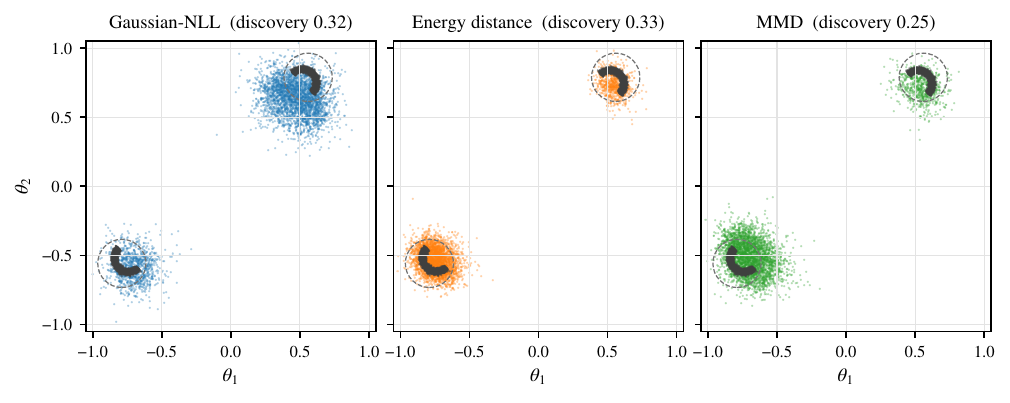}
\caption{Phase-1 samples on Two Moons for each surrogate's profile (observation~1, $10^{4}$ simulations; colour) over the reference posterior (grey). Dashed circles: the regions used for mode discovery ($1.5\times$ each moon's 95th-percentile reference radius). All three surrogates place a cluster on each moon; discovery is limited by how unevenly the mass is split between the moons and by clusters wider than the moons themselves.}
\label{fig:two_moons-modes}
\end{figure}

\begin{figure}[htbp]
\centering
\includegraphics[width=\linewidth]{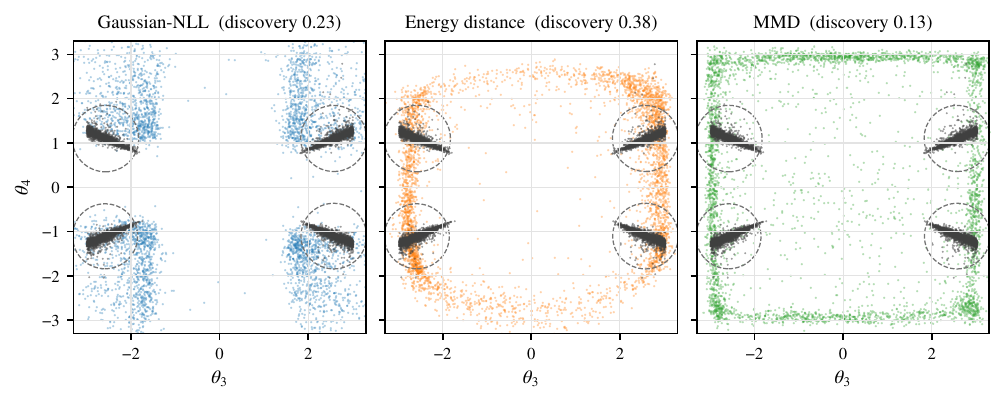}
\caption{Phase-1 samples on SLCP in the mode-separating coordinates $(\theta_3, \theta_4)$ for each surrogate's profile (observation~1, $10^{4}$ simulations; colour) over the reference posterior (grey). Dashed circles: the regions used for mode discovery ($1.5\times$ each mode's 95th-percentile reference radius). The Gaussian-NLL puts its mass beside all four modes ($|\theta_3| \approx 2$), spread along $\theta_4$; the energy distance spreads its mass in a ring around the prior box that passes through them; MMD concentrates on the box edges.}
\label{fig:slcp-modes}
\end{figure}

We compare three surrogate distances $\D(s(\theta,\omega), x_o)$ as the Phase-1
energy: the Gaussian negative log-likelihood (Gaussian-NLL), the energy distance and
the maximum mean discrepancy (MMD). We run the comparison on Two Moons, SLCP and the
Bernoulli GLM, each at its baseline prior. Within a task all surrogates share the same
Phase-1 backbone, and training is capped at the simulation budget
$N_\mathrm{sim} \in \{10^3, 10^4, 10^5\}$. All entries are the mean\,$\pm$\,standard
deviation over 10 observations. Tables~\ref{tab:two_moons-c2st-vs-ref}--\ref{tab:bernoulli_glm-mode-fidelity}
report the results; for the multimodal tasks, Figs.~\ref{fig:two_moons-modes} and~\ref{fig:slcp-modes} show
the Phase-1 samples of each surrogate on one observation.

These ablations are experiments in their own right, separate from the benchmark runs:
rather than reusing the benchmark settings, each surrogate is given its own bandwidth
$\epsilon$ and temperature $T$, so that we compare reasonably tuned surrogates rather than one surrogate's defaults. These settings are chosen using the heuristic method of Sec.~\ref{sec:energy_landscape_opt} from a pilot batch of prior-predictive simulations, which uses no reference posterior. On Two Moons, $\epsilon$ is fixed at the scale of the pilot simulations (a
Gaussian-NLL variance of $1.0 \times 10^{-3}$; an energy-distance length scale and MMD
bandwidth of $0.78$, the median pairwise distance), which is also the value that best
separates the two energy wells of the pilot landscape. $T$ is then set from the same
landscape so that, after well-tempered metadynamics with bias factor $10$, about $3\,kT$
of the median barrier between the wells remains: $T = 12.8$, $0.068$ and $0.032$, with
one simulator replicate per $\theta$. On SLCP the bandwidths are adaptive (as in the
benchmark), and $T$ is set from the energy landscape in the same way: $T = 11.1$ for the
Gaussian-NLL and $T = 0.011$ for MMD, while the energy distance keeps $T = 10^{-3}$ from
short pilot runs on observation~1, with three replicates per $\theta$. On the Bernoulli GLM we keep $T = 1$ for all surrogates, because the
prior is Gaussian and $T$ would temper it as well; sharpness is instead set through a
fixed $\epsilon$ from the pilot: a Gaussian-NLL variance of $0.05$, an energy-distance
length scale of $0.914$ (as in the benchmark runs) and an MMD bandwidth of $54.2$
(about the median pairwise distance of the pilot simulations), with one, two and two
replicates per $\theta$.

Phase~1 only has to seed Phase~2 with a proposal that places mass on every posterior
mode; it does not have to match the posterior. We therefore report two metrics, and a
third, mode discovery, for the multimodal Two Moons and SLCP. The
first is the C2ST of the Phase-1 samples against the reference posterior (lower is
better, $0.5$ means indistinguishable). The second is a \emph{mode fidelity}, which
measures how mass is distribued across modes, something C2ST does not explicitly test. Let
$\tau$ be the $95$th percentile of the nearest-neighbour distances within the reference
sample. The \emph{precision} $P$ is the fraction of Phase-1 samples that lie within
$\tau$ of a reference sample. The \emph{recall} $R_m$ of mode $m$ is the fraction of
reference samples in mode $m$ that have a Phase-1 sample within $\tau$. The mode
fidelity is
\begin{equation}
  F = P \cdot \min_m R_m \in [0, 1],
\end{equation}
with $1$ ideal. $F \to 0$ if Phase~1 misses a mode ($\min_m R_m \to 0$) or spreads its
mass too thinly over the prior ($P \to 0$); $F > 0$ means every mode is covered. For the
unimodal Bernoulli GLM, $F$ reduces to precision times recall. On SLCP we compute $F$ on
$(\theta_3, \theta_4)$ only: these coordinates separate the four modes, whereas the posterior is so
narrow in the full five dimensions that almost no Phase-1 sample falls within $\tau$ of a reference
sample, even when Phase~1 has found all four modes. Because $F$ rewards samples that land on the
reference support rather than near a mode, on Two Moons and SLCP we also report a coarser
\emph{mode discovery}: each Phase-1 sample is assigned to the nearest reference-mode centre (in $\theta$
for Two Moons, in $(\theta_3, \theta_4)$ for SLCP) if it lies within $1.5$ times that mode's
95th-percentile reference radius (at most half the distance between mode centres), and discovery is
$K$ times the smallest per-mode share ($K$ modes; $1$ means all mass split equally over the modes, $0$ a
missed mode). As a reference point, samples drawn from the prior score $0.04$ on Two Moons and $0.23$ on
SLCP. Because the Phase-1
proposal is much broader than the posterior, $F$ is small, and we report it in units of
$10^{-3}$.

\paragraph{Two Moons.}
 Table~\ref{tab:two_moons-c2st-vs-ref}) shows that with larger budgets the
C2ST improves monotonically, to $0.875$ for the energy distance, $0.922$ for MMD and
$0.941$ for the Gaussian-NLL at $10^5$. It never reaches $0.5$, as
expected, because Phase~1 learns the posterior under the surrogate likelihood rather
than the simulator likelihood. Mode fidelity (Table~\ref{tab:two_moons-mode-fidelity})
is positive for every surrogate and budget, so both moons are always seeded. The energy
distance gives by far the best coverage at $10^4$ and $10^5$ simulations, albeit with
large variation across observations. Mode discovery (Table~\ref{tab:two_moons-mode-discovery}, Fig.~\ref{fig:two_moons-modes}) ranks
the surrogates as C2ST does: at $10^3$ simulations all three are barely above the prior ($0.09$--$0.11$
against $0.04$), while at $10^5$ the energy distance reaches $0.64$, MMD $0.53$ and the Gaussian-NLL $0.32$.

\begin{table}[htbp]
\centering
\caption{Phase-1 \textbf{C2ST} on Two Moons (baseline prior) vs.\ the simulation budget (capped), mean\,$\pm$\,std over 10 observations. Lower is better ($0.5$ = indistinguishable from the reference).}
\begin{tabular}{lccc}
\toprule
Surrogate & $10^{3}$ & $10^{4}$ & $10^{5}$ \\
\midrule
Gaussian-NLL & $0.981\!\pm\!0.005$ & $0.955\!\pm\!0.017$ & $0.941\!\pm\!0.011$ \\
Energy distance & $0.982\!\pm\!0.008$ & $0.891\!\pm\!0.047$ & $0.875\!\pm\!0.026$ \\
MMD & $0.981\!\pm\!0.007$ & $0.934\!\pm\!0.026$ & $0.922\!\pm\!0.012$ \\
\bottomrule
\end{tabular}

\label{tab:two_moons-c2st-vs-ref}
\end{table}

\begin{table}[h]
\centering
\caption{Phase-1 \textbf{mode fidelity} on Two Moons (baseline prior) vs.\ the simulation budget (capped), mean\,$\pm$\,std over 10 observations. Values $\times1000$.}
\begin{tabular}{lccc}
\toprule
Surrogate & $10^{3}$ & $10^{4}$ & $10^{5}$ \\
\midrule
Gaussian-NLL & $1.560\!\pm\!1.092$ & $10.560\!\pm\!6.981$ & $19.030\!\pm\!11.230$ \\
Energy distance & $1.270\!\pm\!1.328$ & $57.630\!\pm\!24.147$ & $101.430\!\pm\!37.410$ \\
MMD & $1.480\!\pm\!1.282$ & $21.500\!\pm\!10.500$ & $38.990\!\pm\!12.119$ \\
\bottomrule
\end{tabular}

\label{tab:two_moons-mode-fidelity}
\end{table}

\begin{table}[h]
\centering
\caption{Phase-1 \textbf{mode discovery} on Two Moons (baseline prior) vs.\ the simulation budget (capped), mean\,$\pm$\,std over 10 observations. Higher is better. Samples drawn from the prior score $0.04$.}
\begin{tabular}{lccc}
\toprule
Surrogate & $10^{3}$ & $10^{4}$ & $10^{5}$ \\
\midrule
Gaussian-NLL & $0.114\!\pm\!0.048$ & $0.260\!\pm\!0.105$ & $0.323\!\pm\!0.181$ \\
Energy distance & $0.093\!\pm\!0.063$ & $0.527\!\pm\!0.114$ & $0.639\!\pm\!0.234$ \\
MMD & $0.107\!\pm\!0.055$ & $0.350\!\pm\!0.115$ & $0.530\!\pm\!0.132$ \\
\bottomrule
\end{tabular}

\label{tab:two_moons-mode-discovery}
\end{table}

\FloatBarrier

\paragraph{SLCP.}
All three surrogates give a similar Phase-1 C2ST on SLCP, between $0.97$ and $0.98$ at
every budget (Table~\ref{tab:slcp-c2st-vs-ref}); the differences are within one standard
deviation. Mode fidelity on $(\theta_3, \theta_4)$ (Table~\ref{tab:slcp-mode-fidelity}) is
positive for every surrogate and budget, so all three seed all four modes. It is highest for the
Gaussian-NLL (at $10^3$ and $10^5$ simulations) and the energy distance (at $10^4$); at $10^4$ and
$10^5$ it is three to four times lower for MMD, which places much of its mass on the edges of the
prior box. Mode discovery (Table~\ref{tab:slcp-mode-discovery}, Fig.~\ref{fig:slcp-modes}) shows how
weakly Phase~1 concentrates on SLCP at these budgets: the Gaussian-NLL and the energy distance are only
moderately above the prior ($0.34$--$0.39$ at $10^4$ and $10^5$ simulations, against $0.23$), with the
Gaussian-NLL ahead at $10^3$ ($0.29$), and MMD stays at the level of the prior ($0.16$--$0.24$); the
variation across observations is large. The posterior itself is resolved in Phase~2.

\begin{table}[htbp]
\centering
\caption{Phase-1 \textbf{C2ST} on SLCP (baseline prior) vs.\ the simulation budget (capped), mean\,$\pm$\,std over 10 observations. Lower is better ($0.5$ = indistinguishable from the reference).}
\begin{tabular}{lccc}
\toprule
Surrogate & $10^{3}$ & $10^{4}$ & $10^{5}$ \\
\midrule
Gaussian-NLL & $0.979\!\pm\!0.011$ & $0.976\!\pm\!0.013$ & $0.971\!\pm\!0.016$ \\
Energy distance & $0.983\!\pm\!0.007$ & $0.973\!\pm\!0.015$ & $0.970\!\pm\!0.015$ \\
MMD & $0.981\!\pm\!0.011$ & $0.981\!\pm\!0.008$ & $0.979\!\pm\!0.014$ \\
\bottomrule
\end{tabular}

\label{tab:slcp-c2st-vs-ref}
\end{table}

\begin{table}[htbp]
\centering
\caption{Phase-1 \textbf{mode fidelity} on SLCP (baseline prior) vs.\ the simulation budget (capped), mean\,$\pm$\,std over 10 observations. Values $\times1000$. Mode fidelity is computed on $(\theta_3, \theta_4)$ only, the coordinates that separate the four SLCP modes (and carry the metadynamics bias).}
\begin{tabular}{lccc}
\toprule
Surrogate & $10^{3}$ & $10^{4}$ & $10^{5}$ \\
\midrule
Gaussian-NLL & $68.642\!\pm\!75.682$ & $111.676\!\pm\!133.676$ & $138.893\!\pm\!158.338$ \\
Energy distance & $41.317\!\pm\!46.915$ & $130.302\!\pm\!131.649$ & $112.418\!\pm\!109.132$ \\
MMD & $41.959\!\pm\!40.802$ & $28.202\!\pm\!28.300$ & $36.760\!\pm\!52.349$ \\
\bottomrule
\end{tabular}

\label{tab:slcp-mode-fidelity}
\end{table}

\begin{table}[htbp]
\centering
\caption{Phase-1 \textbf{mode discovery} on SLCP (baseline prior) vs.\ the simulation budget (capped), mean\,$\pm$\,std over 10 observations. Higher is better.  Samples drawn from the prior score $0.23$.}
\begin{tabular}{lccc}
\toprule
Surrogate & $10^{3}$ & $10^{4}$ & $10^{5}$ \\
\midrule
Gaussian-NLL & $0.287\!\pm\!0.241$ & $0.338\!\pm\!0.298$ & $0.377\!\pm\!0.332$ \\
Energy distance & $0.173\!\pm\!0.172$ & $0.358\!\pm\!0.266$ & $0.389\!\pm\!0.273$ \\
MMD & $0.164\!\pm\!0.163$ & $0.224\!\pm\!0.184$ & $0.241\!\pm\!0.183$ \\
\bottomrule
\end{tabular}
\label{tab:slcp-mode-discovery}
\end{table}

\FloatBarrier

\paragraph{Bernoulli GLM.}
The surrogates differ most on the Bernoulli GLM (Table~\ref{tab:bernoulli_glm-c2st-vs-ref}).
The energy distance is the only surrogate whose C2ST improves steadily with the budget,
reaching $0.687$ at $10^5$ simulations, and it also attains the highest mode fidelity,
$F \approx 0.77$ (Table~\ref{tab:bernoulli_glm-mode-fidelity}). The Gaussian-NLL stays
at a C2ST of $0.97$--$0.99$ with lower fidelity. MMD fails to concentrate on the posterior
at all ($\mathrm{C2ST} = 1.000$, $F = 0$).

\begin{table}[htbp]
\centering
\caption{Phase-1 \textbf{C2ST} on Bernoulli GLM (baseline prior) vs.\ the simulation budget (capped), mean\,$\pm$\,std over 10 observations. Lower is better ($0.5$ = indistinguishable from the reference).}
\label{tab:bernoulli_glm-c2st-vs-ref}
\begin{tabular}{lccc}
\toprule
Surrogate & $10^{3}$ & $10^{4}$ & $10^{5}$ \\
\midrule
Gaussian-NLL & $0.969\!\pm\!0.022$ & $0.983\!\pm\!0.011$ & $0.988\!\pm\!0.010$ \\
Energy distance & $0.810\!\pm\!0.066$ & $0.725\!\pm\!0.069$ & $0.687\!\pm\!0.059$ \\
MMD & $1.000\!\pm\!0.000$ & $1.000\!\pm\!0.000$ & $1.000\!\pm\!0.000$ \\
\bottomrule
\end{tabular}

\end{table}

\begin{table}[htbp]
\centering
\caption{Phase-1 \textbf{mode fidelity} on Bernoulli GLM (baseline prior) vs.\ the simulation budget (capped), mean\,$\pm$\,std over 10 observations. Values $\times1000$.}
\begin{tabular}{lccc}
\toprule
Surrogate & $10^{3}$ & $10^{4}$ & $10^{5}$ \\
\midrule
Gaussian-NLL & $86.070\!\pm\!84.689$ & $231.940\!\pm\!129.145$ & $184.920\!\pm\!175.774$ \\
Energy distance & $430.270\!\pm\!200.013$ & $737.610\!\pm\!71.645$ & $767.610\!\pm\!54.824$ \\
MMD & $0.000\!\pm\!0.000$ & $0.000\!\pm\!0.000$ & $0.000\!\pm\!0.000$ \\
\bottomrule
\end{tabular}

\label{tab:bernoulli_glm-mode-fidelity}
\end{table}

\FloatBarrier

\subsection{Surrogate bandwidth ablations}
\label{app:bandwidth-ablations}

Each surrogate has a bandwidth $\epsilon$, namely the per-dimension variance of the Gaussian-NLL
(Appendix~\ref{app:surrogate:nll}), the length scale of the energy distance (Appendix~\ref{app:surrogate:ed})
and the kernel bandwidth of the MMD (Appendix~\ref{app:surrogate:mmd}). Each can be fixed or set adaptively
from the simulations, as defined in those appendices. The benchmark runs use the adaptive variance for the
Gaussian-NLL on SLCP and Lotka--Volterra and the adaptive length scale for the energy distance on Two Moons. They use a fixed energy-distance length scale
of $0.914$ on the Bernoulli-GLM, selected from the heuristic procedure of Appendix~\ref{sec:energy_landscape_opt}, and  $\epsilon=1$ for the energy distance on \texttt{gw\_ringdown}. MMD is not used in the benchmark, and its
adaptive variant is the median heuristic.

To measure how sensitive Phase~1 is to $\epsilon$, we rerun Phase~1 on observation~1 with each surrogate's
ablation settings of Appendix~\ref{app:surrogate-ablations} (temperature, replicates and metadynamics) and
change only $\epsilon$, which we sweep over a log-spaced grid of fixed values. The temperature stays at the
value chosen for the ablation bandwidth, so the sweep isolates $\epsilon$. The budget is
$N_\mathrm{sim} = 10^4$ on SLCP and the Bernoulli GLM, while on Two Moons Phase~1 runs its total benchmark schedule
($4{,}512$ simulations). The grid contains the fixed ablation bandwidth of
Appendix~\ref{app:surrogate-ablations} and, where the benchmark fixes one, the benchmark value ($0.914$ for the
Bernoulli energy distance). On Two Moons it also contains $0.095$, the value at which the benchmark's adaptive
length scale settles once Phase~1 is near the posterior. For the energy distance and a fixed-variance
Gaussian-NLL, $\epsilon$ only rescales the surrogate, so changing it is equivalent to changing the temperature
of the likelihood term. For MMD it also changes the shape of the energy. 

For each task we plot C2ST and mode fidelity $F$ (Appendix~\ref{app:surrogate-ablations}) against $\epsilon$
and summarise the sweep in two tables. The C2ST table compares the profile with the best fixed bandwidth
$\epsilon^\star$, and shows how much C2ST degrades when $\epsilon^\star$ is off by a factor of ten in either
direction. The fidelity table gives the band of $\epsilon$ over which $F > 0.01$, the bandwidth
$\epsilon^{\mathrm{peak}}$ that maximises $F$, with the precision $P$ and worst-mode recall $\min_m R_m$
there, and $F$ for the profile. Outside this band fidelity collapses in one of two ways. A bandwidth that
is too small loses modes ($\min_m R_m \to 0$), and one that is too large spreads mass over the prior
($P \to 0$). On SLCP a third table (Table~\ref{tab:slcp-eps-discovery}) reports mode discovery. We do not consider mode discovery for two moons since all surrogates recover the modes significantly.

\begin{figure}[htbp]
\centering
\includegraphics[width=\linewidth]{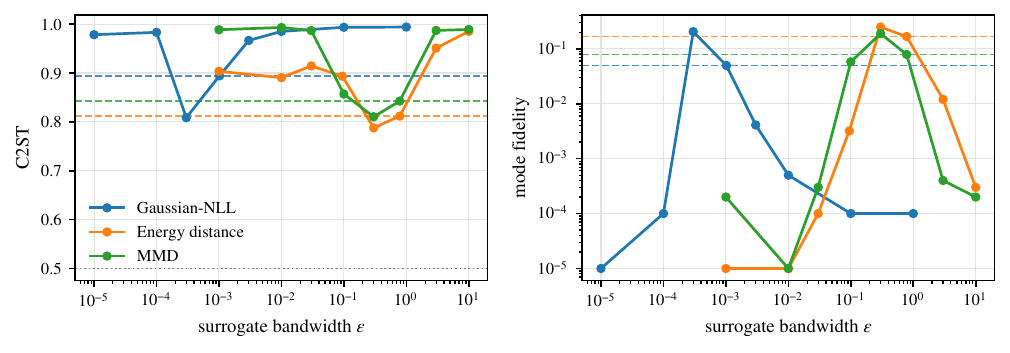}
\caption{Phase-1 sensitivity to the surrogate bandwidth $\epsilon$ on Two Moons (4,512 simulations, one observation). Left: C2ST against the reference (lower is better; dotted: $0.5$). Right: mode fidelity ($=$ precision $\times$ worst-mode recall, log scale; zero drawn at $10^{-5}$). Solid: each surrogate swept over its own bandwidth parameter (Gaussian-NLL variance, energy-distance length scale, MMD kernel bandwidth); dashed: the surrogate's profile setting.}
\label{fig:two_moons-eps}
\end{figure}

\paragraph{Two Moons.}
Misspecifying $\epsilon^\star$ by an order of magnitude degrades the Phase-1 C2ST by
$0.13$--$0.18$, depending on the surrogate (Fig.~\ref{fig:two_moons-eps},
Table~\ref{tab:two_moons-eps-c2st}). The profile bandwidths, chosen without the reference,
lie within half a decade of $\epsilon^\star$ and are within $0.09$ of its C2ST ($0.894$
vs.\ $0.809$ for the Gaussian-NLL, $0.812$ vs.\ $0.788$ for the energy distance and $0.843$
vs.\ $0.810$ for MMD). Mode fidelity is non-trivial only in a narrow band of $\epsilon$
(Table~\ref{tab:two_moons-eps-fidelity}), yet every profile lies inside it ($F = 0.050$,
$0.167$ and $0.079$, against peaks of $0.19$--$0.25$).

\begin{table}[htbp]
\centering
\caption{Phase-1 sensitivity to the surrogate bandwidth $\epsilon$ on Two Moons (4,512 simulations, one observation). \emph{profile}: C2ST at each surrogate's bandwidth and temperature as chosen, without the reference, from the energy-landscape diagnostics. $\epsilon^\star$: the best fixed bandwidth on the sweep (identified using the reference), with its C2ST, and the C2ST when $\epsilon^\star$ is mis-set by one order of magnitude in either direction (change relative to $\epsilon^\star$ in brackets). Lower is better ($0.5$ = indistinguishable from the reference).}
\begin{tabular}{lccccc}
\toprule
Surrogate & profile & $\epsilon^\star$ & C2ST$(\epsilon^\star)$ & C2ST$(0.1\,\epsilon^\star)$ & C2ST$(10\,\epsilon^\star)$ \\
\midrule
Gaussian-NLL & 0.894 & 0.0003 & 0.809 & 0.978\,($+0.170$) & 0.967\,($+0.158$) \\
Energy distance & 0.812 & 0.3 & 0.788 & 0.914\,($+0.127$) & 0.951\,($+0.163$) \\
MMD & 0.843 & 0.3 & 0.810 & 0.987\,($+0.176$) & 0.987\,($+0.177$) \\
\bottomrule
\end{tabular}

\label{tab:two_moons-eps-c2st}
\end{table}

\begin{table}[htbp]
\centering
\caption{Phase-1 mode fidelity ($=$ precision $\times$ worst-mode recall; higher is better) versus the surrogate bandwidth $\epsilon$ on Two Moons (4,512 simulations, one observation). Non-trivial band: the range of $\epsilon$ with mode fidelity above $0.01$ (-- if none); $\epsilon^{\mathrm{peak}}$: the bandwidth with the highest mode fidelity, with its precision and worst-mode recall; \emph{profile}: mode fidelity at the surrogate's profile setting.}
\begin{tabular}{lcccccc}
\toprule
Surrogate & non-trvial band & $\epsilon^{\mathrm{peak}}$ & fidelity & precision & min-recall & profile \\
\midrule
Gaussian-NLL & $[0.0003,\,0.00102]$ & 0.0003 & 0.204 & 0.308 & 0.664 & 0.050 \\
Energy distance & $[0.3,\,3]$ & 0.3 & 0.248 & 0.411 & 0.604 & 0.167 \\
MMD & $[0.1,\,0.783]$ & 0.3 & 0.189 & 0.371 & 0.509 & 0.079 \\
\bottomrule
\end{tabular}

\label{tab:two_moons-eps-fidelity}
\end{table}

\FloatBarrier

\begin{figure}[htbp]
\centering
\includegraphics[width=\linewidth]{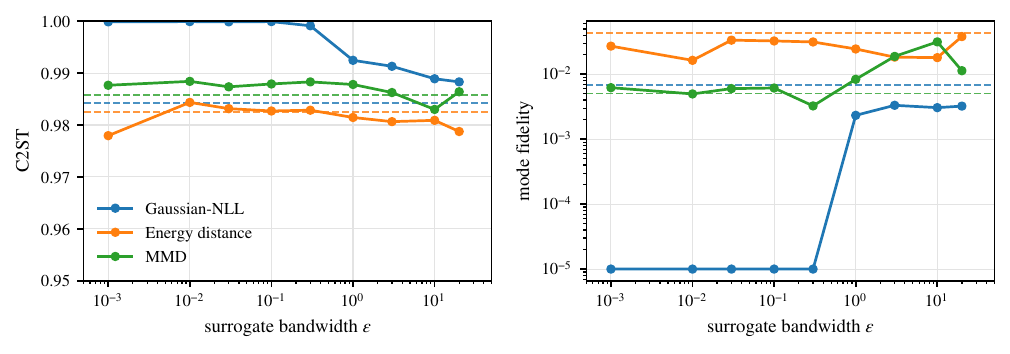}
\caption{Phase-1 sensitivity to the surrogate bandwidth $\epsilon$ on SLCP (10,000 simulations, one observation). Left: C2ST against the reference (lower is better; dotted: $0.5$). Right: mode fidelity ($=$ precision $\times$ worst-mode recall, log scale; zero drawn at $10^{-5}$). Solid: each surrogate swept over its own bandwidth parameter (Gaussian-NLL variance, energy-distance length scale, MMD kernel bandwidth); dashed: the surrogate's profile setting. Mode fidelity is computed on $(\theta_3, \theta_4)$ only, the coordinates that separate the four SLCP modes (and carry the metadynamics bias).}
\label{fig:slcp-eps}
\end{figure}

\paragraph{SLCP.}
Phase-1 C2ST on SLCP is almost insensitive to $\epsilon$: all values lie between
$0.978$ and $1.000$ over four orders of magnitude, and $\epsilon^\star$ falls at the
edge of the grid for the Gaussian-NLL and the energy distance (Fig.~\ref{fig:slcp-eps},
Table~\ref{tab:slcp-eps-c2st}). The profiles, which use adaptive bandwidths on SLCP, are
within $0.005$ of the best fixed value, and better than it for the Gaussian-NLL. For the
Gaussian-NLL a fixed variance discards information: $\theta_3$--$\theta_5$ enter SLCP only through
the noise covariance, which the adaptive per-$\theta$ variance captures and a fixed one does not.
Accordingly, its mode fidelity on $(\theta_3, \theta_4)$ stays at or below $0.003$ for every fixed
variance, against $0.007$ for the adaptive profile (Table~\ref{tab:slcp-eps-fidelity}). The energy
distance keeps $F > 0.01$ over the whole grid (peak $0.038$) and MMD for $\epsilon \geq 3$ (peak
$0.032$). Mode discovery  is shown in Table~\ref{tab:slcp-eps-discovery} and shows the cost of fixing the Gaussian-NLL variance: the adaptive profile reaches
$0.23$, whereas no fixed variance exceeds the prior ($\leq 0.14$), and variances up to $0.1$ miss modes
entirely. For the energy distance the adaptive profile ($0.38$) matches the best fixed length scale
($0.37$); for MMD the adaptive bandwidth ($0.13$) is below every fixed one ($0.14$--$0.26$). We note from the results of Fig.~\ref{fig:slcp-modes} that the mode discovery and fidelity metrics do not consider distinct isolation of the Phase~1 modes present with the GNLL. 

\begin{table}[htbp]
\centering
\caption{Phase-1 sensitivity to the surrogate bandwidth $\epsilon$ on SLCP (10,000 simulations, one observation). \emph{profile}: C2ST at each surrogate's bandwidth and temperature as chosen, without the reference, from the energy-landscape diagnostics. $\epsilon^\star$: the best fixed bandwidth on the sweep (identified using the reference), with its C2ST, and the C2ST when $\epsilon^\star$ is mis-set by one order of magnitude in either direction (change relative to $\epsilon^\star$ in brackets). Lower is better ($0.5$ = indistinguishable from the reference).}
\begin{tabular}{lccccc}
\toprule
Surrogate & profile & $\epsilon^\star$ & C2ST$(\epsilon^\star)$ & C2ST$(0.1\,\epsilon^\star)$ & C2ST$(10\,\epsilon^\star)$ \\
\midrule
Gaussian-NLL & 0.984 & 20 & 0.988 & 0.991\,($+0.003$) & 0.988\,($+0.000$) \\
Energy distance & 0.983 & 0.001 & 0.978 & 0.978\,($+0.000$) & 0.984\,($+0.006$) \\
MMD & 0.986 & 10 & 0.983 & 0.988\,($+0.005$) & 0.986\,($+0.003$) \\
\bottomrule
\end{tabular}

\label{tab:slcp-eps-c2st}
\end{table}

\begin{table}[htbp]
\centering
\caption{Phase-1 mode fidelity ($=$ precision $\times$ worst-mode recall; higher is better) versus the surrogate bandwidth $\epsilon$ on SLCP (10,000 simulations, one observation). Non-trivial band: the range of $\epsilon$ with mode fidelity above $0.01$ (-- if none); $\epsilon^{\mathrm{peak}}$: the bandwidth with the highest mode fidelity, with its precision and worst-mode recall; \emph{profile}: mode fidelity at the surrogate's profile setting. Mode fidelity is computed on $(\theta_3, \theta_4)$ only, the coordinates that separate the four SLCP modes (and carry the metadynamics bias).}
\begin{tabular}{lcccccc}
\toprule
Surrogate & non-trivial & $\epsilon^{\mathrm{peak}}$ & fidelity & precision & min-recall & profile \\
\midrule
Gaussian-NLL & -- & 3 & 0.003 & 0.027 & 0.124 & 0.007 \\
Energy distance & $[0.001,\,20]$ & 20 & 0.038 & 0.132 & 0.288 & 0.043 \\
MMD & $[3,\,20]$ & 10 & 0.032 & 0.083 & 0.378 & 0.005 \\
\bottomrule
\end{tabular}

\label{tab:slcp-eps-fidelity}
\end{table}
\begin{table}[htbp]
\centering
\caption{Phase-1 mode discovery versus the surrogate bandwidth $\epsilon$ on SLCP (10,000 simulations, one observation). \emph{profile}: discovery at the surrogate's profile setting (adaptive or fixed bandwidth); best fixed $\epsilon$: the swept bandwidth with the highest discovery; range: lowest and highest discovery ov er the fixed bandwidths. Mode discovery: each Phase-1 sample is assigned to the nearest reference-mode centre in $(\theta_3, \theta_4)$ if it lies within $1.5\times$ that mode's 95th-percentile reference radius (at most half the distance between mode centres); discovery $= 4 \times$ the smallest per-mode share ($1$ = all mass split equally over the 4 modes, $0$ = a mode missed). Samples drawn from the prior score $0.16$.}
\begin{tabular}{lcccc}
\toprule
Surrogate & profile & best fixed $\epsilon$ & discovery$(\epsilon)$ & range over fixed $\epsilon$ \\
\midrule
Gaussian-NLL & 0.23 (adaptive) & 10 & 0.14 & $[0.00,\,0.14]$ \\
Energy distance & 0.38 (adaptive) & 0.03 & 0.37 & $[0.11,\,0.37]$ \\
MMD & 0.13 (adaptive) & 10 & 0.26 & $[0.14,\,0.26]$ \\
\bottomrule
\end{tabular}

\label{tab:slcp-eps-discovery}
\end{table}

\FloatBarrier

\begin{figure}[htbp]
\centering
\includegraphics[width=\linewidth]{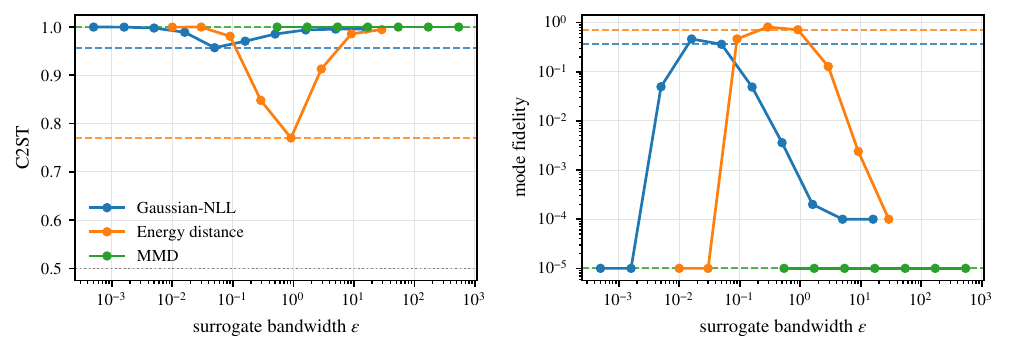}
\caption{Phase-1 sensitivity to the surrogate bandwidth $\epsilon$ on Bernoulli GLM (10,000 simulations, one observation). Left: C2ST against the reference (lower is better; dotted: $0.5$). Right: mode fidelity ($=$ precision $\times$ worst-mode recall, log scale; zero drawn at $10^{-5}$). Solid: each surrogate swept over its own bandwidth parameter (Gaussian-NLL variance, energy-distance length scale, MMD kernel bandwidth); dashed: the surrogate's profile setting.}
\label{fig:bernoulli_glm-eps}
\end{figure}

\paragraph{Bernoulli GLM.}
On the Bernoulli GLM adaptive bandwidths fail (C2ST $\geq 0.995$, $F = 0$ on
observation~1): for the Gaussian-NLL the per-$\theta$ variance estimated from few
replicates of the Bernoulli observations is too noisy. A fixed bandwidth is therefore
required, and it matters (Fig.~\ref{fig:bernoulli_glm-eps},
Table~\ref{tab:bernoulli_glm-eps-c2st}): the profile bandwidths coincide with
$\epsilon^\star$ for both the energy distance (C2ST $0.770$) and the Gaussian-NLL
($0.957$), but the C2ST rises to $0.98$--$1.00$ when they are off by a factor of 10. Both
have a broad band of bandwidths with non-trivial fidelity, peaking at $F = 0.81$ for the
energy distance (Table~\ref{tab:bernoulli_glm-eps-fidelity}). MMD fails at every
bandwidth in the grid. The fixed bandwidth that was used for the benchmarking in the main results was derived from the profile value and corresponds to the best performing bandwidth in this ablation. 

\begin{table}[htbp]
\centering
\caption{Phase-1 sensitivity to the surrogate bandwidth $\epsilon$ on Bernoulli GLM (10,000 simulations, one observation). \emph{profile}: C2ST at each surrogate's bandwidth and temperature as chosen, without the reference, from the energy-landscape diagnostics. $\epsilon^\star$: the best fixed bandwidth on the sweep (identified using the reference), with its C2ST, and the C2ST when $\epsilon^\star$ is mis-set by one order of magnitude in either direction (change relative to $\epsilon^\star$ in brackets). Lower is better ($0.5$ = indistinguishable from the reference).}
\begin{tabular}{lccccc}
\toprule
Surrogate & profile & $\epsilon^\star$ & C2ST$(\epsilon^\star)$ & C2ST$(0.1\,\epsilon^\star)$ & C2ST$(10\,\epsilon^\star)$ \\
\midrule
Gaussian-NLL & 0.957 & 0.05 & 0.957 & 0.998\,($+0.041$) & 0.985\,($+0.028$) \\
Energy distance & 0.770 & 0.914 & 0.770 & 0.981\,($+0.211$) & 0.986\,($+0.216$) \\
MMD & 1.000 & 0.54 & 1.000 & 1.000\,($+0.000$) & 1.000\,($+0.000$) \\
\bottomrule
\end{tabular}

\label{tab:bernoulli_glm-eps-c2st}
\end{table}

\begin{table}[htbp]
\centering
\caption{Phase-1 mode fidelity ($=$ precision $\times$ worst-mode recall; higher is better) versus the surrogate bandwidth $\epsilon$ on Bernoulli GLM (10,000 simulations, one observation). Non-trivial band: the range of $\epsilon$ with mode fidelity above $0.01$ (-- if none); $\epsilon^{\mathrm{peak}}$: the bandwidth with the highest mode fidelity, with its precision and worst-mode recall; \emph{profile}: mode fidelity at the surrogate's profile setting.}
\label{tab:bernoulli_glm-eps-fidelity}
\begin{tabular}{lcccccc}
\toprule
Surrogate & non-trivial band & $\epsilon^{\mathrm{peak}}$ & fidelity & precision & min-recall & profile \\
\midrule
Gaussian-NLL & $[0.005,\,0.16]$ & 0.016 & 0.466 & 0.984 & 0.473 & 0.363 \\
Energy distance & $[0.09,\,2.9]$ & 0.29 & 0.813 & 0.972 & 0.837 & 0.716 \\
MMD & -- & 0.54 & 0.000 & 0.000 & 0.000 & 0.000 \\
\bottomrule
\end{tabular}

\end{table}

\FloatBarrier

\section{Task details}
\label{app:task_details}
This appendix comprises details on computational expense as well as descriptive summaries of each experiemnt conducted in this paper. For the experiments, RSBI required differentiable versions of each simulator taken from the \texttt{sbibm} package. At the true parameters of observations 1 and 5, two-sample Kolmogorov--Smirnov tests on every output dimension, at the 5\% level with Bonferroni correction, find no difference between the original \texttt{sbibm} simulators and our re-implementations for any task. Means agree to within $2.3$ standard errors and standard deviations to within $6\%$, consistent with sampling noise at the sample sizes used ($1.5\times10^3$ to $4\times10^4$ draws).

\subsection{Compute Usage}
\label{app:compute}
Nearly all experiments in \secref{sec:experiments} are run on a single NVIDIA A100 40G GPU, with some run instead on an NVIDIA GTX 1660 Ti laptop GPU.
The time taken for each \M run varies depending on the number of simulations and specific hyperparameters. 
In general, each \M run on the Two Moons simulator using the NVIDIA A100 40G GPU in the $10^{5}$ simulation budget regime takes $\sim25$ minutes to run.
Total compute time on this specific experiment therefore approaches $\sim4$ hours.

\subsection{GW ringdown}
\label{app:gw_task}
Gravitational-wave ringdown is the final stage of a binary black-hole
merger~\citep{2023PhRvD.108d4029C}, occurring after the two black holes
merge and the newly formed remnant settles into a single spinning black
hole. During this phase the signal is dominated by a superposition of
damped sinusoids, the quasi-normal modes, whose frequencies and decay
times are determined entirely by the remnant's mass and spin according
to general relativity. This makes ringdown a uniquely clean probe of
strong-field gravity: measuring these modes constrains the remnant
properties and tests general relativity in the high-curvature
regime~\citep{2018CQGra..35a4002G,Bhagwat:2019dtm}.

Our differentiable ringdown simulator combines two quasi-normal modes
($\ell = m = 2$, $n \in \{0, 1\}$) with the H1 antenna pattern at the
LIGO Hanford detector and additive Gaussian noise drawn from the H1 power
spectral density estimated from LIGO first-observing-run data around
GW150914. The strain is whitened via the Cholesky factor of the Toeplitz
noise covariance built from the autocorrelation implied by this PSD, giving a
unit-variance representation that sets the natural surrogate length scale $\epsilon = 1$.
The analysis window is $T = 0.2\,\text{s}$
at a sampling rate of $4096\,\text{Hz}$, giving $d_x = 820$ samples per
detector. The full
forward map, from waveform generation through antenna projection, noise injection
and whitening, is implemented in PyTorch so that
$\nabla_\theta s(\theta, \omega)$ is available end-to-end, which is
what makes Phase~1's surrogate gradient and Phase~2's adjoint matching
applicable. The noise enters as $L z$ with $z \sim \mathcal{N}(0, I)$ and $L$ the
Cholesky factor, so the simulator is reparameterised exactly. Unlike the other tasks,
this simulator is not a re-implementation. The same class generates the observations,
the reference posteriors, the simulations of the baselines and those of RSBI, with the same
detector, window and sampling settings.

We infer
$\theta = (M, \chi, A_{220}, \varphi_{220}, A_{221}, \varphi_{221},
\cos\iota, \alpha, \cos\delta, \psi)$ with $d_\theta = 10$ in total. For
the experiments reported in the main text the four extrinsic dimensions
$(\cos\iota, \alpha, \cos\delta, \psi)$ are pinned to truth, the
natural conditional setup for a single-detector analysis with known sky
position and distance, reducing the inferred dimension to
$d_\theta = 6$. Two prior scenarios are considered. The \emph{default}
scenario with $M \in [40, 100]\,M_\odot$, $\chi \in [0, 0.99]$,
$A_{220}, A_{221} \in [0, 5]$, $\varphi_{220}, \varphi_{221} \in
[0, 2\pi]$, $\cos\iota, \cos\delta \in [-1, 1]$, $\alpha \in [0, 2\pi]$,
$\psi \in [0, \pi]$ covers astrophysically plausible stellar-mass
remnants. The \emph{wide} scenario with $M \in [10, 200]\,M_\odot$ and
$A_{220}, A_{221} \in [0, 15]$, with the remaining ranges unchanged,
relaxes the astrophysical priors to those a fundamental-physics
analysis might use when remaining agnostic about the source population.
The two scenarios give $\log_{10}\mathcal{R}_p \approx 2.6$ and $2.8$
respectively (Table~\ref{tab:prior_scenarios}).

Because the noise model is exact in whitened coordinates, the likelihood
is analytic given $\theta$ and we can sample a reference posterior
directly. Ten injections are drawn uniformly from a sub-box of the default prior,
with $M \in [60, 80]\,M_\odot$, $A_{220} \in [3, 5]$ and $A_{221} \in [2, 4]$ and the
remaining parameters over their full default ranges, and one noise realisation is
added per injection to produce $x_o$. The
same ten observations are used across both prior scenarios, since widening
the prior does not regenerate the data, only the reference posterior.
Figure~\ref{fig:gw_corner} shows observation~10, whose injected source is
$M = 70.18\,M_\odot$, $\chi = 0.558$, $A_{220} = 4.249$, $\varphi_{220} = 5.873\,\text{rad}$,
$A_{221} = 2.724$ and $\varphi_{221} = 1.099\,\text{rad}$.
We draw $N_\text{ref} = 10^4$ reference samples per observation. The
\texttt{emcee} runs are computed once per (prior, observation) pair
and reused across method runs.

We run RSBI at a single simulation budget of $N_\mathrm{sim} = 10^5$ with $R = 2$ Phase-2 rounds.
Phase~1 uses $5.99\times10^4$ simulations, which seed the first NRE (topped up to $6\times10^4$), and
the second round adds $4\times10^4$ simulations drawn from the round-0 Phase~2 sampler. Phase~1's
energy is the unbiased energy distance of Appendix~\ref{app:surrogate:ed} with a fixed length scale
$\epsilon = 1$ and $K = 4$ replicates. The source is the prior box with a soft boundary
penalty ($\lambda_\mathrm{soft} = 10$), which keeps the bridge from spending capacity on the prior
corners. Well-tempered metadynamics acts in Phase~1 along $\xi(\theta) = (M, \chi)$ to cover the
mass and spin basins, and Phase~2 starts from the Phase~1 samples reweighted by the final bias. Phase~2
runs without a bias, in coordinates whitened by the Phase~1 samples (Appendix~\ref{app:phase2-precondition}).
Both prior scenarios share every setting, and the sampler operates on the prior box rescaled to
$[0,10]^6$. Inference is on the 6D free subspace, with all MLP input dimensions remapped from the
on-disk $d_\theta = 10$ to 6. The full configuration is given in Sec.~\ref{sec:hyperparam_tables}.

\subsection{SLCP}
\label{app:slcp-hparams}

We evaluate \M on the SLCP task of~\citet{pmlr-v130-lueckmann21a}
($\theta \in \mathbb{R}^5$, $x \in \mathbb{R}^8$) across the three prior
boxes \textsc{default}~$=[-3,3]^5$, \textsc{wide5x}~$=[-5,5]^5$, and
\textsc{wide10x}~$=[-10,10]^5$ (see Tab.~\ref{tab:prior_scenarios} for the corresponding prior-to-posterior
volume ratios), each at simulator-call budgets $N_\mathrm{sim} \in \{10^3, 10^4, 10^5\}$, on
observations $1{-}10$ from the \texttt{sbibm} reference set. Posterior quality is reported via the
C2ST against the reference posterior of each prior, evaluated on $10^4$ posterior samples.

The simulator draws four i.i.d.\ points $x_j \sim \mathcal{N}(m_\theta, S_\theta)$, with
$m_\theta = (\theta_1, \theta_2)$ and $S_\theta$ built from $s_1 = \theta_3^2$, $s_2 = \theta_4^2$ and
$\rho = \tanh\theta_5$ as in \texttt{sbibm}. We re-implement it in PyTorch through the
reparameterisation $x_j = m_\theta + L_\theta z_j$, $z_j \sim \mathcal{N}(0, I_2)$, where $L_\theta$ is
the Cholesky factor of $S_\theta$, computed in closed form for the $2 \times 2$ case with a diagonal
jitter of $10^{-6}$, so that $\nabla_\theta x$ is available by automatic differentiation. The baselines
use the original \texttt{sbibm} simulator. The two implementations agree everywhere except where
$\tanh\theta_5$ saturates in single precision ($|\theta_5| \gtrsim 8.3$, reachable only under
\textsc{wide10x}). There \texttt{sbibm} clamps $\rho$ to $\pm 0.9999$, whereas we floor the diagonal
of $L_\theta$ at $10^{-3}$, and both only serve to keep the covariance positive definite.

The controlled-SDE drift $u_\phi(\theta, t)$ and the corrector head are Fourier-feature MLPs with
sinusoidal time embedding and GELU activations, $6 \times 64$ (depth $\times$ width) in Phase~1 and
$4 \times 128$ in Phase~2. The NRE classifier projects $\theta$ and $x$ separately to $32$ features
each, concatenates them and applies a $6 \times 64$ MLP with SiLU activations. Both phases integrate
the controlled SDE on $[0, 1]$ with an EDM-style noise schedule~\citep{karras2022edm} ($\rho = 7$),
using $64$ Euler--Maruyama steps in Phase~1 and $256$ in Phase~2.

Phase~1 fits a controlled SDE that approximately samples the surrogate posterior under the
Gaussian-NLL surrogate distance $\D^\mathrm{NLL}(s(\theta), x_o)$ of Section~\ref{app:surrogate:nll},
with the per-dimension variance estimated adaptively from $K$ replicates of the
differentiable simulator at each evaluation.

Phase~2 runs $R$ rounds, $R = 1$ at $N_\mathrm{sim} = 10^3$ and $R = 2$ at $10^4$ and $10^5$. The
first NRE is trained on the Phase~1 simulations, topped up with draws from the Phase~1 sampler
($10^3$, $5\times10^3$ and $1.4\times10^4$ simulations in total), and the second round adds the
remaining budget ($5\times10^3$ and $8.6\times10^4$) from the previous round's sampler. Each round
refines the controlled SDE, warm-started from the previous round and started from the Phase~1
sampler, against $E_r(\theta) = -\log p(\theta) - \log r_{\psi_r}(\theta, x_o)$. Well-tempered
metadynamics~\citep{PhysRevLett.100.020603, nam2026enhancing} biases the SDE along the collective
variable $\xi(\theta) = (\theta_3, \theta_4)$ to encourage exploration of the four SLCP modes. The bias
is reset to zero on the final round, so that the sampled distribution is unbiased up to importance
reweighting (Section~\ref{sec:stage2}). Hyperparameters are listed in Sec.~\ref{sec:hyperparam_tables}.

\subsection{Two Moons}
\label{app:two_moons-hparams}
We evaluate \M on the Two Moons task of~\citet{pmlr-v130-lueckmann21a}
($\theta \in \mathbb{R}^2$, $x \in \mathbb{R}^2$) across the three prior boxes
\textsc{default}~$=[-1,1]^2$, \textsc{wide10x}~$=[-10,10]^2$, and
\textsc{wide50x}~$=[-50,50]^2$ (see Table~\ref{tab:prior_scenarios} for the
corresponding prior-to-posterior volume ratios), each at simulator-call
budgets $N_\mathrm{sim} \in \{10^3, 10^4, 10^5\}$, on observations
$1{-}10$ from the sbibm reference set. Posterior quality is reported via
the C2ST against the closed-form sbibm reference posterior, sampled by
rejection and evaluated on $10^4$ posterior samples.

The controlled-SDE drift $u_\phi(\theta, t)$ and the corrector head are
Fourier-feature MLPs with sinusoidal time embedding and GELU activations,
$4 \times 128$ (depth $\times$ width) in Phase~1 and $6 \times 64$ in
Phase~2. The NRE classifier projects $\theta$ and $x$ separately to half its
width each, concatenates them and applies a $6$-layer MLP with SiLU
activations, of width $64$ to $256$ depending on the cell. Phase~1 integrates
its controlled SDE on $[0, 1]$ with a Brownian schedule on the \textsc{default}
prior and an EDM-style schedule~\citep{karras2022edm} on the wide priors, over
$128$ Euler--Maruyama steps ($256$ for \textsc{wide50x}). Phase~2 uses an
EDM-style schedule over $256$ steps throughout.

The Two Moons simulator
\[
    s(\theta;\, a, r) \;=\; \Bigl(\,
        r\cos a + 0.25 - \tfrac{|\theta_1 + \theta_2|}{\sqrt{2}},\;\;
        r\sin a + \tfrac{-\theta_1 + \theta_2}{\sqrt{2}}\,
    \Bigr)
\]
is driven by the random angle $a$ and radius $r$.
We re-implement it in PyTorch with the noise drawn independently of $\theta$,
$a = \pi(u - \tfrac12)$ with $u \sim \mathcal{U}(0,1)$ and $r = 0.1 + 0.01\,z$ with $z \sim \mathcal{N}(0,1)$,
as in \texttt{sbibm}, so that $s$ is differentiable in $\theta$ away from the line $\theta_1 = -\theta_2$,
where automatic differentiation returns a subgradient of $|\theta_1 + \theta_2|$. This implementation
replaces the \texttt{sbibm} simulator for all methods, so \M and the baselines use identical code.
Phase~1 uses the energy-distance
surrogate $\D^\mathrm{ED}$ of Section~\ref{app:surrogate:ed} with an adaptive
length scale and $K = 1$ replicate per parameter.

Phase~2 runs $R = 2$ rounds in every cell. The first NRE is trained on the Phase~1
simulations, topped up with draws from the Phase~1 sampler, and the second round
simulates the rest of the budget from the round-1 sampler, with
$600 + 400$ simulations at $N_\mathrm{sim} = 10^3$, and $5{,}000 + 5{,}000$ and
$55{,}000 + 45{,}000$ on the \textsc{default} prior at $10^4$ and $10^5$
($1{,}000 + 9{,}000$ and $5{,}000 + 95{,}000$ on the wide priors). Each round
refines the warm-started controlled SDE against
$E_r(\theta) = -\log p(\theta) - \log r_{\psi_r}(\theta, x_o)$. Well-tempered
metadynamics is used in Phase~2 on the wide priors and on the \textsc{default}
prior at $10^3$ simulations, and in Phase~1 on the \textsc{default} prior at $10^4$
and $10^5$ simulations. It is off elsewhere. All Phase~1, Phase~2 and NRE
hyperparameters, which differ between prior boxes and budgets, are listed in
Sec.~\ref{sec:hyperparam_tables}.

\subsection{Bernoulli GLM}
\label{app:bernoulli-hparams}

We evaluate \M on the Bernoulli GLM task of~\citet{pmlr-v130-lueckmann21a}
($\theta \in \mathbb{R}^{10}$, $x \in \mathbb{R}^{10}$), a generalised linear model in which the spike
train of a neuron, driven by a white-noise stimulus over $100$ time bins, is reduced to ten sufficient
statistics. The prior is Gaussian, $\theta \sim \mathcal{N}(0, B^{-1})$, with the smoothness-inducing
precision $B$ of \texttt{sbibm}. We widen it by scaling the precision by $1$, $0.04$ and $0.01$, which
multiplies the prior standard deviations by $1$, $5$ and $10$ (\textsc{default}, \textsc{medium} and
\textsc{large}; see Tab.~\ref{tab:prior_scenarios} for the corresponding prior-to-posterior volume
ratios). Each prior is run at simulator-call budgets $N_\mathrm{sim} \in \{10^3, 10^4, 10^5\}$ on
observations $1{-}10$ from the \texttt{sbibm} reference set, and posterior quality is reported via the
C2ST against the reference posterior of each prior on $10^4$ posterior samples.

We re-implement the \texttt{sbibm} simulator in PyTorch, batched over $\theta$, with the same design
matrix $X \in \mathbb{R}^{100 \times 10}$ and stimulus. Each spike is drawn as
$y_t \sim \mathrm{Bernoulli}(\sigma(\psi_t))$ with $\psi = X\theta$, and the summary statistics, the
spike count and a nine-bin spike-triggered average, are computed exactly as in \texttt{sbibm}, which the
baselines use. Binary spikes have no reparameterisation gradient. For Phase~1 we therefore
differentiate through a straight-through Gumbel-softmax relaxation~\citep{jang2017categorical} of the
spike draws (temperature $0.5$). The forward pass returns the hard Gumbel-max sample with logits
$(0, \psi_t)$, which is an exact Bernoulli draw, and the backward pass uses the gradient of the
continuous relaxation, which is biased. All simulations, including those reused from Phase~1, are
therefore exact, and the NRE is trained on the exact simulator.

The controlled-SDE drift $u_\phi(\theta, t)$ and the corrector head are Fourier-feature MLPs with
sinusoidal time embedding and GELU activations, $6 \times 128$ (depth $\times$ width) in both phases.
The NRE classifier projects $\theta$ and $x$ separately to $64$ features each, concatenates them and
applies a $6 \times 128$ MLP with SiLU activations. Both phases integrate the controlled SDE on
$[0, 1]$ with $128$ Euler--Maruyama steps and an EDM-style noise schedule~\citep{karras2022edm}, with
$\rho = 5$ in Phase~1 and $\rho = 7$ in Phase~2.

Phase~1 fits a controlled SDE that approximately samples the surrogate posterior under the
energy-distance surrogate of Section~\ref{app:surrogate:ed}, with a fixed length scale
$\epsilon = 0.914$ and $K = 2$ replicates of the differentiable simulator per parameter.
Because the posterior is unimodal, no well-tempered bias is used in either phase. Phase~2 runs $R = 2$
rounds. The first NRE is trained on $25\%$ of the budget, the Phase~1 simulations (topped up with draws
from the Phase~1 sampler at $N_\mathrm{sim} = 10^5$), and the remaining $75\%$ is simulated in the
second round from the round-1 sampler. Each round refines the warm-started controlled SDE against
$E_r(\theta) = -\log p(\theta) - \log r_{\psi_r}(\theta, x_o)$. The reference noise scale
$\sigma_{\max}$ grows with the prior width. All hyperparameters are listed in
Sec.~\ref{sec:hyperparam_tables}.

\subsection{Lotka--Volterra}
\label{sec:lokta-volterra}
Lotka--Volterra is the classic predator--prey model and a standard SBI benchmark task
\citep{pmlr-v130-lueckmann21a}. Four rate parameters $\theta = (\alpha, \beta, \gamma, \delta)$ govern the
oscillating populations of a prey and a predator species, and the goal is to infer them from a short, noisy record
of both populations.

The simulator integrates $\dot u = f(u,\theta)$, $f(u,\theta) = (\alpha u_1 - \beta u_1 u_2,\ -\gamma u_2 + \delta u_1 u_2)$,
from $u(0) = (30, 1)$ and records both species at $K = 10$ times $t_k = k\Delta$, $\Delta = 2.1$, with multiplicative
noise $x_k = u(t_k) \odot \exp(\sigma \xi_k)$, $\xi_k \sim \mathcal{N}(0, I_2)$, $\sigma = 0.1$, giving
$x \in \mathbb{R}^{2K}$. A differentiable fourth-order Runge--Kutta (RK4) scheme (step $0.05$) provides
$\nabla_\theta x$ by backpropagating through the unrolled RK4 steps, rather than by solving an adjoint ODE.
\texttt{sbibm} instead solves the same ODE with Julia's adaptive Tsit5 solver, which the baselines use and
which generated the observations for the references. Our PyTorch re-implementation keeps the same initial state, observation
times, output clipping to $[10^{-10}, 10^4]$ and noise model. To check the fixed step, we compared noise-free
outputs with an RK4 solve at step $0.001$, at the true parameters and $200$ reference posterior draws of every
observation. The largest deviation is $2 \times 10^{-4}$ in $\log x$, three orders of magnitude below the
observation noise $\sigma = 0.1$.
For rates
far in the prior tails the solve can overflow, so populations are clamped to $[10^{-10}, 10^{6}]$ after every step,
simulator gradients that are non-finite for a sample are set to zero and the rest are capped at norm $10^{12}$. The
prior term of the energy is differentiated separately, so such draws are still pulled back towards the support.
\M samples $\phi = \log \theta$, under which the LogNormal prior becomes $\mathcal{N}(\mu_0, s^2 I)$,
with $s = 0.5$, $1$ and $2$ for the \textsc{default}, \textsc{medium} and \textsc{large} priors. This is the
same reparameterisation SNVI applies through its automatic support transform. Samples are mapped back with
$\theta = e^{\phi}$ and evaluated in $\theta$-space against the same reference posteriors as all baselines.

The Phase~1 surrogate is the Gaussian negative log-likelihood with adaptive variance (Sec.~\ref{app:surrogate:nll}) at
temperature $T = 1$. Because each feature is scaled by its own variance, no additional feature standardisation (such
as the per-feature z-scoring of the \texttt{sbi} toolkit, \citealp{tejero2020sbi}) is used.

The controlled-SDE drift $u_\phi(\phi, t)$ and the corrector head are Fourier-feature MLPs with sinusoidal
time embedding, $6 \times 128$ (depth $\times$ width) with GELU activations in both phases. The NRE
classifier projects $\phi$ and $x$ separately to $64$ features each, concatenates them and applies a
$6 \times 128$ MLP with SiLU activations, on inputs standardised by the mean and standard deviation of the
first round's training set. Both phases integrate the controlled SDE on $[0,1]$ with an EDM-style noise
schedule~\citep{karras2022edm}, using $64$ Euler--Maruyama steps in Phase~1 and $128$ in Phase~2, with
$\rho = 5$ in Phase~1 ($\rho = 5$ in Phase~2 at $N_\mathrm{sim} = 10^5$ and $7$ otherwise). The reference noise
scale is $\sigma_{\max} = 0.3$ in Phase~1 ($0.5$ for the wider priors at $10^5$) and $\sigma_{\max} = 1$ in Phase~2,
where it acts in the whitened coordinates of Appendix~\ref{app:phase2-precondition}. Phase~1 uses
two simulator replicates per parameter (three at $10^5$) and energy gradients clipped at norm $256$.

At $N_\mathrm{sim} = 10^3$ and $10^4$, Phase~1 receives $70\%$ of the budget and Phase~2 runs a single round
($R = 1$). Its NRE is trained on the Phase~1 simulations drawn after the horizon and shooting curriculum has ended,
so it never sees masked or shooting-based data, topped up to the full budget with draws from the Phase~1 sampler.
At $10^5$, Phase~1 is capped at $5 \times 10^4$ simulations, the first NRE is seeded with its post-curriculum
simulations, and two sequential rounds of $2.5 \times 10^4$ simulations follow ($R = 3$). For the \textsc{large}
prior at $10^5$, Phase~1 instead receives $8 \times 10^4$ simulations and a single round ($R = 1$) is trained on its
post-curriculum simulations topped up to the full budget. At every budget, Phase~2 is refined against the truncated
ratio with $k = 3$ and $q=0.9$, and k = 2 and q = 0.9 for the \textsc{large} prior at $10^5$, without a well-tempered bias since the posterior is unimodal. All
hyperparameters are listed in Sec.~\ref{sec:hyperparam_tables}.

\paragraph{Surrogate likelihood choice}
Away from the posterior, a parameter error changes the period of the predator--prey cycle, so the phase error
accumulates and late observations fall at essentially arbitrary points of a mismatched cycle. Any discrepancy on the
full trajectory then becomes a rippled plateau in $\phi$ whose gradient carries little information about the
direction of the posterior. We therefore use two modifications of the Phase~1 surrogate. Neither changes the data, the
prior, the reference posterior or the evaluation metric.
\emph{(i) Horizon schedule.} For a horizon $H \le K$, the later features of each simulation are replaced by those of
the observation, $[\Pi_H(x)]_k = x_k$ for $k < H$ and $x^o_k$ otherwise, so they contribute exactly zero to the
discrepancy while the data dimension is unchanged. Over a short prefix the phase error cannot wrap around the cycle,
so $E$ is close to convex over a much larger region. With $\nu$ the fraction of the Phase~1 budget consumed,
$H = 3$ for $\nu < 0.3$, $H = 6$ for $0.3 \le \nu < 0.6$ and $H = K$ afterwards (at $N_\mathrm{sim} = 10^5$ the switches
occur at fixed counts of $1.5 \times 10^4$ and $3 \times 10^4$ simulations), so the end of Phase~1 targets the full
trajectory.
\emph{(ii) Multiple shooting.} With $\Phi_\Delta(u,\theta)$ the RK4 flow over one observation interval, each
observation is predicted from the \emph{observed} state one interval earlier,
$x^{\mathrm{sh}}_k = \Phi_\Delta(x^o_{k-1},\theta) \odot e^{\sigma \xi_k}$ for $k \ge 1$ ($x^{\mathrm{sh}}_0 = u(0)
\odot e^{\sigma\xi_0}$). Phase errors cannot accumulate beyond one interval, and each segment needs $42$ RK4 steps
instead of up to $378$. Conditioning on noisy states shifts the minimiser slightly (an errors-in-variables effect), so
shooting is used only for $\nu < 0.5$, and the remaining bias is removed by the ratio-corrected Phase~2.

Both of the above are classical remedies for long-horizon gradients in differential-equation models, whose
sensitivity to $\theta$ grows polynomially for oscillatory and exponentially for chaotic dynamics
\citep{metz2021gradients}. Growing the fitted window is a continuation (curriculum) strategy
\citep{bengio2009curriculum}. Multiple shooting splits the trajectory into short segments started from estimated or
observed states \citep{bock1984multiple,voss2004nonlinear} and was recently shown to rescue neural ODE fitting on
oscillatory data \citep{turan2022multiple}. Smoothing-based estimators avoid long-horizon integration altogether
\citep{ramsay2007parameter}.
Both modifications apply only to the Phase~1 surrogate, the energy that steers the variational sampler towards the
posterior region before Phase~2 corrects it, and they cost no simulations beyond those Phase~1 already uses. They
cannot be applied in the same way to SNPE or SNVI. These methods train a density or ratio estimator on a corpus of
simulated pairs $(\theta, x)$, and that estimator is their final posterior. Masking the late observations of the
corpus, or replacing its simulations with shooting segments, would change the conditional they estimate to
$p(\theta \mid x_{<H})$ or to that of a different simulator, with no later phase to correct it. A horizon curriculum
could at most supply an intermediate proposal, at the cost of training additional estimators on part of the budget,
which neither method does. In \M, by contrast, the NRE and the Phase~2 target only ever see full, unmodified
simulations.
The horizon schedule needs only black-box access to $\theta \mapsto x$. Multiple shooting additionally assumes that
the simulator can be restarted from an arbitrary state and that the full state is observed, both of which hold for
Lotka--Volterra. It uses the simulator's latent process in the spirit of \citet{brehmer2020mining}, the same
white-box access \M already relies on through $\nabla_\theta s(\theta)$.
As a rule, if the pilot gradient field in Sec.~\ref{sec:energy_landscape_opt} is incoherent away from the posterior,
use the horizon schedule. If the state is fully observed and restartable, multiple shooting is a further option. If
the field is already coherent, keep the full-trajectory surrogate.

\subsection{Reference posteriors}
\label{app:references}

We require a reference posterior $p(\theta\!\mid\!x_o)$ for every
$(\text{task}, \text{scenario}, \text{observation})$ triple in the
benchmark grid, since C2ST in Section~\ref{sec:experiments} is computed
against it. The sampler used to obtain these samples differs by task; we
always draw $N_\text{ref} = 10^4$ samples per observation and reuse them
across method runs.

For Two~Moons the likelihood is fully analytic on the polar variables
recovered by \texttt{sbibm}'s
\texttt{TwoMoons.\_likelihood}~\citep{pmlr-v130-lueckmann21a}. The simulator can be therefore inverted in closed form. \texttt{sbibm}'s reference sampler draws the simulator noise, maps it to the parameter that reproduces $x_o$ (choosing between the two branches at random), and rejects only draws outside the prior box. Each prior scenario is redrawn independently with its own box. For eight of the ten observations the posterior lies inside $[-1,1]^2$, so all scenarios share the same reference. For observations 5 and 10 the default box truncates one moon, removing about $57\%$ and $12\%$ of the wide-prior posterior mass, and the wider scenarios recover the full moon.

For SLCP the analytic likelihood induces a four-mode posterior across
which Markov chains mix poorly, so we use \texttt{sbibm}'s shipped
SIR+NSF reference sampler for all three scenarios (\textsc{default},
\textsc{wide5x}, \textsc{wide10x}). On the $20\times$-larger
\textsc{wide10x} box this sampler fails for observations~4 and~10: the
posterior occupies a tiny fraction of the box, the spline-flow proposal
misses parts of it, and rejection sampling stalls at an acceptance rate of
about $10^{-6}$. For these two cases we exploit the closed-form SLCP
likelihood directly. Because it depends on $\theta_3$ and $\theta_4$ only
through $\theta_3^2$ and $\theta_4^2$, the posterior is exactly symmetric
under sign flips of these two coordinates, so we sample the folded posterior
($\theta_3, \theta_4 \geq 0$) and attach independent random signs
afterwards, which gives the four modes exactly equal mass. A short
ensemble-MCMC pilot run~\citep{2013PASP..125..306F} on the folded posterior
locates the posterior mass; a mixture of Student-$t$ distributions fitted to
it, plus a $5\%$ uniform component on the box, serves as an importance
proposal against the exact posterior, from which we resample $10^4$ points.
As a check, the same procedure applied to observation~4 under the
\textsc{default} prior reproduces the \texttt{sbibm} reference
(C2ST $= 0.50$).

For \texttt{gw\_ringdown} the whitened-strain log-likelihood is the
analytic Gaussian
$-\tfrac12 \|L^{-1}(x_o - s(\theta))\|_2^2 + \text{const}$, but the
posterior has pronounced structure---periodic identifications on the
phase dimensions and a near-degeneracy between $A_{220}$ and $A_{221}$
at low signal-to-noise---so we sample via affine-invariant ensemble MCMC
with \texttt{emcee}~\citep{2013PASP..125..306F}: $256$ walkers initialized
in a $0.01$-ball around the injected truth,$10^5$ steps per walker, $10^3$ steps of burn-in, thinning factor 2, and an equal mixture of differential-evolution and stretch moves.
Convergence is diagnosed via per-chain effective sample size
(ESS\,$\gtrsim 100$ per parameter) and trace-plot inspection, rerunning
with longer chains when either flag fires.

For Bernoulli GLM, we use the reference-posterior sampler that ships with
\texttt{sbibm}~\citep{pmlr-v130-lueckmann21a} unchanged, and replace only the
prior with the scenario prior (Table~\ref{tab:prior_scenarios}). The scenarios
scale the precision matrix of the \texttt{sbibm} smoothness prior, so the prior
remains Gaussian and the \texttt{sbibm} sampler applies directly. That sampler
is an exact Gibbs sampler based on P\'olya--Gamma data
augmentation~\citep{polson2013polyagamma}. It is initialised at the true
parameters, discards $25{,}000$ warm-up iterations and keeps every $25$th draw,
giving $10^4$ samples.

For Lotka--Volterra, the default-prior reference is the \texttt{sbibm} reference posterior, obtained with
its ODE solver and its log-normal observation model (noise scale $0.1$) as the likelihood. A slice-sampling
MCMC chain is started at the true parameters and run for $10^4$ warm-up steps. A neural spline
flow is then fitted to its samples and mixed with the prior (weight $0.1$) to form a proposal, and
rejection sampling from this proposal yields $10^4$ exact posterior samples. The wider scenarios only widen
the scale of the log-normal prior, with its location unchanged (Table~\ref{tab:prior_scenarios}). Their
references are obtained by importance-resampling the default reference with weights
$p_s(\theta)/p_{0.5}(\theta)$, the ratio of the scenario prior to the default prior, drawing $10^4$ samples
with replacement. Because every observation's posterior lies well inside the default prior, the weights are
nearly uniform. The effective sample size is at least $86\%$ of $N$ for the medium prior and $78\%$ for the
large prior across the ten observations, and the resampled sets contain about $6{,}000$ to $6{,}400$ distinct
draws.

\subsection{Prior scenarios and volume ratios}
\label{app:prior_volums}

\begin{table}[h]
\caption{Prior scenarios and mean log prior-to-posterior volume ratios, averaged over the ten observations.
Posterior volumes are estimated from the reference samples as the product of per-dimension $95\%$ marginal
widths and as the volume of the sample-covariance ellipsoid. For Gaussian priors, the prior volume is measured
in the same way from its covariance (marginal widths $2 \times 1.96\,\sigma_d$). Lotka--Volterra volumes are
computed in $\log\theta$. $\Sigma_0$ denotes the default \texttt{sbibm} Bernoulli GLM prior covariance and
$\mu_0 = (-0.125, -3, -0.125, -3)$.}
\label{tab:prior_scenarios}
\centering
\small
\begin{tabular}{lllcc}
\toprule
Task & Scenario & Prior &
$\log_{10}\mathcal{R}_p^{\text{marg.}}$ &
$\log_{10}\mathcal{R}_p^{\text{ell.}}$ \\
\midrule
Two moons & default & $\mathcal{U}([-1,1]^2)$    & 0.89 & 1.69 \\
           & wide10x & $\mathcal{U}([-10,10]^2)$  & 2.89 & 3.65 \\
           & wide50x & $\mathcal{U}([-50,50]^2)$  & 4.28 & 5.04 \\
\midrule
SLCP       & default & $\mathcal{U}([-3,3]^5)$    & 1.57 & 3.46 \\
           & wide5x  & $\mathcal{U}([-5,5]^5)$    & 1.93 & 4.05 \\
           & wide10x & $\mathcal{U}([-10,10]^5)$  & 2.69 & 4.99 \\
\midrule
Bernoulli GLM & default & $\mathcal{N}(0, \Sigma_0)$      & 5.38 & 3.55 \\
           & medium  & $\mathcal{N}(0, 25\,\Sigma_0)$  & 9.86 & 8.19 \\
           & large   & $\mathcal{N}(0, 100\,\Sigma_0)$ & 12.18 & 10.68 \\
\midrule
Lotka--Volterra & default & $\log\theta \sim \mathcal{N}(\mu_0, 0.5^2 I)$ & 5.05 & 6.26 \\
           & medium  & $\log\theta \sim \mathcal{N}(\mu_0, I)$       & 6.24 & 7.46 \\
           & large   & $\log\theta \sim \mathcal{N}(\mu_0, 2^2 I)$   & 7.44 & 8.67 \\
\midrule
\texttt{gw\_ringdown} & default & $\mathcal{U}([40,100],[0,0.99],[0,5],[0,2\pi],[0,5],[0,2\pi])$ & 2.62 & 6.41 \\
           & wide & $\mathcal{U}([10,200],[0,0.99],[0,15],[0,2\pi],[0,15],[0,2\pi])$ & 2.84 & 6.73 \\
\bottomrule
\end{tabular}
\end{table}

For each task we consider a default prior and two wider ones (one for \texttt{gw\_ringdown}): uniform boxes
for Two Moons, SLCP and \texttt{gw\_ringdown}, and Gaussian priors with inflated covariance for Bernoulli GLM
and Lotka--Volterra (Table~\ref{tab:prior_scenarios}). For each $(\text{task}, \text{scenario}, \text{obs})$
triple we estimate $\log_{10}\mathcal{R}_p$ from the reference samples using a per-dimension $95\%$-marginal
box estimator and a sample-covariance ellipsoid estimator (see Appendix~\ref{sec:volume}). For Gaussian priors
the prior volume is measured with the same two estimators, from its covariance. The marginal estimator is the
one we report throughout the main text, including the legends of the C2ST-versus-budget figures, because it
is insensitive to multimodality. The ellipsoid estimator, which accounts for correlations but is inflated
when the posterior has several modes (as for SLCP), is given in the table for comparison only. Across tasks
the scenarios span marginal $\log_{10}\mathcal{R}_p$ from about $0.9$ to $12$.

The \texttt{gw\_ringdown} ratio changes little between the two scenarios ($2.62$ and $2.84$), even though the
wide box has about $1.4$ orders of magnitude more prior volume, because the posterior widens with the prior.
The overtone mode is only weakly constrained by single-detector data. Its phase $\phi_{221}$ is essentially
unconstrained, and the $95\%$ width of its amplitude $A_{221}$ grows from about $4.8$ to $14$, nearly filling
the wider amplitude range.

\section{Baselines}
\label{app:baselines}

In this work we compare \M against two established neural-SBI methods.
{SNPE} is configured as in
\texttt{sbibm}~\citep{pmlr-v130-lueckmann21a}: a neural spline
flow (\texttt{nsf}) density estimator with $50$ hidden features, but with $R=2$ with $50\%$ of the budget allocated per round, and the
\texttt{sbibm} training schedule. 

{SNVI+fKL} is configured as in
\citet{DBLP:conf/iclr/GlocklerDM22}: a \texttt{resnet} ratio estimator
paired with an autoregressive spline flow proposal, but with $R = 5$ sequential
rounds with $256$ variational particles per round, and the
$\beta$-divergence schedule released with the original implementation.

\paragraph{Software versions.}
All baselines of the main benchmark run on \texttt{sbibm} 1.1.1
(\url{https://github.com/sbi-benchmark/sbibm}), extended with our prior scenarios and the differentiable
gravitational-wave task. SNPE runs on the \texttt{sbi} 0.16.0 fork released with the SNVI paper, and SNVI+fKL runs
on that fork together with its \texttt{sbivibm} runners, both from the original implementation
(\url{https://github.com/mackelab/snvi_repo}). These run under Python 3.10 and PyTorch 2.11. The Phase~2 ablations
of Appendix~\ref{app:tm-phase2-ablation} use the current \texttt{sbi} release, 0.26.1
(\url{https://github.com/sbi-dev/sbi}), under Python 3.12 and PyTorch 2.10.

The careful reader will notice that we use fewer sequential rounds for SNPE and SNVI+fKL than mentioned in Refs. \cite{pmlr-v130-lueckmann21a,DBLP:conf/iclr/GlocklerDM22}. Indeed, the results in Fig. \ref{fig:c2st_main} corresponding to the \texttt{sbibm} default slightly differ from these publications. This was done due to computational limitations. We also restrict the Phase-2 sequential posterior refinement of RSBI to a maximum of two rounds (matching SNPE) except for Lotka-Volterra for which we use three - in this case we checked the SNPE method with 3 rounds, yielding negligible difference to those recorded in the main results in Sec.~\ref{sec:exp_main}. Furthermore, for SNPE and SNVI we use the default training procedures of the sbi package. The neural density estimator (SNPE) and the classifier-based ratio estimator underlying SNVI are trained with Adam, holding out 10\% of the simulations for validation and stopping once the validation loss has not improved for 20 consecutive epochs; the weights with the best validation loss are kept. In addition, each round of SNVI fits a normalizing-flow variational posterior to the learned ratio by minimizing the forward KL divergence for at most 1,000 iterations, stopping early once the average slope of the running-mean loss over the last 50 iterations falls below $5 \times 10^{-5}$ in absolute value. We do note however that the authors of the SNVI paper mention in the appendix, that the benchmark convergence criteria were "chosen conservative to avoid early stopping". This could indicate using as many as 1000 iterations in each VI stage, therefore, differences with the published scores could potentially be a result of this.

Lastly, both the SNVI release and \texttt{sbi} 0.26.1 estimate the forward KL with self-normalised importance weights
$w_i \propto \exp(\log \tilde p(\theta_i) - \log q(\theta_i))$, where $\tilde p$ is the prior times the learned
ratio. They exponentiate the log-weights before normalising, so a single log-weight above about $88$ overflows in
single precision and the loss becomes undefined. On such a step the optimiser restores the last finite state of
$q$ and refits it to the prior before continuing. On wide priors, a sharp ratio estimator easily produces such
log-weights, and the step can recur until the variational posterior is spread back over the prior. We found
this in the Phase~2 ablation (Appendix~\ref{app:tm-phase2-ablation}), where it affected 9 of 10 observations at $10^4$ simulations. For that ablation we compute the weights in log space with a softmax, which gives the same estimator without overflow, and on a
non-finite step we only restore the last finite state of $q$. For the main benchmark we use the released implementation unmodified. We checked that starting from the prior, the variational posterior is broad and so the log-weights stay far below the overflow threshold. Rerunning three observations at $10^4$ simulations on the widest Two Moons prior with the stabilised weights fix gave no non-finite steps, and recovered essentially the same C2ST for the released implementation that was used to generate the main results of this paper.

\section{Volume estimator for \texorpdfstring{$\R$}{Rp}}
\label{sec:volume}

For each task and prior scenario we estimate the posterior support volume
non-parametrically from the $N_\text{ref} = 10{,}000$ reference samples
$\{\theta^{(n)}\}_{n=1}^{N_\text{ref}} \subset \mathbb{R}^{d_\theta}$. Writing
$\widehat{\mu}$ and $\widehat{\Sigma}$ for the sample mean and covariance, the
unit-Mahalanobis ellipsoid
$\mathcal{E} = \{\theta : (\theta - \widehat{\mu})^{\!\top}
\widehat{\Sigma}^{-1}(\theta - \widehat{\mu}) \le 1\}$ has Lebesgue volume
\begin{equation}
    V_\text{ell} \;=\; c_{d_\theta}\,\sqrt{\det \widehat{\Sigma}},
    \qquad
    c_{d_\theta} \;=\; \frac{\pi^{d_\theta/2}}{\Gamma\!\left(\tfrac{d_\theta}{2}+1\right)},
\end{equation}
where $c_{d_\theta}$ is the volume of the unit $d_\theta$-ball and
$\sqrt{\det \widehat{\Sigma}}$ maps it to the covariance ellipsoid. Working
in log-space for numerical stability,
\begin{equation}
    \log V_\text{ell}
    \;=\;
    \tfrac{1}{2}\log\det \widehat{\Sigma}
    \;+\;
    \tfrac{d_\theta}{2}\log\pi
    \;-\;
    \log \Gamma\!\left(\tfrac{d_\theta}{2}+1\right),
\end{equation}
with $\log\det\widehat{\Sigma}$ computed via \texttt{slogdet}. For a
rectangular prior with uncorrelated distributions, $\theta \sim \mathcal{U}(\prod_i [a_i, b_i])$, the ellipsoid
log-ratio is
\begin{equation}
    \log_{10}\!\mathcal{R}_p^{\text{ell.}}
    \;=\;
    \frac{1}{\ln 10}\bigl(\log V_\text{prior} - \log V_\text{ell}\bigr),
    \qquad
    V_\text{prior} \;=\; \prod_{i=1}^{d_\theta} (b_i - a_i).
\end{equation}

The ellipsoid is the one-$\sigma$ contour of the best-fit Gaussian to the
reference samples and therefore captures elongation and orientation that the
per-dimension marginal-box estimator
$V_\text{marg.} = \prod_i (q_{0.975}^{(i)} - q_{0.025}^{(i)})$ ignores. For uniform
priors it is also much smaller than the marginal box, which spans about $\pm 2\sigma$
per dimension, so $\log_{10}\!\mathcal{R}_p^{\text{ell.}}$ exceeds its marginal
counterpart in Table~\ref{tab:prior_scenarios}. For Gaussian priors, where prior and
posterior are measured with the same estimator, these constants cancel and the two
log-ratios differ by $\tfrac12(\log\det C_\text{prior} - \log\det\widehat{C})$, with $C$
the correlation matrices. This is positive for the isotropic Lotka--Volterra prior and
negative for the strongly correlated smoothness prior of the Bernoulli GLM, whose
ellipsoid ratio is therefore the smaller of the two. If $\widehat{\Sigma}$ were
rank-deficient we would floor its eigenvalues at $10^{-300}$ before taking the
log-determinant. This was not needed for any cell in Table~\ref{tab:prior_scenarios}.

For the Bernoulli GLM, whose prior is Gaussian,
$\theta \sim \mathcal{N}(0, \Sigma_\text{prior})$ with
$\Sigma_\text{prior} = (m\,B^{-1})^{-1}$ ($B^{-1}$ the smoothness precision of
\citet{pmlr-v130-lueckmann21a} and $m$ the scenario's precision multiplier), the
prior has no finite support volume, so we measure it with the same estimator as the
posterior. The prior's unit-Mahalanobis ellipsoid has volume
$c_{d_\theta}\sqrt{\det\Sigma_\text{prior}}$, and the ball constant cancels in the ratio,
\begin{equation}
    \log_{10}\!\mathcal{R}_p^{\text{ell.}}
    \;=\;
    \frac{1}{2\ln 10}\bigl(\log\det\Sigma_\text{prior} - \log\det\widehat{\Sigma}\bigr),
\end{equation}
while the marginal-box counterpart replaces the box widths $b_i - a_i$ by the prior's
central 95\% widths $2 z_{0.975}\,\sigma_i$, with $\sigma_i^2 = (\Sigma_\text{prior})_{ii}$.
The Lotka--Volterra prior is log-normal,
$\log\theta \sim \mathcal{N}(\mu_0, s^2 I)$, so we apply the same Gaussian formulae in
$\log\theta$, the coordinates in which the prior is Gaussian (and in which RSBI samples),
with $\Sigma_\text{prior} = s^2 I$ and $\widehat{\Sigma}$ the covariance of the
log-transformed reference samples.

\section{\M Hyperparameters}\label{sec:hyperparam_tables}

In this section we report hyperparameters used throughout the paper's various experiments.

\subsection{Main Benchmarks}
Tables~\ref{tab:hparams-two_moons}--\ref{tab:hparams-slcp} list the \M hyperparameters of every
prior $\times$ budget cell in \figref{fig:c2st_main}. The prior scenarios are defined in Table~\ref{tab:prior_scenarios}.

\begin{table}[p]
\caption{\M hyperparameters for the Two Moons results in Figure~\ref{fig:c2st_main}. One column per prior $\times$ budget cell. A value spanning several columns applies to all of them, and --- marks a setting that is not used in that cell.}
\label{tab:hparams-two_moons}
\centering\scriptsize\setlength{\tabcolsep}{2.5pt}\renewcommand{\arraystretch}{1.045}
\begin{tabular}{@{}lccccccccc@{}}\toprule
 & \multicolumn{3}{c}{default} & \multicolumn{3}{c}{wide10x} & \multicolumn{3}{c}{wide50x} \\
\cmidrule(lr){2-4} \cmidrule(lr){5-7} \cmidrule(lr){8-10}
 & $10^3$ & $10^4$ & $10^5$ & $10^3$ & $10^4$ & $10^5$ & $10^3$ & $10^4$ & $10^5$ \\\midrule
\multicolumn{10}{@{}l}{\emph{Simulation budget}}\\
Phase-1 simulations & 536 & 4768 & 54\,272 & \multicolumn{3}{c}{524} & \multicolumn{3}{c}{524} \\
NRE sims added, first round & 64 & 232 & 728 & 76 & 476 & 4476 & 76 & 476 & 4476 \\
NRE sims added, later rounds & 400 & 5000 & 45\,000 & 400 & 9000 & 95\,000 & 400 & 9000 & 95\,000 \\
NRE buffer size & 1000 & 10\,000 & 100\,000 & 1000 & 10\,000 & 100\,000 & 1000 & 10\,000 & 100\,000 \\
\midrule
\multicolumn{10}{@{}l}{\emph{Phase 1}}\\
Surrogate & \multicolumn{9}{c}{energy distance ($\epsilon$ from a 256-sample pilot, EMA 0.95)} \\
Temperature $T$ & 0.1 & 0.1 & 0.3 & \multicolumn{3}{c}{0.1} & \multicolumn{3}{c}{0.05} \\
MC replicates $K$ & \multicolumn{9}{c}{1} \\
Prior boundary softness $\lambda$ & 10 & 1 & 1 & \multicolumn{3}{c}{10} & \multicolumn{3}{c}{10} \\
SDE schedule & \multicolumn{3}{c}{Brownian} & \multicolumn{3}{c}{EDM} & \multicolumn{3}{c}{EDM} \\
$\sigma$ & 0.2 & 0.2 & 0.3 & \multicolumn{3}{c}{---} & \multicolumn{3}{c}{---} \\
$\sigma_{\min}$ / $\sigma_{\max}$ & \multicolumn{3}{c}{---} & \multicolumn{3}{c}{0.3 / 5} & \multicolumn{3}{c}{0.3 / 10} \\
$\rho$ & \multicolumn{3}{c}{---} & \multicolumn{3}{c}{2} & \multicolumn{3}{c}{2} \\
Integration steps & \multicolumn{3}{c}{128} & \multicolumn{3}{c}{128} & \multicolumn{3}{c}{256} \\
Buffer size & 128 & 1024 & 8192 & \multicolumn{3}{c}{32} & \multicolumn{3}{c}{32} \\
Resample size & 8 & 32 & 256 & \multicolumn{3}{c}{4} & \multicolumn{3}{c}{4} \\
Batch size & 8 & 32 & 256 & \multicolumn{3}{c}{4} & \multicolumn{3}{c}{4} \\
ASBS stages & 3 & 3 & 5 & \multicolumn{3}{c}{5} & \multicolumn{3}{c}{5} \\
Epochs per stage & 10 & 55 & 60 & \multicolumn{3}{c}{20} & \multicolumn{3}{c}{20} \\
Iterations per epoch & 50 & 100 & 100 & \multicolumn{3}{c}{15} & \multicolumn{3}{c}{15} \\
Learning rate / weight decay & \multicolumn{9}{c}{$5\times10^{-4}$ / $10^{-7}$} \\
LR warm-up steps & \multicolumn{9}{c}{100} \\
Clip grad norm & 1000 & 1000 & 100 & \multicolumn{3}{c}{1000} & \multicolumn{3}{c}{1000} \\
Clip energy grad norm & 1000 & 1000 & 100 & \multicolumn{3}{c}{1000} & \multicolumn{3}{c}{1000} \\
Controller MLP & \multicolumn{9}{c}{4 $\times$ 128, GELU} \\
WT height $h_{\text{wt}}$ & --- & 0.05 & $5\times10^{-4}$ & \multicolumn{3}{c}{---} & \multicolumn{3}{c}{---} \\
WT width $\sigma_\mathrm{cv}$ & --- & 0.07 & 0.07 & \multicolumn{3}{c}{---} & \multicolumn{3}{c}{---} \\
WT bias factor $\gamma$ & --- & 30 & 30 & \multicolumn{3}{c}{---} & \multicolumn{3}{c}{---} \\
WT grid bins & --- & 32 & 32 & \multicolumn{3}{c}{---} & \multicolumn{3}{c}{---} \\
WT grid range & --- & $\pm1.1$ & $\pm1.1$ & \multicolumn{3}{c}{---} & \multicolumn{3}{c}{---} \\
WT collective variable & --- & all of $\theta$ & all of $\theta$ & \multicolumn{3}{c}{---} & \multicolumn{3}{c}{---} \\
\midrule
\multicolumn{10}{@{}l}{\emph{Phase 2}}\\
Stages $R$ (NRE rounds) & \multicolumn{9}{c}{2} \\
Prior boundary $\lambda$ & 10 & 500\,000 & 500\,000 & \multicolumn{3}{c}{10} & \multicolumn{3}{c}{10} \\
SDE schedule & \multicolumn{9}{c}{EDM} \\
$\sigma_{\min}$ & 0.001 & $10^{-6}$ & $10^{-6}$ & \multicolumn{3}{c}{0.001} & \multicolumn{3}{c}{0.001} \\
$\sigma_{\max}$ & 0.2 & 0.2 & 0.1 & 0.2 & 0.3 & 0.3 & 0.2 & 0.3 & 0.3 \\
$\rho$ & 4 & 7 & 7 & \multicolumn{3}{c}{4} & \multicolumn{3}{c}{4} \\
Integration steps & \multicolumn{9}{c}{256} \\
Buffer size & 16\,384 & 8096 & 8096 & 16\,384 & 16\,384 & 8192 & \multicolumn{3}{c}{16\,384} \\
Resample size & 512 & 1024 & 512 & \multicolumn{3}{c}{512} & \multicolumn{3}{c}{512} \\
Batch size & 4096 & 1024 & 512 & 4096 & 4096 & 512 & \multicolumn{3}{c}{4096} \\
ASBS stages & 5 & 3 & 3 & \multicolumn{3}{c}{5} & \multicolumn{3}{c}{5} \\
Epochs per stage & 60 & 400 & 100 & \multicolumn{3}{c}{60} & \multicolumn{3}{c}{60} \\
Iterations per epoch & 10 & 20 & 100 & 10 & 10 & 100 & \multicolumn{3}{c}{10} \\
Learning rate & 0.01 & 0.001 & 0.001 & 0.01 & 0.01 & 0.001 & \multicolumn{3}{c}{0.01} \\
Weight decay & \multicolumn{9}{c}{$10^{-7}$} \\
LR warm-up steps & 1000 & 400 & 400 & \multicolumn{3}{c}{1000} & \multicolumn{3}{c}{1000} \\
Clip grad / energy grad norm & 100 / 100 & 100 / 500 & 100 / 800 & \multicolumn{3}{c}{100 / 100} & \multicolumn{3}{c}{100 / 100} \\
Controller MLP & \multicolumn{9}{c}{6 $\times$ 64, GELU} \\
WT height $h_{\text{wt}}$ & $5\times10^{-4}$ & --- & --- & \multicolumn{3}{c}{$5\times10^{-4}$} & \multicolumn{3}{c}{$5\times10^{-4}$} \\
WT width $\sigma_\mathrm{cv}$ & 0.02 & --- & --- & \multicolumn{3}{c}{0.02} & \multicolumn{3}{c}{0.02} \\
WT bias factor $\gamma$ & 10 & --- & --- & \multicolumn{3}{c}{10} & 10 & 20 & 20 \\
WT grid bins & 150 & --- & --- & \multicolumn{3}{c}{350} & \multicolumn{3}{c}{350} \\
WT grid range & $\pm1.1$ & --- & --- & \multicolumn{3}{c}{$\pm5.1$} & \multicolumn{3}{c}{$\pm5.1$} \\
WT collective variable & all of $\theta$ & --- & --- & \multicolumn{3}{c}{all of $\theta$} & \multicolumn{3}{c}{all of $\theta$} \\
\midrule
\multicolumn{10}{@{}l}{\emph{NRE}}\\
NRE MLP & 6 $\times$ 64 & 6 $\times$ 256 & 6 $\times$ 128 & 6 $\times$ 64 & 6 $\times$ 64 & 6 $\times$ 256 & 6 $\times$ 64 & 6 $\times$ 64 & 6 $\times$ 128 \\
NRE epochs, first round & 1500 & 500 & 50 & 1500 & 1500 & 25 & 1500 & 1500 & 50 \\
NRE epochs, later rounds & 2500 & 200 & 300 & 2500 & 500 & 100 & 2500 & 500 & 100 \\
NRE batch size & 128 & 200 & 200 & \multicolumn{3}{c}{128} & \multicolumn{3}{c}{128} \\
NRE learning rate & $5\times10^{-4}$ & 0.001 & 0.001 & \multicolumn{3}{c}{$5\times10^{-4}$} & \multicolumn{3}{c}{$5\times10^{-4}$} \\
NRE contrast classes $K_{\text{class}}$ & 9 & 100 & 100 & \multicolumn{3}{c}{9} & \multicolumn{3}{c}{9} \\
\bottomrule\end{tabular}
\end{table}

\begin{table}[p]
\caption{\M hyperparameters for the Lotka--Volterra results in Figure~\ref{fig:c2st_main}. One column per prior $\times$ budget cell. A value spanning several columns applies to all of them, and --- marks a setting that is not used in that cell. $^\dagger$~the Phase-1 simulations are discarded and the first NRE round trains on fresh draws from the Phase-1 sampler.}
\label{tab:hparams-lotka_volterra}
\centering\scriptsize\setlength{\tabcolsep}{2.5pt}\renewcommand{\arraystretch}{1.105}
\begin{tabular}{@{}lccccccccc@{}}\toprule
 & \multicolumn{3}{c}{$10^3$} & \multicolumn{3}{c}{$10^4$} & \multicolumn{3}{c}{$10^5$} \\
\cmidrule(lr){2-4} \cmidrule(lr){5-7} \cmidrule(lr){8-10}
 & default & medium & large & default & medium & large & default & medium & large \\\midrule
\multicolumn{10}{@{}l}{\emph{Simulation budget}}\\
Phase-1 simulations & \multicolumn{3}{c}{700} & \multicolumn{3}{c}{7000} & 49\,725 & 50\,000 & 80\,000 \\
NRE sims added, first round & \multicolumn{3}{c}{300} & \multicolumn{3}{c}{3000} & 275 & 0 & 20\,000 \\
NRE sims added, later rounds & \multicolumn{3}{c}{---} & \multicolumn{3}{c}{---} & 25\,000 & 25\,000 & --- \\
NRE buffer size & \multicolumn{3}{c}{---} & \multicolumn{3}{c}{---} & 100\,000 & 100\,000 & --- \\
Phase-1 simulation cap & \multicolumn{3}{c}{700} & \multicolumn{3}{c}{7000} & 50\,000 & 50\,000 & 80\,000 \\
NRE top-up to & \multicolumn{3}{c}{1000} & \multicolumn{3}{c}{10\,000} & 50\,000 & 50\,000 & 80\,000 \\
\midrule
\multicolumn{10}{@{}l}{\emph{Phase 1}}\\
Surrogate & \multicolumn{9}{c}{Gaussian NLL (adaptive variance)} \\
Temperature $T$ & \multicolumn{9}{c}{1} \\
MC replicates $K$ & \multicolumn{3}{c}{2} & \multicolumn{3}{c}{2} & \multicolumn{3}{c}{3} \\
ODE horizons & \multicolumn{9}{c}{3, 6, 10} \\
Horizon switch points & \multicolumn{3}{c}{0, 0.3, 0.6$\,\times$cap} & \multicolumn{3}{c}{0, 0.3, 0.6$\,\times$cap} & \multicolumn{3}{c}{0, 15\,000, 30\,000 sims} \\
Shooting until & \multicolumn{9}{c}{0.5} \\
SDE schedule & \multicolumn{9}{c}{EDM} \\
$\sigma_{\min}$ & \multicolumn{9}{c}{$10^{-5}$} \\
$\sigma_{\max}$ & \multicolumn{3}{c}{0.3} & \multicolumn{3}{c}{0.3} & 0.3 & 0.5 & 0.5 \\
$\rho$ & \multicolumn{9}{c}{5} \\
Integration steps & \multicolumn{9}{c}{64} \\
Buffer size & \multicolumn{3}{c}{100} & \multicolumn{3}{c}{100} & \multicolumn{3}{c}{2000} \\
Resample size & \multicolumn{3}{c}{50} & \multicolumn{3}{c}{50} & 25 & 50 & 50 \\
Batch size & \multicolumn{3}{c}{100} & \multicolumn{3}{c}{100} & \multicolumn{3}{c}{1000} \\
ASBS stages & \multicolumn{3}{c}{5} & \multicolumn{3}{c}{7} & \multicolumn{3}{c}{7} \\
Epochs per stage & \multicolumn{3}{c}{2} & \multicolumn{3}{c}{19} & 146 & 78 & 130 \\
Iterations per epoch & \multicolumn{3}{c}{200} & \multicolumn{3}{c}{200} & \multicolumn{3}{c}{20} \\
Learning rate / weight decay & \multicolumn{9}{c}{$5\times10^{-4}$ / $10^{-7}$} \\
LR warm-up steps & \multicolumn{9}{c}{10} \\
Clip grad / energy grad norm & \multicolumn{9}{c}{$10^{6}$ / 256} \\
Controller MLP & \multicolumn{9}{c}{6 $\times$ 128, GELU} \\
\midrule
\multicolumn{10}{@{}l}{\emph{Phase 2}}\\
Stages $R$ (NRE rounds) & \multicolumn{3}{c}{1} & \multicolumn{3}{c}{1} & 3 & 3 & 1$^\dagger$ \\
SDE schedule & \multicolumn{9}{c}{EDM} \\
$\sigma_{\min}$ / $\sigma_{\max}$ & \multicolumn{9}{c}{$10^{-5}$ / 1} \\
$\rho$ & \multicolumn{3}{c}{7} & \multicolumn{3}{c}{7} & \multicolumn{3}{c}{5} \\
Integration steps & \multicolumn{9}{c}{128} \\
Buffer size & \multicolumn{9}{c}{2000} \\
Resample size & \multicolumn{9}{c}{512} \\
Batch size & \multicolumn{3}{c}{128} & \multicolumn{3}{c}{128} & \multicolumn{3}{c}{500} \\
Stages $\times$ epochs $\times$ iterations & \multicolumn{3}{c}{5 $\times$ 50 $\times$ 50} & \multicolumn{3}{c}{5 $\times$ 50 $\times$ 50} & \multicolumn{3}{c}{7 $\times$ 150 $\times$ 20} \\
Learning rate / weight decay & \multicolumn{9}{c}{$5\times10^{-4}$ / 0.003} \\
LR warm-up steps & \multicolumn{9}{c}{100} \\
Clip grad norm & \multicolumn{3}{c}{$10^{8}$} & \multicolumn{3}{c}{$10^{8}$} & $10^{8}$ & 100 & 100 \\
Clip energy grad norm & \multicolumn{3}{c}{$10^{8}$} & \multicolumn{3}{c}{$10^{8}$} & $10^{6}$ & 1000 & 1000 \\
Controller MLP & \multicolumn{9}{c}{6 $\times$ 128, GELU} \\
\midrule
\multicolumn{10}{@{}l}{\emph{NRE}}\\
NRE MLP & \multicolumn{9}{c}{6 $\times$ 128} \\
NRE epochs, first round & \multicolumn{3}{c}{200} & \multicolumn{3}{c}{200} & \multicolumn{3}{c}{300} \\
NRE epochs, later rounds & \multicolumn{3}{c}{---} & \multicolumn{3}{c}{---} & 150 & 150 & --- \\
NRE batch size / learning rate & \multicolumn{3}{c}{1000 / $6\times10^{-4}$} & \multicolumn{3}{c}{1000 / $6\times10^{-4}$} & \multicolumn{3}{c}{350 / $6\times10^{-4}$} \\
NRE contrast classes $K_{\text{class}}$ & \multicolumn{9}{c}{10} \\
NRE input standardised & \multicolumn{9}{c}{yes} \\
\midrule
\multicolumn{10}{@{}l}{\emph{Run-time guards}}\\
NRE $\log r$ cap quantile & \multicolumn{9}{c}{0.999} \\
NRE $\log r$ cap margin & \multicolumn{9}{c}{2} \\
NRE truncation $k$ & \multicolumn{3}{c}{3} & \multicolumn{3}{c}{3} & 3 & 3 & 2 \\
NRE truncation trim quantile & \multicolumn{3}{c}{0.99} & \multicolumn{3}{c}{0.99} & 0.99 & 0.99 & 0.9 \\
NRE truncation strength & \multicolumn{9}{c}{50} \\
Round draw filter (prior s.d.) & \multicolumn{9}{c}{5} \\
Phase-2 whitening trim quantile & \multicolumn{3}{c}{0.99} & \multicolumn{3}{c}{0.99} & 0.99 & 0.99 & 0.9 \\
\bottomrule\end{tabular}
\end{table}

\begin{table}[p]
\caption{\M hyperparameters for the Bernoulli GLM results in Figure~\ref{fig:c2st_main}. One column per prior $\times$ budget cell. A value spanning several columns applies to all of them, and --- marks a setting that is not used in that cell.}
\label{tab:hparams-bernoulli_glm}
\centering\footnotesize\setlength{\tabcolsep}{3pt}\renewcommand{\arraystretch}{1.225}
\begin{tabular}{@{}lccccccccc@{}}\toprule
 & \multicolumn{3}{c}{$10^3$} & \multicolumn{3}{c}{$10^4$} & \multicolumn{3}{c}{$10^5$} \\
\cmidrule(lr){2-4} \cmidrule(lr){5-7} \cmidrule(lr){8-10}
 & default & medium & large & default & medium & large & default & medium & large \\\midrule
\multicolumn{10}{@{}l}{\emph{Simulation budget}}\\
Phase-1 simulations & \multicolumn{3}{c}{250} & \multicolumn{3}{c}{2500} & 22\,980 & 24\,980 & 24\,980 \\
NRE sims added, first round & \multicolumn{3}{c}{0} & \multicolumn{3}{c}{0} & 2020 & 20 & 20 \\
NRE sims added, later rounds & \multicolumn{3}{c}{750} & \multicolumn{3}{c}{7500} & \multicolumn{3}{c}{75\,000} \\
NRE buffer size & \multicolumn{3}{c}{1000} & \multicolumn{3}{c}{10\,000} & \multicolumn{3}{c}{100\,000} \\
\midrule
\multicolumn{10}{@{}l}{\emph{Phase 1}}\\
Surrogate & \multicolumn{9}{c}{energy distance ($\epsilon=0.914$)} \\
Temperature $T$ & \multicolumn{9}{c}{1} \\
MC replicates $K$ & \multicolumn{9}{c}{2} \\
Simulator temperature & \multicolumn{9}{c}{0.5} \\
SDE schedule & \multicolumn{9}{c}{EDM} \\
$\sigma_{\min}$ & \multicolumn{9}{c}{$10^{-5}$} \\
$\sigma_{\max}$ & 1.5 & 10 & 10 & 1.5 & 10 & 10 & 1.5 & 10 & 10 \\
$\rho$ & \multicolumn{9}{c}{5} \\
Integration steps & \multicolumn{9}{c}{128} \\
Buffer size & 51 & 26 & 26 & 501 & 251 & 251 & \multicolumn{3}{c}{2500} \\
Resample size & \multicolumn{3}{c}{1} & \multicolumn{3}{c}{1} & \multicolumn{3}{c}{10} \\
Batch size & 51 & 26 & 26 & 501 & 251 & 251 & \multicolumn{3}{c}{512} \\
ASBS stages & 5 & 3 & 3 & 5 & 3 & 3 & 7 & 3 & 3 \\
Epochs per stage & 25 & 50 & 50 & 250 & 500 & 500 & 225 & 500 & 500 \\
Iterations per epoch & \multicolumn{3}{c}{50} & \multicolumn{3}{c}{5} & \multicolumn{3}{c}{5} \\
Learning rate / weight decay & \multicolumn{9}{c}{$5\times10^{-4}$ / $10^{-7}$} \\
LR warm-up steps & \multicolumn{9}{c}{10} \\
Clip grad / energy grad norm & \multicolumn{9}{c}{$10^{6}$ / $10^{6}$} \\
Controller MLP & \multicolumn{9}{c}{6 $\times$ 128, GELU} \\
\midrule
\multicolumn{10}{@{}l}{\emph{Phase 2}}\\
Stages $R$ (NRE rounds) & \multicolumn{9}{c}{2} \\
SDE schedule & \multicolumn{9}{c}{EDM} \\
$\sigma_{\min}$ & \multicolumn{9}{c}{$10^{-5}$} \\
$\sigma_{\max}$ & 3 & 10 & 15 & 3 & 10 & 15 & 2.5 & 5 & 15 \\
$\rho$ & \multicolumn{9}{c}{7} \\
Integration steps & \multicolumn{9}{c}{128} \\
Buffer size & \multicolumn{9}{c}{2000} \\
Resample size & \multicolumn{9}{c}{512} \\
Batch size & \multicolumn{9}{c}{128} \\
Stages $\times$ epochs $\times$ iterations & \multicolumn{3}{c}{5 $\times$ 50 $\times$ 50} & \multicolumn{3}{c}{5 $\times$ 50 $\times$ 50} & \multicolumn{3}{c}{7 $\times$ 50 $\times$ 50} \\
Learning rate / weight decay & \multicolumn{9}{c}{$5\times10^{-4}$ / 0.003} \\
LR warm-up steps & \multicolumn{9}{c}{100} \\
Clip grad / energy grad norm & \multicolumn{9}{c}{$10^{8}$ / $10^{8}$} \\
Controller MLP & \multicolumn{9}{c}{6 $\times$ 128, GELU} \\
\midrule
\multicolumn{10}{@{}l}{\emph{NRE}}\\
NRE MLP & \multicolumn{9}{c}{6 $\times$ 128} \\
NRE epochs, first round & \multicolumn{9}{c}{150} \\
NRE epochs, later rounds & \multicolumn{3}{c}{50} & \multicolumn{3}{c}{50} & 50 & 25 & 25 \\
NRE batch size / learning rate & \multicolumn{3}{c}{50 / $6\times10^{-4}$} & \multicolumn{3}{c}{300 / $6\times10^{-4}$} & \multicolumn{3}{c}{300 / $6\times10^{-4}$} \\
NRE contrast classes $K_{\text{class}}$ & \multicolumn{3}{c}{16} & \multicolumn{3}{c}{100} & \multicolumn{3}{c}{100} \\
\bottomrule\end{tabular}
\end{table}

\begin{table}[p]
\caption{\M hyperparameters for the SLCP results in Figure~\ref{fig:c2st_main}. One column per prior $\times$ budget cell. A value spanning several columns applies to all of them, and --- marks a setting that is not used in that cell.}
\label{tab:hparams-slcp}
\centering\footnotesize\setlength{\tabcolsep}{3pt}\renewcommand{\arraystretch}{1.13}
\begin{tabular}{@{}lccccccccc@{}}\toprule
 & \multicolumn{3}{c}{$10^3$} & \multicolumn{3}{c}{$10^4$} & \multicolumn{3}{c}{$10^5$} \\
\cmidrule(lr){2-4} \cmidrule(lr){5-7} \cmidrule(lr){8-10}
 & default & wide5x & wide10x & default & wide5x & wide10x & default & wide5x & wide10x \\\midrule
\multicolumn{10}{@{}l}{\emph{Simulation budget}}\\
Phase-1 simulations & \multicolumn{3}{c}{576} & \multicolumn{3}{c}{4896} & \multicolumn{3}{c}{13\,248} \\
NRE sims added, first round & \multicolumn{3}{c}{424} & \multicolumn{3}{c}{104} & \multicolumn{3}{c}{752} \\
NRE sims added, later rounds & \multicolumn{3}{c}{---} & \multicolumn{3}{c}{5000} & \multicolumn{3}{c}{86\,000} \\
NRE buffer size & \multicolumn{3}{c}{---} & \multicolumn{3}{c}{8096} & \multicolumn{3}{c}{50\,000} \\
\midrule
\multicolumn{10}{@{}l}{\emph{Phase 1}}\\
Surrogate & \multicolumn{9}{c}{Gaussian NLL (adaptive variance)} \\
Temperature $T$ & \multicolumn{9}{c}{1} \\
MC replicates $K$ & \multicolumn{3}{c}{3} & \multicolumn{3}{c}{3} & \multicolumn{3}{c}{4} \\
Prior boundary softness $\lambda$ & \multicolumn{9}{c}{20} \\
SDE schedule & \multicolumn{9}{c}{EDM} \\
$\sigma_{\min}$ / $\sigma_{\max}$ & \multicolumn{9}{c}{0.01 / 1} \\
$\rho$ & \multicolumn{9}{c}{7} \\
Integration steps & \multicolumn{9}{c}{64} \\
Buffer size & \multicolumn{9}{c}{128} \\
Resample size & \multicolumn{9}{c}{16} \\
Batch size & \multicolumn{9}{c}{32} \\
Stages $\times$ epochs $\times$ iterations & \multicolumn{3}{c}{1 $\times$ 5 $\times$ 30} & \multicolumn{3}{c}{1 $\times$ 95 $\times$ 30} & \multicolumn{3}{c}{1 $\times$ 200 $\times$ 30} \\
Learning rate / weight decay & \multicolumn{9}{c}{0.005 / 0.001} \\
LR warm-up steps & \multicolumn{9}{c}{50} \\
Clip grad / energy grad norm & \multicolumn{9}{c}{1 / 100} \\
Controller MLP & \multicolumn{9}{c}{6 $\times$ 64, GELU} \\
\midrule
\multicolumn{10}{@{}l}{\emph{Phase 2}}\\
Stages $R$ (NRE rounds) & \multicolumn{3}{c}{1} & \multicolumn{3}{c}{2} & \multicolumn{3}{c}{2} \\
SDE schedule & \multicolumn{9}{c}{EDM} \\
$\sigma_{\min}$ & \multicolumn{9}{c}{$10^{-5}$} \\
$\sigma_{\max}$ & \multicolumn{3}{c}{3} & 3 & 5 & 5 & \multicolumn{3}{c}{5} \\
$\rho$ & \multicolumn{9}{c}{7} \\
Integration steps & \multicolumn{9}{c}{256} \\
Buffer size & \multicolumn{9}{c}{2048} \\
Resample size & \multicolumn{3}{c}{128} & \multicolumn{3}{c}{1024} & \multicolumn{3}{c}{128} \\
Batch size & \multicolumn{9}{c}{128} \\
Stages $\times$ epochs $\times$ iterations & \multicolumn{3}{c}{5 $\times$ 200 $\times$ 100} & \multicolumn{3}{c}{5 $\times$ 500 $\times$ 100} & \multicolumn{3}{c}{5 $\times$ 500 $\times$ 100} \\
Learning rate / weight decay & \multicolumn{9}{c}{0.001 / 0.001} \\
LR warm-up steps & \multicolumn{9}{c}{100} \\
Clip grad norm & \multicolumn{9}{c}{100} \\
Clip energy grad norm & 1 & --- & 100 & 100 & --- & --- & \multicolumn{3}{c}{---} \\
Controller MLP & \multicolumn{9}{c}{4 $\times$ 128, GELU} \\
WT height $h_{\text{wt}}$ & \multicolumn{9}{c}{0.005} \\
WT width $\sigma_\mathrm{cv}$ & \multicolumn{9}{c}{0.1} \\
WT bias factor $\gamma$ & \multicolumn{9}{c}{10} \\
WT grid bins & \multicolumn{9}{c}{128} \\
WT grid range & $\pm3.5$ & $\pm5.5$ & $\pm10.5$ & $\pm3.5$ & $\pm5.5$ & $\pm10.5$ & $\pm3.5$ & $\pm5.5$ & $\pm10.5$ \\
WT collective variable & \multicolumn{9}{c}{$(\theta_{3}, \theta_{4})$} \\
\midrule
\multicolumn{10}{@{}l}{\emph{NRE}}\\
NRE MLP & \multicolumn{9}{c}{6 $\times$ 64} \\
NRE epochs, first round & \multicolumn{3}{c}{300} & \multicolumn{3}{c}{100} & \multicolumn{3}{c}{500} \\
NRE epochs, later rounds & \multicolumn{3}{c}{---} & \multicolumn{3}{c}{100} & \multicolumn{3}{c}{100} \\
NRE batch size / learning rate & \multicolumn{3}{c}{10 / $5\times10^{-4}$} & \multicolumn{3}{c}{128 / $5\times10^{-4}$} & \multicolumn{3}{c}{1028 / $5\times10^{-4}$} \\
NRE contrast classes $K_{\text{class}}$ & \multicolumn{9}{c}{4} \\
\bottomrule\end{tabular}
\end{table}

\clearpage

\subsection{\texttt{gw\_ringdown}}

Table~\ref{tab:hparams-gw_ringdown} lists the \M hyperparameters of the \texttt{gw\_ringdown} experiment
(Table~\ref{tab:gw} and \figref{fig:gw_corner}) for the two prior scenarios of Table~\ref{tab:prior_scenarios}.
All ten observations of a scenario use the same configuration; the sampler runs on the prior box rescaled to
$[0, 10]^6$, so the schedule and well-tempered settings are given in those coordinates.

\begin{table}[!ht]
\caption{\M hyperparameters for the \texttt{gw\_ringdown} results in Table~\ref{tab:gw} and Figure~\ref{fig:gw_corner} ($N_\text{sim} = 10^5$). The default and wide prior scenarios share every setting.}
\label{tab:hparams-gw_ringdown}
\centering\small\setlength{\tabcolsep}{3pt}\renewcommand{\arraystretch}{1.0}
\makebox[\linewidth][c]{%
\begin{tabular}[t]{@{}lc@{}}\toprule
\multicolumn{2}{@{}l}{\emph{Simulation budget}}\\
Phase-1 simulations & 59\,900 \\
NRE sims added, first round & 100 \\
NRE sims added, later rounds & 40\,000 \\
NRE buffer size & 100\,000 \\
\midrule
\multicolumn{2}{@{}l}{\emph{Phase 1}}\\
Sampler coordinates & prior box $\mapsto [0, 10]^6$ \\
Surrogate & energy distance ($\epsilon=1$) \\
Temperature $T$ & 0.025 \\
MC replicates $K$ & 4 \\
Prior boundary softness $\lambda$ & 20 \\
SDE schedule & EDM \\
$\sigma_{\min}$ / $\sigma_{\max}$ & $10^{-9}$ / 1 \\
$\rho$ & 5 \\
Integration steps & 256 \\
Buffer size & 2000 \\
Resample size & 25 \\
Batch size & 1000 \\
Stages $\times$ epochs $\times$ iterations & 3 $\times$ 260 $\times$ 20 \\
Learning rate / weight decay & 0.01 / $10^{-7}$ \\
LR warm-up steps & 975 \\
Clip grad / energy grad norm & 1000 / 1000 \\
Controller MLP & 6 $\times$ 128, GELU \\
WT height $h_{\text{wt}}$ & 0.05 \\
WT width $\sigma_\mathrm{cv}$ & 0.5 \\
WT bias factor $\gamma$ & 10 \\
WT grid bins & 60 \\
WT grid range & $[-0.5, 10.5]$ \\
WT collective variable & $(\theta_{1}, \theta_{2})$ \\
\bottomrule\end{tabular}%
\hspace{1em}%
\begin{tabular}[t]{@{}lc@{}}\toprule
\multicolumn{2}{@{}l}{\emph{Phase 2}}\\
Stages $R$ (NRE rounds) & 2 \\
SDE schedule & EDM \\
$\sigma_{\min}$ / $\sigma_{\max}$ & $10^{-5}$ / 1 \\
$\rho$ & 5 \\
Integration steps & 256 \\
Buffer size & 5000 \\
Resample size & 256 \\
Batch size & 256 \\
Stages $\times$ epochs $\times$ iterations & 7 $\times$ 200 $\times$ 100 \\
Learning rate / weight decay & 0.001 / 0.003 \\
LR warm-up steps & 100 \\
Clip grad / energy grad norm & 100 / 1000 \\
Controller MLP & 4 $\times$ 128, GELU \\
Source reweighed by WT bias & yes \\
\midrule
\multicolumn{2}{@{}l}{\emph{NRE}}\\
NRE MLP & 4 $\times$ 64 \\
NRE epochs, first round & 200 \\
NRE epochs, later rounds & 200 \\
NRE batch size / learning rate & 600 / $6\times10^{-4}$ \\
NRE contrast classes $K_{\text{class}}$ & 20 \\
\midrule
\multicolumn{2}{@{}l}{\emph{Run-time guards}}\\
Phase-2 whitening trim quantile & 0.99 \\
Phase-2 whitening fit samples & 10\,000 \\
\bottomrule\end{tabular}%
}
\end{table}

\clearpage

\end{document}